\documentclass{article}

\PassOptionsToPackage{comma, numbers, sort}{natbib}
\usepackage[final]{neurips_2025}

\usepackage[utf8]{inputenc} %
\usepackage[T1]{fontenc}    %
\usepackage{url}            %
\usepackage{booktabs}       %
\usepackage{amsfonts}       %
\usepackage{nicefrac}       %
\usepackage{microtype}      %
\usepackage{xcolor}         %
\usepackage{tcolorbox}
\usepackage{soul}
\usepackage{caption}
\usepackage{subcaption}
\usepackage{wrapfig}
\usepackage{lipsum}
\usepackage{enumitem}
\usepackage{multirow}
\usepackage{tikz}
\usetikzlibrary{arrows.meta}
\usepackage{basic_commands}
\usepackage{algorithm}
\usepackage[endLComment=,italicComments=false]{algpseudocodex}

\definecolor{myred}{HTML}{F54254}
\definecolor{myorange}{HTML}{FFB135}
\definecolor{mygreen}{HTML}{10BD35}
\definecolor{myblue}{HTML}{598BE7}
\definecolor{mypurple}{HTML}{9A1C6B}
\definecolor{plgray}{HTML}{999999}
\renewcommand{\tt}[1]{\texttt{#1}}
\renewcommand{\phi}{\varphi}

\newcommand{\pl}[1]{{\color{plgray} #1}}

\hypersetup{urlcolor=myblue,citecolor=myblue,linkcolor=myblue}

\newcommand{\cutsectionup}{\vspace{0pt}}
\newcommand{\cutsectiondown}{\vspace{0pt}}
\newcommand{\cutsubsectionup}{\vspace{0pt}}
\newcommand{\cutsubsectiondown}{\vspace{0pt}}

\setlist[itemize]{itemsep=1pt, leftmargin=20pt}
\newtcolorbox{mybox}[1][]
{
  colframe=myblue,
  colback=myblue!8,
  coltitle=white,
  toprule=1pt,
  titlerule=1pt,
  leftrule=1pt,
  rightrule=1pt,
  bottomrule=1pt,
  #1,
}

\title{Horizon Reduction Makes RL Scalable}

\author{%
  \textbf{Seohong Park}$^{1}$ \quad \textbf{Kevin Frans}$^{1}$ \quad \textbf{Deepinder Mann}$^{1}$ \\
  \textbf{Benjamin Eysenbach}$^{2}$ \quad \textbf{Aviral Kumar}$^{3}$ \quad \textbf{Sergey Levine}$^{1}$ \\
  $^{1}$University of California, Berkeley \quad $^{2}$Princeton University \quad $^{3}$Carnegie Mellon University \\
  \texttt{seohong@berkeley.edu} \\
}

\begin{document}

\maketitle

\begin{abstract}
In this work, we study the \emph{scalability} of offline reinforcement learning (RL) algorithms.
In principle, a truly scalable offline RL algorithm should be able to solve any given problem,
regardless of its complexity, given sufficient data, compute, and model capacity.
We investigate if and how current offline RL algorithms match up to this promise
on diverse, challenging, previously unsolved tasks,
using datasets up to $1000\times$ larger than typical offline RL datasets.
We observe that despite scaling up data,
many existing offline RL algorithms exhibit poor scaling behavior,
saturating well below the maximum performance.
We hypothesize that the \emph{horizon} is the main cause behind the poor scaling of offline RL.
We empirically verify this hypothesis through several analysis experiments,
showing that long horizons indeed present a fundamental barrier to scaling up offline RL.
We then show that various horizon reduction%
\footnote{Throughout this work, we use the term ``horizon reduction'' to refer to techniques
that reduce the \emph{effective} decision horizon,
such as $n$-step returns and hierarchical policies.}
techniques substantially enhance scalability on challenging tasks.
Based on our insights, we also introduce a minimal yet scalable method named SHARSA that effectively reduces the horizon.
SHARSA achieves the best asymptotic performance and scaling behavior among our evaluation methods,
showing that explicitly reducing the horizon unlocks the scalability of offline RL.

Code: \url{https://github.com/seohongpark/horizon-reduction}
\end{abstract}

\cutsectionup
\section{Introduction}
\cutsectiondown
\label{sec:intro}
Scalability,
the ability to consistently improve performance with more data and compute,
is at the core of the success of modern machine learning algorithms,
across natural language processing (NLP), computer vision (CV), and robotics.
In this work, we are interested in the scalability of \emph{offline reinforcement learning (RL)},
a framework that can leverage large-scale offline datasets to learn performant policies.
While prior works have shown that
current offline RL methods scale to \emph{more} (but not necessarily harder) tasks
with larger models and datasets~\citep{scaledql_kumar2023, pac_springenberg2024},
it remains unclear how RL scales with data to \emph{more challenging} tasks,
especially those that require more complex, longer-horizon sequential decision making.

Our main question, posed informally is:

\begin{minipage}{\linewidth}
\centering
\emph{To what extent can current offline RL algorithms solve complex tasks}\\
\emph{simply by scaling up data and compute?}
\end{minipage}

In principle, a truly scalable offline RL algorithm should be able to master
\emph{any} given task, \emph{no matter how complex and long-horizon it is},
given a sufficient amount of data (of sufficient coverage),
compute, and model capacity.
Studying how current offline RL algorithms live up to this promise is important,
because it will tell us whether we are ready to scale existing offline RL methods,
or if we must further improve offline RL algorithms before scaling them.

\begin{figure}[t!]
    \centering
    \vspace{-10pt}
    \includegraphics[width=1.0\textwidth]{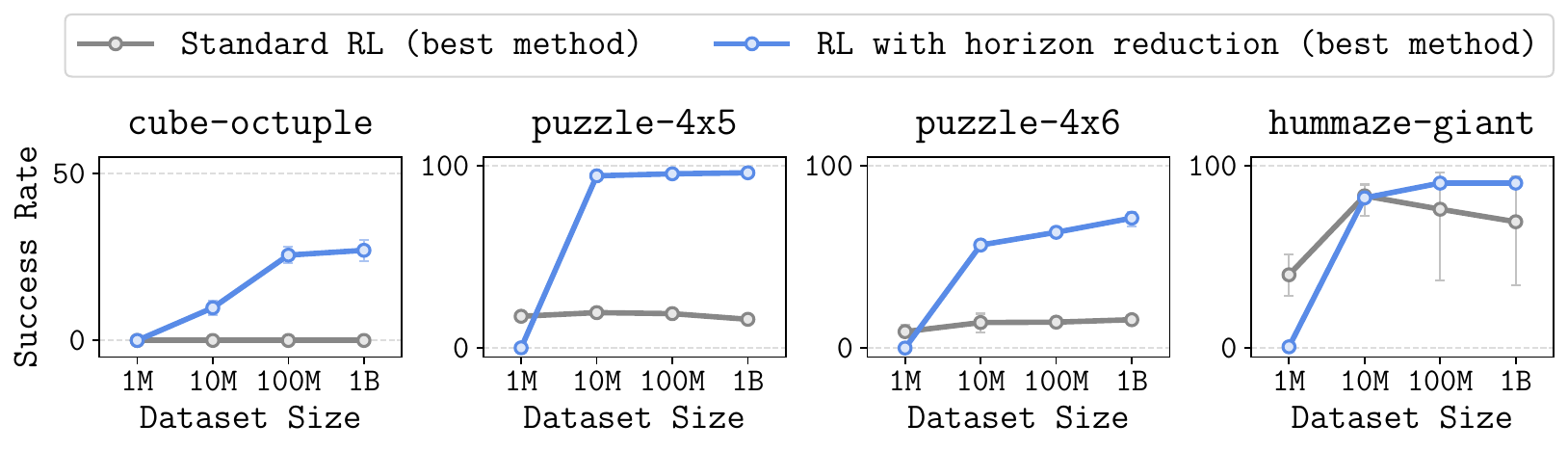}
    \vspace{-10pt}
    \caption{
    \footnotesize
    \textbf{Horizon reduction makes RL scalable.}
    Standard offline RL methods struggle to scale on highly challenging tasks,
    not improving performance with more data.
    We show that this is mainly because the \emph{long horizon} can fundamentally inhibit scaling,
    and that horizon reduction techniques unlock the scaling of offline RL.
    }
    \label{fig:teaser}
\end{figure}

To answer this question, we generate large-scale datasets for tasks
that require highly complex, long-horizon reasoning,
and study how current offline RL algorithms scale with data.
Specifically, on complex simulated robotics tasks across diverse domains in OGBench~\citep{ogbench_park2025},
we collect a dataset with up to \textbf{one billion} transitions for each environment,
which is $1000\times$ larger than standard $1$M-sized offline RL datasets used in prior work~\citep{d4rl_fu2020, ogbench_park2025}.
To isolate the fundamental sequential decision-making capabilities of RL algorithms,
we also idealize environments by removing other potential confounding factors,
such as visual representation learning.
In these controlled yet challenging environments,
we evaluate the performance of state-of-the-art offline RL algorithms while varying the amount of data.

We observe that many existing offline RL algorithms
struggle to scale,
even with orders of magnitude more data in these idealized environments.
Specifically, we show algorithms
such as IQL~\citep{iql_kostrikov2022}, SAC+BC~(\Cref{sec:algo_flat}), CRL~\citep{crl_eysenbach2022}, and FQL~\citep{fql_park2025}
often either completely fail to solve complex tasks,
or require an excessive amount of compute and model capacity
to reach even moderate performance.
Their performance
often saturates far below the maximum possible performance (\Cref{fig:teaser}), especially on complex, long-horizon tasks,
suggesting that there exist scalability challenges in offline RL.
We hypothesize that the reason behind this poor scaling 
is due to \textbf{the curse of horizon} in both value learning and policy learning.
In value learning,
we argue that the temporal difference (TD) learning objective used in many offline RL algorithms
has a fundamental limitation that inhibits scaling to longer horizons:
biases (errors) in the target Q values \emph{accumulate over the horizon}.
Through controlled analysis experiments,
we show that this bias accumulation is strongly correlated with poor performance.
Moreover, we show that increasing the model size or adjusting other hyperparameters, does \emph{not} effectively
mitigate this issue,
suggesting that the horizon fundamentally hinders scaling.
In policy learning,
we argue that the complexity of the mapping between states and optimal actions
in long-horizon tasks poses a major challenge,
and support this claim with experiments.
We then demonstrate that
methods that
explicitly reduce the value or policy horizons%
exhibit substantially better scaling (\Cref{fig:teaser}).
For example, we show that even simple techniques to reduce the value horizon,
such as $n$-step returns,
can substantially improve scaling curves and even asymptotic performance.
Based on our insights,
we also propose a minimal yet scalable RL method
called \textbf{SHARSA}
that reduces both the value and policy horizons.
Our method relies only on simple objectives that do \emph{not} require excessive hyperparameter tuning,
such as SARSA and behavioral cloning,
while effectively reducing the horizon.
Despite the simplicity, we show that
SHARSA generally exhibits the best scaling behavior and asymptotic performance among our evaluation methods.
\textbf{Contributions.}
Our main contributions are threefold.
First, through our $1$B-scale data-scaling analysis,
we empirically demonstrate that many standard offline RL algorithms
scale poorly on complex, long-horizon tasks.
Second, we identify the \emph{horizon} as a main obstacle to RL scaling,
and empirically show that horizon reduction techniques can effectively address this challenge. %
Third, we propose a simple method, SHARSA,
\mbox{that exhibits strong asymptotic performance and scaling behavior.}

\cutsectionup
\section{Related work}
\cutsectiondown
\label{sec:related}

\textbf{Offline RL.} Offline RL aims to train a reward-maximizing policy from a static dataset without online interactions~\citep{offline_levine2020}.
The main challenge in offline RL is
to maximize rewards while staying close to the dataset distribution to avoid distributional shift.
Previous works have proposed a number of techniques to address this challenge
based on
behavioral regularization~\citep{brac_wu2019, td3bc_fujimoto2021, rebrac_tarasov2023, fql_park2025},
conservatism~\citep{cql_kumar2020},
weighted regression~\citep{rwr_peters2007, awr_peng2019, crr_wang2020},
in-sample maximization~\citep{iql_kostrikov2022, sql_xu2023, xql_garg2023},
uncertainty minimization~\citep{edac_an2021, sacrnd_nikulin2023},
one-step RL~\citep{onestep_brandfonbrener2021, crl_eysenbach2022},
model-based RL~\citep{morel_kidambi2020, mopo_yu2020, combo_yu2021},
and more~\citep{optidice_lee2021, dt_chen2021, tt_janner2021, diffuser_janner2022, idql_hansenestruch2023, qrl_wang2023, dualrl_sikchi2024}.
Among these methods, we mainly consider three distinct representative model-free algorithms
that have been reported to achieve state-of-the-art performance on standard benchmarks~\citep{corl_tarasov2023, ogbench_park2025, fql_park2025},
IQL~\citep{iql_kostrikov2022}, SAC+BC (\Cref{sec:algo_flat}), and CRL~\citep{crl_eysenbach2022},
as the main subject of our scaling analysis.
We leave the scaling study of offline model-based RL for future work.

\textbf{Scaling RL.}
Prior work has studied the scalability of RL algorithms in various aspects.
Many previous works focus on scaling \emph{online} RL
to solve more diverse tasks with larger models~\citep{tdmpc2_hansen2024, bro_nauman2024, dreamerv3_hafner2025},
more compute~\cite{rlscale_rybkin2025},
and parallel simulation~\cite{impala_espeholt2018, pql_li2023, sapg_singla2024, pqn_gallici2025}.
More recently, several works have also explored the scaling of online, on-policy RL on language tasks~\citep{openaio1_jaech2024, kimi_team2025}.
Unlike these works that study online RL, we focus on the scalability of \emph{offline} RL algorithms.

Many prior works on scaling offline RL focus on scalability to \emph{more} tasks
by training a large, multi-task agent that is capable of solving more diverse (but not necessarily harder)
tasks~\citep{gato_reed2022, mgdt_lee2022, scaledql_kumar2023, qt_chebotar2023, pac_springenberg2024, jowa_cheng2025}.
Unlike these works,
we focus the ability to solve more \emph{challenging} tasks that require highly complex sequential decision making, 
given more data and compute.
This is analogous to scalability along the ``depth'' axis, as opposed to the ``width'' axis that the prior works have explored.
This is an important, complementary axis to study, as it will let us know
whether offline RL is currently bottlenecked by the amount of data and compute, or the fundamental learning capabilities of algorithms.
A closely related work is \citet{bottleneck_park2024},
which shows that poor policy extraction and generalization can bottleneck the scaling of offline RL.
We study the scalability of offline RL to more complex (in particular, longer-horizon) tasks when these bottlenecks are removed,
with the use of more expressive policy classes and with datasets $100 \times$ as large as this prior work.
This makes our study distinct from and complementary to the challenges that \citet{bottleneck_park2024} highlight.
\textbf{Horizon reduction and hierarchical RL.}
In this work,
we identify the horizon length as one of the main factors that inhibit the scaling of RL.
Prior works have developed diverse techniques to reduce the effective horizon
with multi-step or hierarchical value functions~\citep{feudal_dayan1992, option_sutton1999, option_stolle2002, oc_bacon2017, hiro_nachum2018, hac_levy2019, opal_ajay2021, archer_zhou2024},
hierarchical policy extraction~\citep{ril_gupta2019, play_lynch2019, hiql_park2023},
or high-level planning~\citep{sptm_savinov2018, sorb_eysenbach2019, mss_huang2019, leap_nasiriany2019, sfl_hoang2021, higl_kim2021, higoc_li2022, ptp_fang2022, pig_kim2023, hilp_park2024}.
While these works in hierarchical RL have mainly focused on
exploration~\citep{why_nachum2019}, representation learning~\citep{nearoptimal_nachum2019, hiql_park2023}, and planning~\citep{sorb_eysenbach2019},
we focus on \emph{scalability},
showing that horizon reduction mitigates bias accumulation
and unlocks the scaling of offline RL.
In this work, we also propose a new, minimal method (SHARSA) to reduce the horizon.
SHARSA is related to previous hierarchical methods that use rejection sampling for subgoal selection~\citep{iris_mandlekar2020, opal_ajay2021, ghilglue_hatch2025}.
Inspired by these works, SHARSA uses a minimal set of techniques (\eg, flow behavioral cloning and SARSA) that address the horizon issue in a scalable manner (see \Cref{sec:hrl_techniques} for further discussions).

\cutsectionup
\section{Experimental setup}
\cutsectiondown
\label{sec:setup}

\textbf{Problem setting.}
We aim to understand the degree to which current offline RL methods
can solve complex tasks simply by scaling data and compute.
In particular, we are interested in the capabilities of offline RL algorithms to solve challenging tasks
that require \emph{complex, long-hoziron} sequential decision-making given enough data.
To this end, we focus on the \emph{offline goal-conditioned RL} setting~\citep{ogbench_park2025},
where we want to train agents that are able to reach any goal state from any other initial state in the fewest number of steps,
from a static, pre-collected dataset of behaviors.
This problem poses a substantial learning challenge,
as the agent must learn complex, long-horizon, multi-task behaviors
purely from binary sparse rewards and an unlabeled (reward-free) dataset.
We note that although we mainly focus on goal-conditioned tasks in this work,
our claims are not limited to goal-conditioned settings (\Cref{sec:add_results}).

Formally, we consider a controlled Markov process defined as $\gM = (\gS, \gA, \mu, p)$,
where $\gS$ is the state space, $\gA$ is the action space,
$\mu(\pl{s}) \in \Delta(\gS)$ is the initial state distribution and $p(\pl{s'} \mid \pl{s}, \pl{a}): \gS \times \gA \to \Delta(\gS)$
is the transition dynamics kernel.
Here, $\Delta(\gX)$ denotes the set of probability distributions on space $\gX$,
and we denote placeholder variables in \pl{gray}.
We denote the discount factor as $\gamma$.
The dataset $\gD = \{\tau^{(n)}\}_{n \in \{1, 2, \ldots, N\}}$ consists of $N$ length-$H$ state-action trajectories,
$\tau = (s_0, a_0, s_1, a_1, \ldots, s_H)$.
We assume that these trajectories are collected in an unsupervised, task-agnostic manner.
\clearpage

\begin{wrapfigure}{r}{0.35\textwidth}
    \centering
    \vspace{11pt}
    \raisebox{0pt}[\dimexpr\height-1.0\baselineskip\relax]{
        \begin{subfigure}[t]{1.0\linewidth}
        \includegraphics[width=\linewidth]{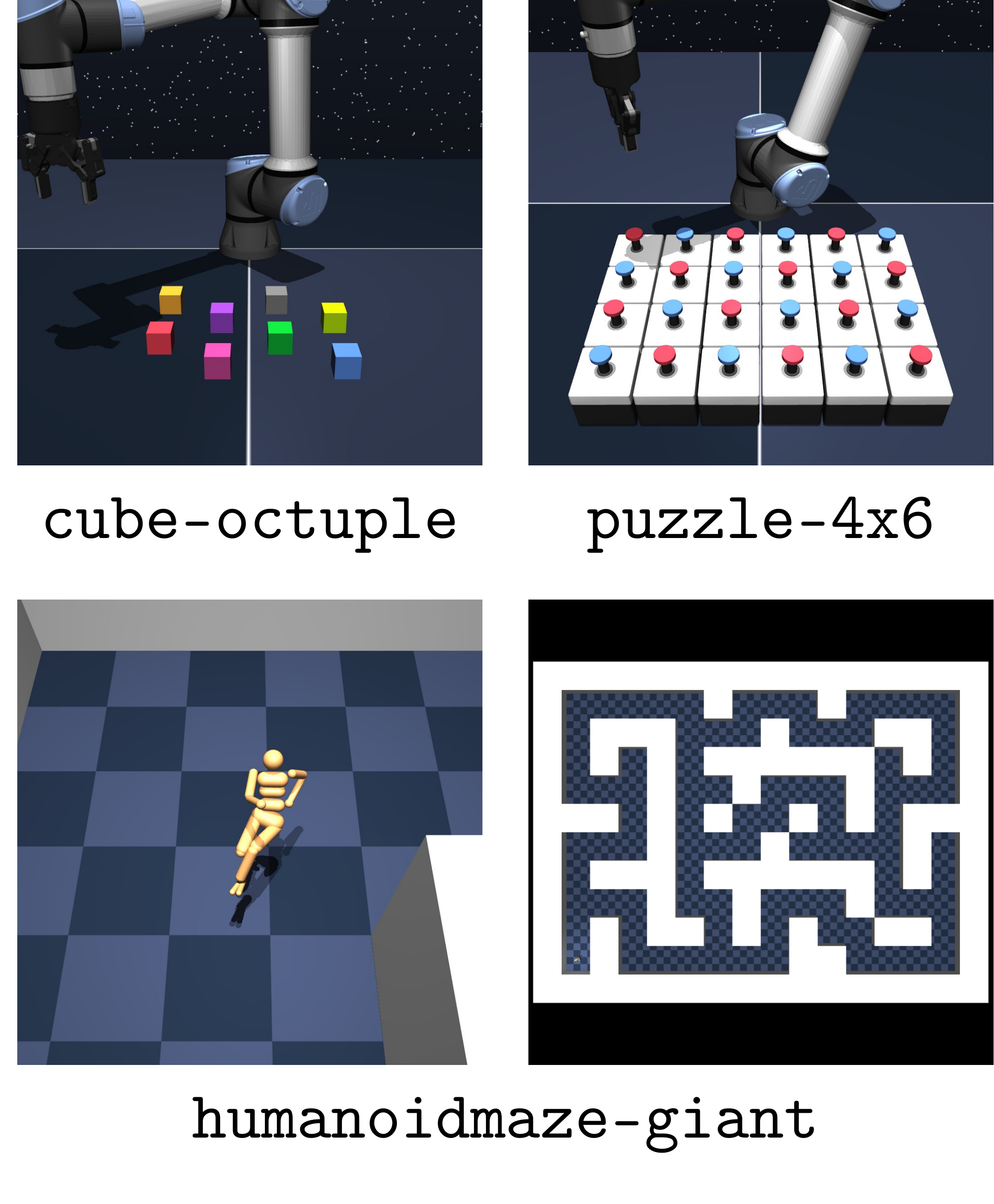}
        \end{subfigure}
    }
    \vspace{-35pt}
\end{wrapfigure}
\textbf{Environments and datasets.}
We employ four highly challenging offline goal-conditioned RL tasks in robotics
from the OGBench task suite~\citep{ogbench_park2025}. %
Among these tasks, \texttt{cube} involves sequential pick-and-place manipulation of multiple cube objects,
\texttt{puzzle} involves solving a combinatorial puzzle called ``Lights Out''\footnote{\url{https://en.wikipedia.org/wiki/Lights_Out_(game)}.}
with a robot arm,
and \texttt{humanoidmaze} involves whole-body control of a humanoid agent to navigate a given maze.
These OGBench tasks provide multiple variants with varying levels of difficulty,
and we employ the hardest tasks (\tt{cube-octuple}, \tt{puzzle-\{4x5, 4x6\}}, and \tt{humanoidmaze-giant})
to maximally challenge offline RL algorithms.
To our knowledge, no current offline RL algorithm has been reported to achieve non-trivial performance
(\ie, non-zero performance on most evaluation goals)
on these hardest tasks with the original OGBench datasets~\citep{ogbench_park2025}.
On these environments, we generate up to $\mathbf{1}$\textbf{B} transitions %
using the scripted policies provided by OGBench.
These datasets consist of ``play''-style~\citep{play_lynch2019} task-agnostic trajectories
to ensure sufficient coverage and diversity
(see also the discussion about dataset coverage in \Cref{sec:flat}).
Specifically, they contain trajectories that randomly navigate the maze (\tt{humanoidmaze}),
sequentially perform random pick-and-place (\tt{cube}),
or press random buttons (\tt{puzzle}).
These task-agnostic, unsupervised datasets conceptually resemble unlabeled Internet-scale data
used to train vision and language foundation models.
We also note that our $1$B-sized datasets contain about $1$M trajectories and $10$M atomic behaviors in manipulation environments,
which is similar or even larger than one of the largest robotics datasets to date~\citep{oxe_collaboration2024}.
\textbf{Idealization.}
To isolate the core sequential decision-making capabilities of RL from other confounding factors,
such as challenges with visual representation learning, distributional shift, and data coverage,
we idealize environments and tasks in our analysis experiments.
While these challenges are certainly important in practice,
our rationale is to first understand how current offline RL algorithms can solve highly challenging tasks
in an idealized, controlled setting with near-infinite data.

Specifically, we employ low-dimensional state-based observations with oracle goal representations
to alleviate challenges in visual representation learning.
We also remove some evaluation goals that may require out-of-distribution generalization 
to ensure all tasks remain in-distribution.
Finally, we ensure that the datasets have sufficient coverage and optimality,
by verifying that our data-collecting script enables achieving near-perfect performance
on the same environment with fewer objects (see \Cref{sec:flat} for the full discussion).
We refer to \Cref{sec:exp_details} for the details.

\textbf{Methods we evaluate.}
In this work, we mainly consider three performant, widely-used offline model-free RL algorithms across different categories:
IQL, CRL, and SAC+BC.
IQL~\citep{iql_kostrikov2022} is based on in-sample maximization,
CRL~\citep{crl_eysenbach2022} is based on contrastive learning and one-step RL~\citep{onestep_brandfonbrener2021},
and SAC+BC (\Cref{sec:algo_flat}) is based on behavioral regularization~\citep{brac_wu2019, td3bc_fujimoto2021}.
Additionally, we employ flow behavioral cloning (flow BC)~\citep{diffusionpolicy_chi2023, pi0_black2024}
to understand the scalability of behavioral cloning as well.
Due to high computational costs,
we use four random seeds in our scaling experiments (unless otherwise noted),
and report $95\%$ confidence intervals with shaded areas in the plots.
A full description and implementation details of the algorithms are provided in \Cref{sec:algo,sec:exp_details}.

\cutsectionup
\section{Standard offline RL methods struggle to scale}
\cutsectiondown
\label{sec:flat}

\begin{figure}[t!]
    \centering
    \vspace{-5pt}
    \includegraphics[width=1.0\textwidth]{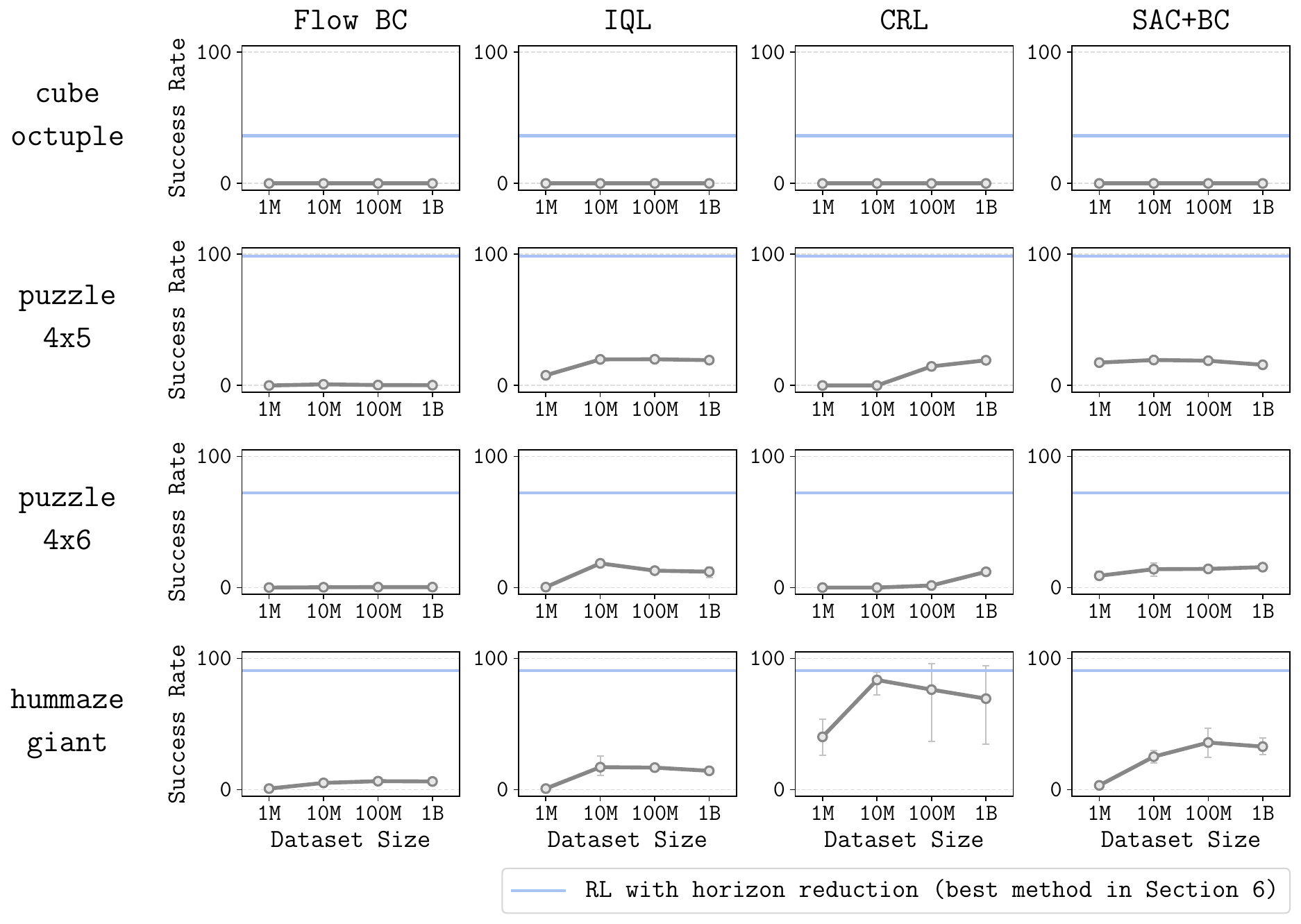}
    \vspace{-10pt}
    \caption{
    \footnotesize
    \textbf{Standard offline RL methods struggle to scale on challenging tasks.}
    We train four offline RL methods with $1$M, $10$M, $100$M, and $1$B data on four complex, long-horizon tasks.
    However, even with $1$B data, their performance often saturates
    far below the maximum performance ($100\%$).
    }
    \label{fig:flat_data}
\end{figure}

We now evaluate the degree to which four standard offline RL methods
(flow BC, IQL, CRL, and SAC+BC) can solve the four challenging tasks
by simply scaling up data.
\Cref{fig:flat_data} shows the scaling plots
of the four methods
with $1$M, $10$M, $100$M, and $1$B-sized datasets (see \Cref{fig:flat_main} for the full training curves).
These methods are trained for $5$M steps with $[1024, 1024, 1024, 1024]$-sized multi-layer perceptrons (MLPs).

In aggregate, our results show that
\textbf{none} of these four standard offline RL methods
are able to solve all four tasks,
even with the largest $1$B-sized datasets.
Notably, all of them completely fail on the hardest \texttt{cube-octuple} task.
Moreover, their performance often quickly plateaus well below the optimal success rate (\ie, $100\%$),
despite scaling up data.
In other words, these standard offline RL methods struggle to scale on these tasks.

A keen reader may already have several questions about this result.
Before proceeding further, we first address those potential questions.

\ul{\textbf{Q: How do you know these tasks are solvable with the given datasets?}}

\textbf{A:}
We will see in \Cref{sec:hrl} that
it is indeed possible to solve these tasks, or at least achieve non-trivial performance (denoted in {\color{myblue}blue} in \Cref{fig:flat_data}).
In \Cref{sec:flat_easy}, we also show that these algorithms can solve the same tasks with fewer objects,
using datasets collected by the same scripted policy.
This verifies that the dataset \emph{distribution} (induced by the scripted policy) provides sufficient coverage to learn a near-optimal policy.

\ul{\textbf{Q: Have you tried further increasing the model size?}}

\begin{figure}[t!]
    \centering
    \includegraphics[width=1.0\textwidth]{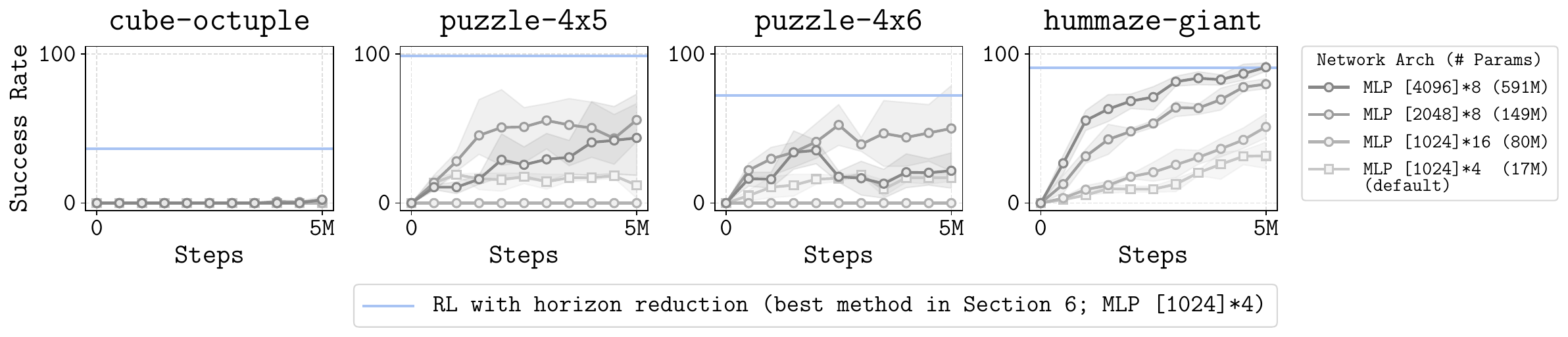}
    \vspace{-10pt}
    \caption{
    \footnotesize
    \textbf{Increasing model capacity alone is not sufficient to master the tasks.}
    }
    \label{fig:sacbc_size}
\end{figure}

\textbf{A:}
A natural confounder in the results above is the model size.
To understand how this affects performance, we train SAC+BC,
the best method on \tt{cube-double} and \tt{puzzle-4x4} (\Cref{fig:flat_easy}),
using up to $35\times$ larger models with $591$M parameters, on the largest $1$B datasets.
\Cref{fig:sacbc_size} shows the training curves.
The results suggest that while using larger networks can improve performance on some tasks to some degree,
this alone is not sufficient to master the tasks, especially the hardest \texttt{cube-octuple} task.
Moreover, the performance often saturates (or sometimes degrades) despite using larger models.
In \Cref{sec:flat_abl}, we provide more ablations with different architectures (residual MLPs and Transformers),
which show similar trends.

While an even larger network with a smaller learning rate\footnote{We already use a decreased learning rate for the largest $591$M model; see \Cref{sec:flat_abl} for the details.} might further improve performance
(which unfortunately we could not afford, as $591$M models already require $8$ days of training),
we are interested in performance within a reasonably bounded total compute budget.
If an algorithm is unable to achieve good performance within
a practical amount of compute,
we deem it a challenge in scalability.
In contrast, in \Cref{sec:hrl}, we will show that
horizon reduction techniques enable achieving significantly better
scaling behavior and asymptotic performance (denoted in {\color{myblue}blue} in the above figure)
even with the original $[1024] \times 4$-sized models.

\ul{\textbf{Q: Are you sure this isn't just a hyperparameter or design choice issue?}}

\textbf{A:}
While there is always a possibility of achieving better performance with better hyperparameters,
despite our extensive efforts in adjusting hyperparameters and design choices,
we were unable to achieve promising scaling results with these methods.
In \Cref{sec:flat_abl}, we present $\mathbf{9}$ ablation studies on
policy classes (Gaussian and flow policies),
network architectures (MLPs and Transformers),
value ensembles,
regularization techniques,
learning rates,
target network update rates,
batch sizes,
and gradient steps,
showing that \textbf{none} of these changes substantially improves scalability or asymptotic performance across the board.

\cutsectionup
\section{The curse of horizon}
\cutsectiondown
\label{sec:horizon}

Why do current offline RL methods exhibit poor scaling behavior on these challenging tasks?
In the previous section,
we observed that adjusting model sizes or other hyperparameters
does \emph{not} effectively improve scaling on complex, long-horizon tasks,
even though they scale on simpler tasks (see \Cref{fig:flat_easy}).
This suggests that there may exist a fundamental obstacle that inhibits the scaling of offline RL.
We hypothesize that this obstacle is the \textbf{horizon}.
In this section,
we discuss and analyze \emph{the curse of horizon} along two orthogonal axes: value and policy.

\cutsubsectionup
\subsection{The curse of horizon in value learning}
\cutsubsectiondown
\label{sec:horizon_value}

Many offline RL algorithms train Q functions via temporal difference (TD) learning.
Unfortunately, the TD learning objective has a fundamental limitation:
at any gradient step, the prediction target that the algorithm chases is \emph{biased}~\citep{rl_sutton2005},
and these biases \emph{accumulate} over the horizon. %
Such biases do not exist (or at least they do not accumulate)
in many scalable supervised and unsupervised learning objectives, such as next-token prediction.
As such, we hypothesize that the presence of bias accumulation in TD learning
is one of the fundamental causes behind the poor scaling result in \Cref{sec:flat}.
This hypothesis partly explains
why CRL, which is not based on TD learning, achieves a significantly better asymptotic performance on \tt{humanoidmaze-giant} in \Cref{fig:flat_data}.

\begin{figure}[h!]
    \centering
    \includegraphics[width=0.5\textwidth]{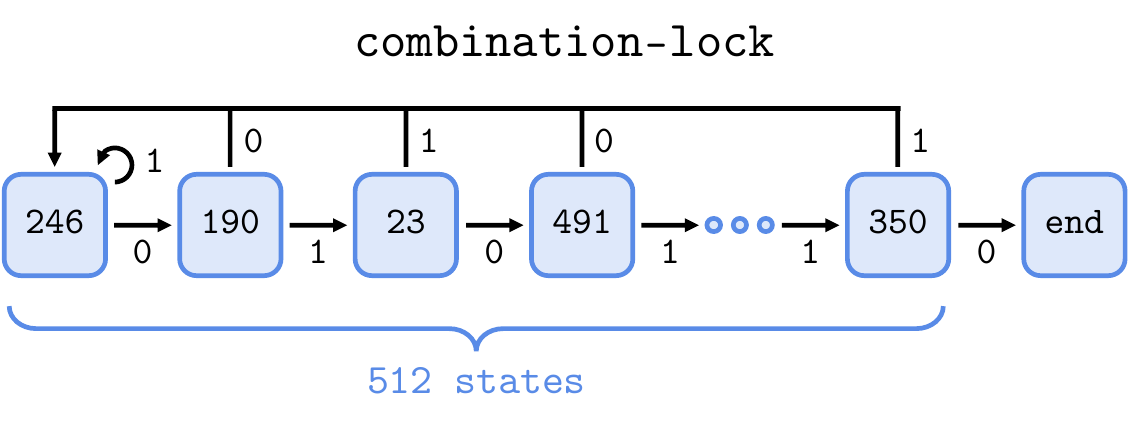}
    \vspace{-5pt}
    \caption{
    \footnotesize
    \textbf{The \tt{combination-lock} task with $\mathbf{H = 512}$ states.}
    }
    \label{fig:toy_env}
\end{figure}

\textbf{Didactic task setup.} We empirically validate this hypothesis
by performing an analysis on a didactic task
named \tt{combination-lock} (\Cref{fig:toy_env}).
This environment has $H$ states and two discrete actions.
The states are linearly ordered, and each state has an ``answer'' action.
The state order and answer actions are randomly chosen (and kept fixed) when instantiating the environment.
The agent starts from the first state,
and whenever it selects the correct action, it moves forward by one step;
otherwise, it is sent back to the first state.
The agent always gets a reward of $-1$ at each step,
except at the final (goal) state, where it gets a reward of $0$ and the episode terminates.
Hence, the agent must memorize all $H$ answer actions to reach the goal.
To understand the effect of bias accumulation in deep TD learning,
we evaluate two offline Q learning algorithms with different TD horizons:
standard ($1$-step) DQN~\citep{dqn_mnih2013} and $n$-step DQN (see \Cref{sec:exp_details_didactic} for details).%
\footnote{Although these are online RL algorithms, we can also use them for offline RL without modification, as our datasets have uniform coverage and thus do not require conservatism.}
Note that the optimal Q functions for both algorithms are the same
(under the optimal, full-coverage datasets) and thus have the same learning complexity,
but the latter involves $n$ times fewer TD recursions.
To compare the maximum possible performance of these two algorithms in a fair way,
we employ two types of datasets that have uniform coverage of length-$\{1, n\}$ trajectory segments,
evaluate each method on both datasets,
and select the best one for each method.
In this experiment, we do not use a discount factor (\ie, $\gamma = 1$) as the task has a finite horizon.
We refer the reader to \Cref{sec:exp_details_didactic} for the full experimental details.

\begin{figure}[h!]
    \centering
    \includegraphics[width=1.0\textwidth]{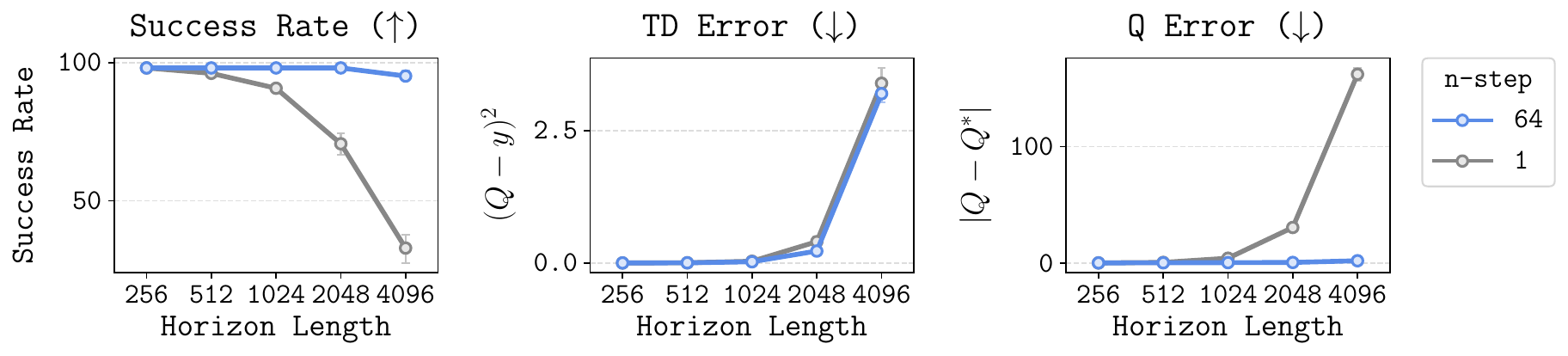}
    \vspace{-5pt}
    \caption{
    \footnotesize
    \textbf{$\mathbf{1}$-step TD learning suffers bias accumulation (\ie, high Q errors).}
    }
    \label{fig:toy_horizon}
\end{figure}

We train $1$-step and $64$-step DQN on \tt{combination-lock}
with different horizon lengths, ranging from $H=256$ to $H=4096$.
First, we measure their performance.
The first plot in \Cref{fig:toy_horizon} shows that
the performance of $1$-step DQN drops faster than that of $64$-step DQN
as the horizon increases.

\begin{wrapfigure}{r}{0.388\textwidth}
    \centering
    \vspace{-4pt}
    \raisebox{0pt}[\dimexpr\height-1.0\baselineskip\relax]{
        \begin{subfigure}[t]{1.0\linewidth}
        \includegraphics[width=\linewidth]{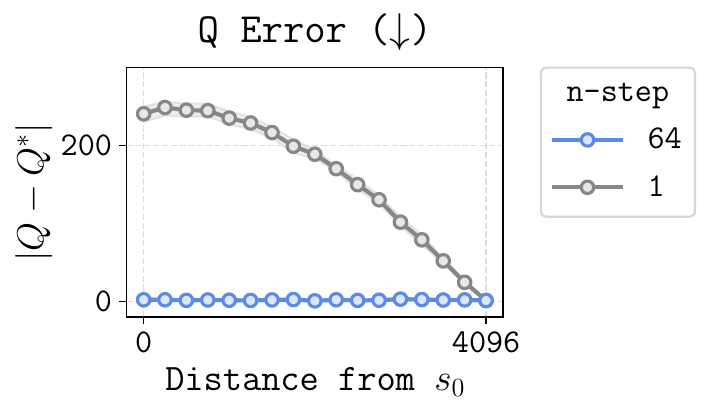}
        \end{subfigure}
    }
    \vspace{-15pt}
    \caption{
    \footnotesize
    \textbf{Biases accumulate.}
    }
    \vspace{-7pt}
    \label{fig:toy_accumulation}
\end{wrapfigure}
Next,
we measure two metrics: the TD error and the Q error.
The \emph{TD error} 
measures the difference against the TD target $y$,
and the \emph{Q error} measures against the ground-truth Q value $Q^*$.
The results are presented in
the second and third plots in \Cref{fig:toy_horizon}.
They show that
$1$-step DQN has significantly larger Q errors than $64$-step DQN,
even though they have similar TD errors.
Since the Q error corresponds to compounded error in the learned Q function,
this strongly suggests that bias accumulation happens in practice,
and that it can substantially affect performance on long-horizon tasks.
We further corroborate this point by measuring how Q errors vary across state positions in a single episode.
\Cref{fig:toy_accumulation} shows that Q errors indeed become larger as the distance from the end increases.

\begin{figure}[h!]
    \centering
    \vspace{-5pt}
    \includegraphics[width=1.0\textwidth]{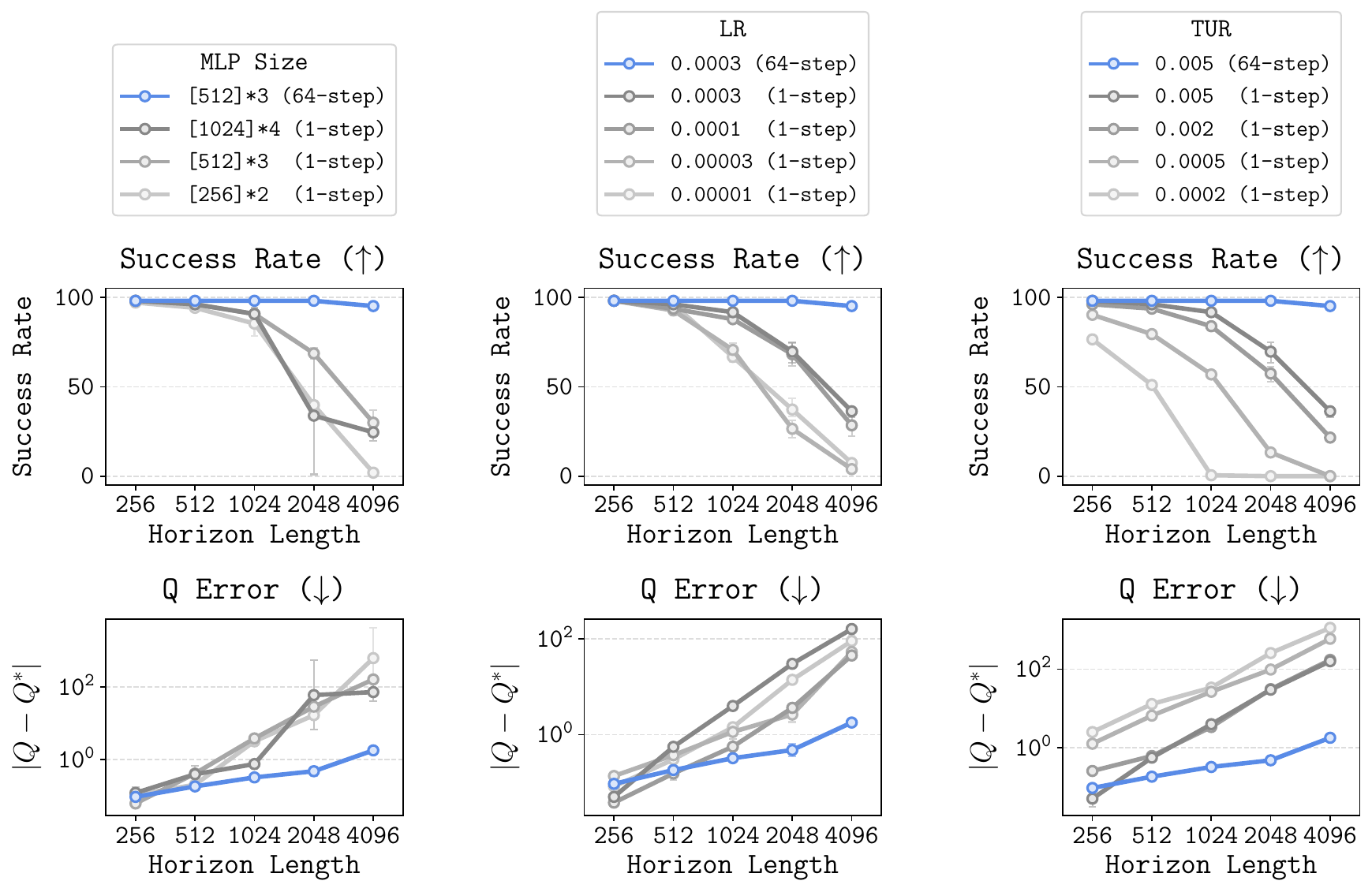}
    \vspace{-10pt}
    \caption{
    \footnotesize
    \textbf{Regardless of hyperparameters, $\mathbf{1}$-step TD learning struggles to handle a long horizon.}
    }
    \label{fig:toy_abl}
\end{figure}

Then, is it possible to fix error accumulation in $1$-step TD learning by tuning hyperparameters,
or is it a fundamental limitation of deep TD learning?
As in \Cref{sec:flat},
we adjust diverse hyperparameters,
such as the model size, learning rate (LR), and target network update rate (TUR),
and present the ablation results in \Cref{fig:toy_abl}.
The results suggest that
simply increasing the model size or decreasing LR or TUR
provides limited or no improvement in both performance and bias accumulation (measured by Q errors).
This matches the observation in \Cref{sec:flat}.
In contrast, $64$-step DQN achieves significantly better performance and Q errors,
even with the default-sized network.
This suggests that error accumulation over the horizon may be a fundamental factor
that obstructs scaling up TD learning.

Of course, there is a possibility that under certain hyperparameter settings,
such as with a very low learning rate and a much larger network,
$1$-step DQN might be able to converge to the optimal policy on long-horizon tasks.
While it is impossible to experimentally eliminate this possibility entirely,
our results do suggest $1$-step TD learning \emph{scales poorly} in horizon,
in the sense that it may require an excessive amount of compute, model capacity,
and the practitioner's time to achieve good performance.
In contrast, our results show that techniques that explicitly reduce the effective horizon,
such as $n$-step returns (as one example),
can potentially be much more effective in addressing this issue.
\cutsubsectionup
\subsection{The curse of horizon in policy learning}
\cutsubsectiondown
\label{sec:horizon_policy}

Orthogonal to the bias accumulation issue in value learning discussed in the previous section,
the policy may also independently suffer from the curse of horizon.
This is because, even when the value function is perfect,
the policy still needs to \emph{fit} the mapping between states and optimal actions
prescribed by the Q function,
where this mapping can be increasingly complex as the horizon becomes longer.
For example, in the goal-conditioned setting,
the mapping between the optimal actions and distant goals can be highly complex~\citep{ogbench_park2025},
as it depends on the entire topology of the state space.

Analogous to $n$-step returns in value learning,
we can mitigate the curse of horizon in policy learning
by reducing the effective horizon using a \emph{hierarchical} policy~\citep{ril_gupta2019, play_lynch2019, hiql_park2023, ogbench_park2025}.
For example, we can decompose a goal-conditioned policy $\pi(\pl{a} \mid \pl{s}, \pl{g})$ into
a high-level policy $\pi^h(\pl{w} \mid \pl{s}, \pl{g})$ that outputs a subgoal $w$,
and a low-level policy $\pi^\ell(\pl{a} \mid \pl{s}, \pl{w})$ that outputs actions given the subgoal.
Since the complexity  of the individual hierarchical policies is
often (much) lower than that of the flat (\ie, non-hierarchical) policy~\citep{hiql_park2023},
this can lead to a policy that both performs and \emph{generalizes} better~\citep{ogbench_park2025}.
This is akin to how chain-of-thought reasoning~\citep{cot_wei2022} improves the performance of language models,
which shows that decomposing a problem into multiple simpler subtasks is more effective than
producing an answer directly.
While we do not perform a separate didactic experiment for this point
(as it is relatively well studied and analyzed in prior work~\citep{ril_gupta2019, hiql_park2023, ogbench_park2025}),
we will empirically demonstrate how hierarchical policies can substantially improve the scalability of offline RL
in challenging environments in the next section.

\cutsectionup
\section{Horizon reduction makes RL \emph{scale} better}
\cutsectiondown
\label{sec:hrl}

Based on the insights in \Cref{sec:horizon},
we now apply value and policy horizon reduction techniques to our four challenging benchmark tasks,
and evaluate how they improve the scalability of offline RL.

\cutsubsectionup
\subsection{Horizon reduction techniques}
\cutsubsectiondown
\label{sec:hrl_techniques}

As discussed in the previous section,
there are two orthogonal axes of horizons in RL:
the value horizon (\Cref{sec:horizon_value}) and policy horizon (\Cref{sec:horizon_policy}).
In our experiments, we consider four representative techniques that reduce one or both types of horizons.
We refer to \Cref{sec:algo_hrl} for the full details.

\textbf{Value horizon reduction.}
For value horizon reduction, we consider \textbf{$\bm{n}$-step SAC+BC}, a variant of SAC+BC with $n$-step TD updates,
analogous to $n$-step DQN in \Cref{sec:horizon_value}.
This method reduces the value horizon, but not the policy horizon,
as it learns a flat policy.

\textbf{Policy horizon reduction.}
We consider two techniques that reduce the policy horizon, but not the value horizon.
\textbf{Hierarchical flow BC (hierarchical FBC)} trains a hierarchical policy ($\pi^h$ and $\pi^\ell$)
with flow behavioral cloning, without performing RL.
\textbf{HIQL}~\citep{hiql_park2023} trains a flat value function with goal-conditioned IQL,
but extract a hierarchical policy from it.
These methods will tell us the degree to which having a hierarchical policy \emph{alone}
can improve performance.

\textbf{Value \emph{and} policy horizon reduction.}
We can reduce both the value and policy horizons with full-fledged hierarchical RL.
While there are several approaches that perform full hierarchical offline RL
with a (potentially complex) high-level \emph{planner} (\Cref{sec:related}),
there exist only a handful of planning-free methods that reduce \emph{both} the value and policy horizons~\citep{iris_mandlekar2020, opal_ajay2021, ghilglue_hatch2025}.
Since these methods are either based on (less scalable) variational autoencoders or recurrent networks~\citep{iris_mandlekar2020, opal_ajay2021},
or only applicable to language-based tasks~\citep{ghilglue_hatch2025},
we propose a new method called \textbf{SHARSA} in the following section.

\cutsubsectionup
\subsection{SHARSA: a minimal, scalable offline RL method for horizon reduction}
\cutsubsectiondown
\label{sec:sharsa}

We propose a simple, scalable offline RL method that reduces both the value and policy horizons for continuous control.
Our main goal here is, rather than designing a completely novel technique that achieves state-of-the-art performance,
to empirically demonstrate how reducing both types of horizons improves scalability, even with otherwise simple ingredients.

The main challenge with full hierarchical offline RL (\ie, value and policy horizon reduction)
is \emph{high-level policy extraction}:
learning a subgoal policy $\pi^h(\pl{w} \mid \pl{s}, \pl{g})$ that maximizes values while not deviating too much from the data distribution.
For low-level or flat policies (whose output space is $\gA$),
policy extraction is typically best done by reparameterized gradients in the action space~\citep{bottleneck_park2024}
(\eg, DDPG+BC~\citep{td3bc_fujimoto2021, bottleneck_park2024}).
However, the same technique does not necessarily work for high-level policies (whose output space is $\gS$),
since first-order gradient information in the \emph{state} space may not be semantically meaningful
(\eg, the button states of \tt{puzzle} are discrete, and thus first-order gradients in the state space are not even well-defined).

To address this challenge,
we employ \emph{rejection sampling}~\citep{iris_mandlekar2020, sfbc_chen2023, idql_hansenestruch2023, ghilglue_hatch2025}
with an expressive \emph{flow} policy~\citep{flow_lipman2023, flow_liu2023, flow_albergo2023, pi0_black2024, fql_park2025}
for high-level policy extraction:
we first sample $N$ subgoals from a high-level flow BC policy $\pi_\beta^h$
and pick the best one based on a high-level (goal-conditioned) value function $Q^h$:
\begin{align}
    \pi^h(s, g) \stackrel{d}{=} \argmax_{w_1, \ldots, w_N: w_i \sim \pi_\beta^h(\pl{w} \mid s, g)} Q^h(s, w_i, g), \label{eq:sharsa}
\end{align}
where $\stackrel{d}{=}$ denotes equality in distribution.
This is beneficial because it does not use first-order information (unlike reparameterized gradients)
while leveraging the expressivity of a flow policy~\citep{bottleneck_park2024, fql_park2025}.
For the value function $Q^h$ in \Cref{eq:sharsa},
we employ high-level SARSA~\citep{rl_sutton2005} for simplicity,
which trains behavioral value functions with the following losses:
\begin{align}
    L^V(V^h) &= \E \Big[ D\big( V^h(s_h, g), \bar Q^h(s_h, s_{h+n}, g)\big)\Big], \\
    L^Q(Q^h) &= \E \Big[ D\big(Q^h(s_h, s_{h+n}, g), \sum_{i=0}^{n-1} \gamma^i r(s_{h+i}, g) + \gamma^n V^h(s_{h+n}, g) \big) \Big],
\end{align}
where $V^h$ is a high-level state value function,
$\bar Q^h$ is the target network~\citep{dqn_mnih2013},
$D$ is a loss function (regression or binary cross-entropy; we use the latter),
and the expectations are taken over length-$n$ trajectories $(s_h, a_h, \ldots, s_{h+n})$ and goals $g$ sampled from the dataset.
We refer to \Cref{sec:algo_sharsa} for the full details.
We note that one can use any \emph{decoupled} value learning methods
(\ie, those that do not involve policy learning, such as IQL~\citep{iql_kostrikov2022})
in place of SARSA.
For the low-level policy, we can either simply employ goal-conditioned flow BC,
or do another round of similar rejection sampling based on a low-level behavioral (SARSA) value function.
We call the former variant \textbf{SHARSA}%
\footnote{This acronym stands for (state)--(high-level action)--(reward)--(state)--(high-level action).}
and the latter variant \textbf{double SHARSA}.
We provide the pseudocode for SHARSA and double SHARSA in \Cref{alg:sharsa,alg:double_sharsa}.

SHARSA is appealing for two reasons.
First, it is simple and easy to use.
SHARSA is only based on behavioral cloning and SARSA,
both of which do not require extensive hyperparameter tuning,
unlike typical offline RL algorithms~\citep{corl_tarasov2023, bottleneck_park2024}.
Second, it reduces both the value and policy horizon lengths with an expressive flow policy.
This mitigates the curse of horizon in a scalable way.

\begin{algorithm}[t!]
\caption{SHARSA}
\label{alg:sharsa}
\begin{algorithmic}
\footnotesize

\LComment{\color{myblue} Training loop}
\While{not converged}
\State Sample batch $\{(s_h, a_h, \ldots, s_{h+n}, g)\}$ from $\gD$
\BeginBox[fill=white]
\LComment{Hierarchical flow BC}
\State Update high-level flow BC policy $\pi_\beta^h(\pl{s_{h+n}} \mid \pl{s_h}, \pl{g})$ with flow-matching loss (\Cref{eq:sharsa_pi_high})
\State Update low-level flow BC policy $\pi_\beta^\ell(\pl{a_h} \mid \pl{s_h}, \pl{s_{h+n}})$ with flow-matching loss (\Cref{eq:sharsa_pi_low})
\EndBox
\BeginBox[fill=white]
\LComment{High-level ($n$-step) SARSA value learning}
\State Update $V^h$ to minimize $\E\left[D\left(V^h(s_h, g), \bar Q^h(s_h, s_{h+n}, g)\right)\right]$
\State Update $Q^h$ to minimize $\E\left[D\left(Q^h(s_h, s_{h+n}, g), \sum_{i=0}^{n-1} \gamma^i r(s_{h+i}, g) + \gamma^n V^h(s_{h+n}, g)\right)\right]$
\EndBox
\EndWhile
\Return $\pi(\pl{s}, \pl{g})$ defined below

\vspace{5pt}

\LComment{\color{myblue} Resulting policy}
\Function{$\pi(s, g)$}{}
\BeginBox[fill=white]
\LComment{High-level: rejection sampling}
\State Sample $w_1, \ldots, w_N \sim \pi_\beta^h(s, g)$
\State Set $w \gets \argmax_{w_1, \ldots, w_N} Q^h(s, w_i, g)$
\EndBox
\BeginBox[fill=white]
\LComment{Low-level: behavioral cloning} 
\State Sample $a \sim \pi_\beta^\ell(s, w)$\vskip 30pt
\EndBox
\State \Return $a$
\EndFunction

\end{algorithmic}
\end{algorithm}

\cutsubsectionup
\subsection{Results}
\cutsubsectiondown

\begin{figure}[h!]
    \centering
    \vspace{-5pt}
    \includegraphics[width=1.0\textwidth]{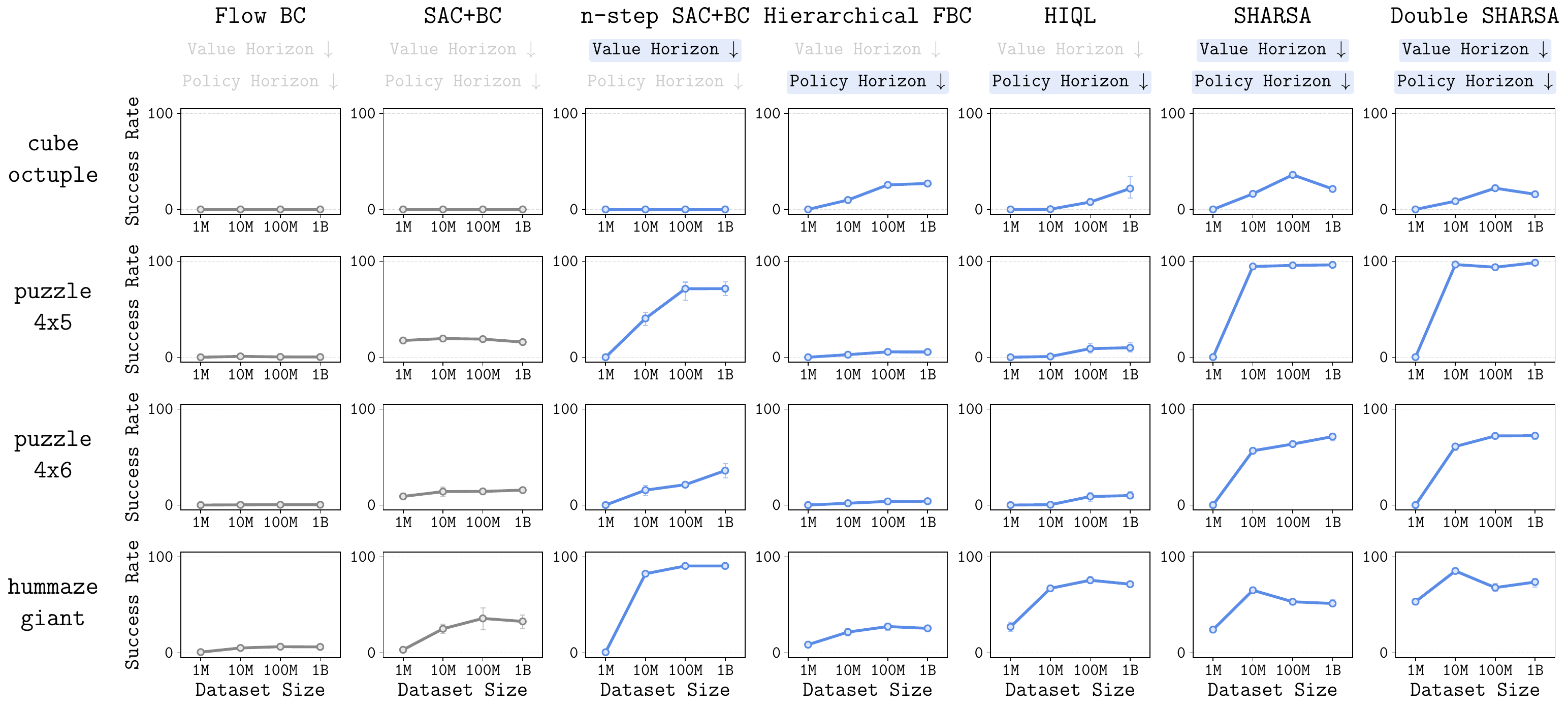}
    \vspace{-10pt}
    \caption{
    \footnotesize
    \textbf{Horizon reduction makes RL scalable.}
    Value and policy horizon reduction techniques often result in substantially better scaling and asymptotic performance.
    }
    \label{fig:hrl_data}
\end{figure}

We now evaluate the performance 
of various horizon reduction techniques
on the main benchmark tasks.
We present the data-scaling curves in \Cref{fig:hrl_data} (see \Cref{fig:hrl_main} for the training curves).
The results show that horizon reduction techniques can indeed unlock the scalability of offline RL on these challenging tasks.
We highlight three particularly informative comparisons:

\textbf{Value horizon reduction: SAC+BC vs. $\bm{n}$-step SAC+BC}
shows that simply reducing the \emph{value} horizon with $n$-step returns ($n$-step SAC+BC)
substantially improves scalability and even \emph{asymptotic} performance on many tasks.
This matches our didactic experiments in \Cref{sec:horizon_value}.
We note that their network sizes and training objectives are identical, except for the use of $n$-step returns.

\textbf{Policy horizon reduction: Flow BC vs. hierarchical flow BC}
shows that reducing the \emph{policy} horizon also significantly improves performance, but on a different set of tasks.
In particular, it shows that, on some tasks (\eg, \tt{cube-octuple}),
it is challenging to achieve even non-zero performance without reducing the policy horizon.

\textbf{Value \emph{and} policy horizon reduction: SHARSA vs. the others}
shows that reducing \emph{both} value and policy horizons
leads to the best of both worlds.
In particular, (double) SHARSA is the \emph{only} method that achieves non-trivial performance on all four tasks in our experiments.
In \Cref{sec:sharsa_abl}, we present several ablation studies on SHARSA, discussing the relative importance of various design choices
(\eg, alternative policy extraction strategies and value learning objectives).

\cutsectionup
\section{Call for research: offline RL algorithms should be evaluated for \emph{scalability}}
\label{sec:call}
\cutsectiondown

In this work, we empirically showed that standard, non-hierarchical offline RL methods struggle to scale on complex tasks.
We hypothesized that this is due to the curse of horizon,
and demonstrated that techniques that explicitly reduce the horizon length, including SHARSA,
can unlock scalability.

However, this is far from the end of the story.
Empirically, still none of these techniques enable \emph{mastering}
all four tasks (\ie, achieving a $100\%$ performance), even with $1$B data.
Methodologically, these hierarchical methods only \emph{mitigate}
the error accumulation issue in TD learning with two-level hierarchies,
rather than fundamentally solving it. 
Moreover, SHARSA and other $n$-step return-based methods implicitly assume that
dataset trajectories are near-optimal within short segments
(although double SHARSA relaxes this assumption to some extent).
Finally, our results still indicate room for improvement over SHARSA, as in some cases the performance does not always scale monotonically with increasing dataset sizes (\Cref{fig:hrl_data}).
These limitations of current approaches open up a number of fruitful research questions in scalable reinforcement learning:
\begin{itemize}
\item Can we completely avoid TD learning while performing RL
(\eg, potentially with model-based RL~\citep{dreamerv3_hafner2025}, linear programming~\citep{mdp_puterman2014, qrl_wang2023}, or shortest path algorithms~\citep{gcrl_kaelbling1993, fwrl_dhiman2018})? 
\item Can we find a \emph{simple}, scalable way to extend beyond two-level hierarchies to deal with horizons of arbitrary length?
\item Is the curse of horizon fundamentally impossible to solve? The RL theory community suggests otherwise~\citep{mvp_zhang2021, mvp_zhang2024},
and can we instantiate such a principle within deep RL?
\end{itemize}

We conclude this paper by calling for research on \emph{scalable} offline RL algorithms,
that is, algorithmic research done on large-scale datasets and complex tasks.
Currently, offline RL research is often mainly conducted on standard datasets (\eg, D4RL~\citep{d4rl_fu2020}, OGBench~\citep{ogbench_park2025}, etc.)
with $1$M--$5$M transitions.
However, success on small-scale tasks and datasets does not necessarily guarantee success
on datasets that are $1000\times$ larger,
as not every algorithm \emph{scales} equally~\citep{bitter_sutton2019, scaling_tay2023}.
Hence, to assess their potential at scale,
it is important to directly evaluate
new methods on substantially more challenging tasks and larger datasets
and measure scaling trends.
To facilitate this, we open-source our tasks, datasets, and implementations (\href{https://github.com/seohongpark/horizon-reduction}{link}),
where we have made them as easy to use as possible.
We hope that our insights in this work, as well as our open-source implementation,
serve as a foundation for the development of scalable offline RL objectives
that unlock the full potential of data-driven RL.

\begin{ack}
We thank Oleh Rybkin for helpful discussions.
This work was partly supported by the Korea Foundation for Advanced Studies (KFAS),
National Science Foundation Graduate Research Fellowship Program under Grant No. DGE 2146752,
ONR N00014-22-1-2773,
and NSF IIS-2150826.
This research used the Savio computational cluster resource provided by the Berkeley Research Computing program at UC Berkeley.
\end{ack}

{\footnotesize
\bibliographystyle{plainnat}
\bibliography{neurips_2025}

\begin{thebibliography}{106}
\providecommand{\natexlab}[1]{#1}
\providecommand{\url}[1]{\texttt{#1}}
\expandafter\ifx\csname urlstyle\endcsname\relax
  \providecommand{\doi}[1]{doi: #1}\else
  \providecommand{\doi}{doi: \begingroup \urlstyle{rm}\Url}\fi

\bibitem[Ajay et~al.(2021)Ajay, Kumar, Agrawal, Levine, and
  Nachum]{opal_ajay2021}
Anurag Ajay, Aviral Kumar, Pulkit Agrawal, Sergey Levine, and Ofir Nachum.
\newblock Opal: Offline primitive discovery for accelerating offline
  reinforcement learning.
\newblock In \emph{International Conference on Learning Representations
  (ICLR)}, 2021.

\bibitem[Albergo and Vanden-Eijnden(2023)]{flow_albergo2023}
Michael~S Albergo and Eric Vanden-Eijnden.
\newblock Building normalizing flows with stochastic interpolants.
\newblock In \emph{International Conference on Learning Representations
  (ICLR)}, 2023.

\bibitem[An et~al.(2021)An, Moon, Kim, and Song]{edac_an2021}
Gaon An, Seungyong Moon, Jang-Hyun Kim, and Hyun~Oh Song.
\newblock Uncertainty-based offline reinforcement learning with diversified
  q-ensemble.
\newblock In \emph{Neural Information Processing Systems (NeurIPS)}, 2021.

\bibitem[Ba et~al.(2016)Ba, Kiros, and Hinton]{ln_ba2016}
Jimmy Ba, Jamie~Ryan Kiros, and Geoffrey~E. Hinton.
\newblock Layer normalization.
\newblock \emph{ArXiv}, abs/1607.06450, 2016.

\bibitem[Bacon et~al.(2017)Bacon, Harb, and Precup]{oc_bacon2017}
Pierre-Luc Bacon, Jean Harb, and Doina Precup.
\newblock The option-critic architecture.
\newblock In \emph{AAAI Conference on Artificial Intelligence (AAAI)}, 2017.

\bibitem[Black et~al.(2024)Black, Brown, Driess, Esmail, Equi, Finn, Fusai,
  Groom, Hausman, Ichter, et~al.]{pi0_black2024}
Kevin Black, Noah Brown, Danny Driess, Adnan Esmail, Michael Equi, Chelsea
  Finn, Niccolo Fusai, Lachy Groom, Karol Hausman, Brian Ichter, et~al.
\newblock $\pi_0$: A vision-language-action flow model for general robot
  control.
\newblock \emph{ArXiv}, abs/2410.24164, 2024.

\bibitem[Brandfonbrener et~al.(2021)Brandfonbrener, Whitney, Ranganath, and
  Bruna]{onestep_brandfonbrener2021}
David Brandfonbrener, William~F. Whitney, Rajesh Ranganath, and Joan Bruna.
\newblock Offline rl without off-policy evaluation.
\newblock In \emph{Neural Information Processing Systems (NeurIPS)}, 2021.

\bibitem[Chebotar et~al.(2023)Chebotar, Vuong, Irpan, Hausman, Xia, Lu, Kumar,
  Yu, Herzog, Pertsch, Gopalakrishnan, Ibarz, Nachum, Sontakke, Salazar, Tran,
  Peralta, Tan, Manjunath, Singht, Zitkovich, Jackson, Rao, Finn, and
  Levine]{qt_chebotar2023}
Yevgen Chebotar, Quan~Ho Vuong, Alex Irpan, Karol Hausman, F.~Xia, Yao Lu,
  Aviral Kumar, Tianhe Yu, Alexander Herzog, Karl Pertsch, Keerthana
  Gopalakrishnan, Julian Ibarz, Ofir Nachum, Sumedh~Anand Sontakke, Grecia
  Salazar, Huong Tran, Jodilyn Peralta, Clayton Tan, Deeksha Manjunath, Jaspiar
  Singht, Brianna Zitkovich, Tomas Jackson, Kanishka Rao, Chelsea Finn, and
  Sergey Levine.
\newblock Q-transformer: Scalable offline reinforcement learning via
  autoregressive q-functions.
\newblock In \emph{Conference on Robot Learning (CoRL)}, 2023.

\bibitem[Chen et~al.(2023)Chen, Lu, Ying, Su, and Zhu]{sfbc_chen2023}
Huayu Chen, Cheng Lu, Chengyang Ying, Hang Su, and Jun Zhu.
\newblock Offline reinforcement learning via high-fidelity generative behavior
  modeling.
\newblock In \emph{International Conference on Learning Representations
  (ICLR)}, 2023.

\bibitem[Chen et~al.(2021)Chen, Lu, Rajeswaran, Lee, Grover, Laskin, Abbeel,
  Srinivas, and Mordatch]{dt_chen2021}
Lili Chen, Kevin Lu, Aravind Rajeswaran, Kimin Lee, Aditya Grover, Michael
  Laskin, P.~Abbeel, A.~Srinivas, and Igor Mordatch.
\newblock Decision transformer: Reinforcement learning via sequence modeling.
\newblock In \emph{Neural Information Processing Systems (NeurIPS)}, 2021.

\bibitem[Cheng et~al.(2025)Cheng, Qiao, Xiong, Miao, Ma, Li, Li, and
  Lv]{jowa_cheng2025}
Jie Cheng, Ruixi Qiao, Gang Xiong, Qinghai Miao, Yingwei Ma, Binhua Li, Yongbin
  Li, and Yisheng Lv.
\newblock Scaling offline model-based rl via jointly-optimized world-action
  model pretraining.
\newblock In \emph{International Conference on Learning Representations
  (ICLR)}, 2025.

\bibitem[Chi et~al.(2023)Chi, Xu, Feng, Cousineau, Du, Burchfiel, Tedrake, and
  Song]{diffusionpolicy_chi2023}
Cheng Chi, Zhenjia Xu, Siyuan Feng, Eric Cousineau, Yilun Du, Benjamin
  Burchfiel, Russ Tedrake, and Shuran Song.
\newblock Diffusion policy: Visuomotor policy learning via action diffusion.
\newblock In \emph{Robotics: Science and Systems (RSS)}, 2023.

\bibitem[Collaboration et~al.(2024)Collaboration, O’Neill, Rehman, Maddukuri,
  Gupta, Padalkar, Lee, Pooley, Gupta, Mandlekar, Jain,
  et~al.]{oxe_collaboration2024}
Open X-Embodiment Collaboration, Abby O’Neill, Abdul Rehman, Abhiram
  Maddukuri, Abhishek Gupta, Abhishek Padalkar, Abraham Lee, Acorn Pooley,
  Agrim Gupta, Ajay Mandlekar, Ajinkya Jain, et~al.
\newblock Open x-embodiment: Robotic learning datasets and rt-x models.
\newblock In \emph{IEEE International Conference on Robotics and Automation
  (ICRA)}, 2024.

\bibitem[Dayan and Hinton(1992)]{feudal_dayan1992}
Peter Dayan and Geoffrey~E. Hinton.
\newblock Feudal reinforcement learning.
\newblock In \emph{Neural Information Processing Systems (NeurIPS)}, 1992.

\bibitem[Dhiman et~al.(2018)Dhiman, Banerjee, Siskind, and
  Corso]{fwrl_dhiman2018}
Vikas Dhiman, Shurjo Banerjee, Jeffrey~M Siskind, and Jason~J Corso.
\newblock Floyd-warshall reinforcement learning: Learning from past experiences
  to reach new goals.
\newblock \emph{ArXiv}, abs/1809.09318, 2018.

\bibitem[Espeholt et~al.(2018)Espeholt, Soyer, Munos, Simonyan, Mnih, Ward,
  Doron, Firoiu, Harley, Dunning, Legg, and Kavukcuoglu]{impala_espeholt2018}
Lasse Espeholt, Hubert Soyer, R{\'e}mi Munos, Karen Simonyan, Volodymyr Mnih,
  Tom Ward, Yotam Doron, Vlad Firoiu, Tim Harley, Iain Dunning, Shane Legg, and
  Koray Kavukcuoglu.
\newblock Impala: Scalable distributed deep-rl with importance weighted
  actor-learner architectures.
\newblock In \emph{International Conference on Machine Learning (ICML)}, 2018.

\bibitem[Eysenbach et~al.(2019)Eysenbach, Salakhutdinov, and
  Levine]{sorb_eysenbach2019}
Benjamin Eysenbach, Ruslan Salakhutdinov, and Sergey Levine.
\newblock Search on the replay buffer: Bridging planning and reinforcement
  learning.
\newblock In \emph{Neural Information Processing Systems (NeurIPS)}, 2019.

\bibitem[Eysenbach et~al.(2022)Eysenbach, Zhang, Salakhutdinov, and
  Levine]{crl_eysenbach2022}
Benjamin Eysenbach, Tianjun Zhang, Ruslan Salakhutdinov, and Sergey Levine.
\newblock Contrastive learning as goal-conditioned reinforcement learning.
\newblock In \emph{Neural Information Processing Systems (NeurIPS)}, 2022.

\bibitem[Fang et~al.(2022)Fang, Yin, Nair, and Levine]{ptp_fang2022}
Kuan Fang, Patrick Yin, Ashvin Nair, and Sergey Levine.
\newblock Planning to practice: Efficient online fine-tuning by composing goals
  in latent space.
\newblock In \emph{IEEE/RSJ International Conference on Intelligent Robots and
  Systems (IROS)}, 2022.

\bibitem[Farebrother et~al.(2024)Farebrother, Orbay, Vuong, Ta{\"\i}ga,
  Chebotar, Xiao, Irpan, Levine, Castro, Faust,
  et~al.]{hlgauss_farebrother2024}
Jesse Farebrother, Jordi Orbay, Quan Vuong, Adrien~Ali Ta{\"\i}ga, Yevgen
  Chebotar, Ted Xiao, Alex Irpan, Sergey Levine, Pablo~Samuel Castro,
  Aleksandra Faust, et~al.
\newblock Stop regressing: Training value functions via classification for
  scalable deep rl.
\newblock In \emph{International Conference on Machine Learning (ICML)}, 2024.

\bibitem[Fu et~al.(2020)Fu, Kumar, Nachum, Tucker, and Levine]{d4rl_fu2020}
Justin Fu, Aviral Kumar, Ofir Nachum, G.~Tucker, and Sergey Levine.
\newblock D4rl: Datasets for deep data-driven reinforcement learning.
\newblock \emph{ArXiv}, abs/2004.07219, 2020.

\bibitem[Fujimoto and Gu(2021)]{td3bc_fujimoto2021}
Scott Fujimoto and Shixiang~Shane Gu.
\newblock A minimalist approach to offline reinforcement learning.
\newblock In \emph{Neural Information Processing Systems (NeurIPS)}, 2021.

\bibitem[Fujimoto et~al.(2018)Fujimoto, van Hoof, and Meger]{td3_fujimoto2018}
Scott Fujimoto, Herke van Hoof, and David Meger.
\newblock Addressing function approximation error in actor-critic methods.
\newblock In \emph{International Conference on Machine Learning (ICML)}, 2018.

\bibitem[Gallici et~al.(2025)Gallici, Fellows, Ellis, Pou, Masmitja, Foerster,
  and Martin]{pqn_gallici2025}
Matteo Gallici, Mattie Fellows, Benjamin Ellis, Bartomeu Pou, Ivan Masmitja,
  Jakob~Nicolaus Foerster, and Mario Martin.
\newblock Simplifying deep temporal difference learning.
\newblock In \emph{International Conference on Learning Representations
  (ICLR)}, 2025.

\bibitem[Garg et~al.(2023)Garg, Hejna, Geist, and Ermon]{xql_garg2023}
Divyansh Garg, Joey Hejna, Matthieu Geist, and Stefano Ermon.
\newblock Extreme q-learning: Maxent rl without entropy.
\newblock In \emph{International Conference on Learning Representations
  (ICLR)}, 2023.

\bibitem[Gupta et~al.(2019)Gupta, Kumar, Lynch, Levine, and
  Hausman]{ril_gupta2019}
Abhishek Gupta, Vikash Kumar, Corey Lynch, Sergey Levine, and Karol Hausman.
\newblock Relay policy learning: Solving long-horizon tasks via imitation and
  reinforcement learning.
\newblock In \emph{Conference on Robot Learning (CoRL)}, 2019.

\bibitem[Haarnoja et~al.(2018{\natexlab{a}})Haarnoja, Zhou, Abbeel, and
  Levine]{sac_haarnoja2018}
Tuomas Haarnoja, Aurick Zhou, Pieter Abbeel, and Sergey Levine.
\newblock Soft actor-critic: Off-policy maximum entropy deep reinforcement
  learning with a stochastic actor.
\newblock In \emph{International Conference on Machine Learning (ICML)},
  2018{\natexlab{a}}.

\bibitem[Haarnoja et~al.(2018{\natexlab{b}})Haarnoja, Zhou, Hartikainen,
  Tucker, Ha, Tan, Kumar, Zhu, Gupta, Abbeel, and Levine]{saces_haarnoja2018}
Tuomas Haarnoja, Aurick Zhou, Kristian Hartikainen, G.~Tucker, Sehoon Ha, Jie
  Tan, Vikash Kumar, Henry Zhu, Abhishek Gupta, Pieter Abbeel, and Sergey
  Levine.
\newblock Soft actor-critic algorithms and applications.
\newblock \emph{ArXiv}, abs/1812.05905, 2018{\natexlab{b}}.

\bibitem[Hafner et~al.(2025)Hafner, Pasukonis, Ba, and
  Lillicrap]{dreamerv3_hafner2025}
Danijar Hafner, Jurgis Pasukonis, Jimmy Ba, and Timothy Lillicrap.
\newblock Mastering diverse control tasks through world models.
\newblock \emph{Nature}, 640:\penalty0 647--653, 2025.

\bibitem[Hansen et~al.(2024)Hansen, Su, and Wang]{tdmpc2_hansen2024}
Nicklas Hansen, Hao Su, and Xiaolong Wang.
\newblock Td-mpc2: Scalable, robust world models for continuous control.
\newblock In \emph{International Conference on Learning Representations
  (ICLR)}, 2024.

\bibitem[Hansen-Estruch et~al.(2023)Hansen-Estruch, Kostrikov, Janner, Kuba,
  and Levine]{idql_hansenestruch2023}
Philippe Hansen-Estruch, Ilya Kostrikov, Michael Janner, Jakub~Grudzien Kuba,
  and Sergey Levine.
\newblock Idql: Implicit q-learning as an actor-critic method with diffusion
  policies.
\newblock \emph{ArXiv}, abs/2304.10573, 2023.

\bibitem[Hasselt et~al.(2016)Hasselt, Guez, and Silver]{ddqn_hasselt2016}
H.~V. Hasselt, Arthur Guez, and David Silver.
\newblock Deep reinforcement learning with double q-learning.
\newblock In \emph{AAAI Conference on Artificial Intelligence (AAAI)}, 2016.

\bibitem[Hatch et~al.(2025)Hatch, Balakrishna, Mees, Nair, Park, Wulfe, Itkina,
  Eysenbach, Levine, Kollar, and Burchfiel]{ghilglue_hatch2025}
Kyle~B. Hatch, Ashwin Balakrishna, Oier Mees, Suraj Nair, Seohong Park, Blake
  Wulfe, Masha Itkina, Benjamin Eysenbach, Sergey Levine, Thomas Kollar, and
  Benjamin Burchfiel.
\newblock Ghil-glue: Hierarchical control with filtered subgoal images.
\newblock In \emph{IEEE International Conference on Robotics and Automation
  (ICRA)}, 2025.

\bibitem[Hendrycks and Gimpel(2016)]{gelu_hendrycks2016}
Dan Hendrycks and Kevin Gimpel.
\newblock Gaussian error linear units (gelus).
\newblock \emph{ArXiv}, abs/1606.08415, 2016.

\bibitem[Hoang et~al.(2021)Hoang, Sohn, Choi, Carvalho, and Lee]{sfl_hoang2021}
Christopher Hoang, Sungryull Sohn, Jongwook Choi, Wilka Carvalho, and Honglak
  Lee.
\newblock Successor feature landmarks for long-horizon goal-conditioned
  reinforcement learning.
\newblock In \emph{Neural Information Processing Systems (NeurIPS)}, 2021.

\bibitem[Huang et~al.(2019)Huang, Liu, and Su]{mss_huang2019}
Zhiao Huang, Fangchen Liu, and Hao Su.
\newblock Mapping state space using landmarks for universal goal reaching.
\newblock In \emph{Neural Information Processing Systems (NeurIPS)}, 2019.

\bibitem[Imani and White(2018)]{hlgauss_imani2018}
Ehsan Imani and Martha White.
\newblock Improving regression performance with distributional losses.
\newblock In \emph{International Conference on Machine Learning (ICML)}, 2018.

\bibitem[Jaech et~al.(2024)Jaech, Kalai, Lerer, Richardson, El-Kishky, Low,
  Helyar, Madry, Beutel, Carney, et~al.]{openaio1_jaech2024}
Aaron Jaech, Adam Kalai, Adam Lerer, Adam Richardson, Ahmed El-Kishky, Aiden
  Low, Alec Helyar, Aleksander Madry, Alex Beutel, Alex Carney, et~al.
\newblock Openai o1 system card.
\newblock \emph{ArXiv}, abs/2412.16720, 2024.

\bibitem[Janner et~al.(2021)Janner, Li, and Levine]{tt_janner2021}
Michael Janner, Qiyang Li, and Sergey Levine.
\newblock Reinforcement learning as one big sequence modeling problem.
\newblock In \emph{Neural Information Processing Systems (NeurIPS)}, 2021.

\bibitem[Janner et~al.(2022)Janner, Du, Tenenbaum, and
  Levine]{diffuser_janner2022}
Michael Janner, Yilun Du, Joshua~B. Tenenbaum, and Sergey Levine.
\newblock Planning with diffusion for flexible behavior synthesis.
\newblock In \emph{International Conference on Machine Learning (ICML)}, 2022.

\bibitem[Kaelbling(1993)]{gcrl_kaelbling1993}
Leslie~Pack Kaelbling.
\newblock Learning to achieve goals.
\newblock In \emph{International Joint Conference on Artificial Intelligence
  (IJCAI)}, 1993.

\bibitem[Kalashnikov et~al.(2018)Kalashnikov, Irpan, Pastor, Ibarz, Herzog,
  Jang, Quillen, Holly, Kalakrishnan, Vanhoucke, and
  Levine]{qtopt_kalashnikov2018}
Dmitry Kalashnikov, Alex Irpan, Peter Pastor, Julian Ibarz, Alexander Herzog,
  Eric Jang, Deirdre Quillen, Ethan Holly, Mrinal Kalakrishnan, Vincent
  Vanhoucke, and Sergey Levine.
\newblock Qt-opt: Scalable deep reinforcement learning for vision-based robotic
  manipulation.
\newblock In \emph{Conference on Robot Learning (CoRL)}, 2018.

\bibitem[Kidambi et~al.(2020)Kidambi, Rajeswaran, Netrapalli, and
  Joachims]{morel_kidambi2020}
Rahul Kidambi, Aravind Rajeswaran, Praneeth Netrapalli, and Thorsten Joachims.
\newblock Morel : Model-based offline reinforcement learning.
\newblock In \emph{Neural Information Processing Systems (NeurIPS)}, 2020.

\bibitem[Kim et~al.(2021)Kim, Seo, and Shin]{higl_kim2021}
Junsu Kim, Younggyo Seo, and Jinwoo Shin.
\newblock Landmark-guided subgoal generation in hierarchical reinforcement
  learning.
\newblock In \emph{Neural Information Processing Systems (NeurIPS)}, 2021.

\bibitem[Kim et~al.(2023)Kim, Seo, Ahn, Son, and Shin]{pig_kim2023}
Junsu Kim, Younggyo Seo, Sungsoo Ahn, Kyunghwan Son, and Jinwoo Shin.
\newblock Imitating graph-based planning with goal-conditioned policies.
\newblock In \emph{International Conference on Learning Representations
  (ICLR)}, 2023.

\bibitem[Kingma and Ba(2015)]{adam_kingma2015}
Diederik~P. Kingma and Jimmy Ba.
\newblock Adam: A method for stochastic optimization.
\newblock In \emph{International Conference on Learning Representations
  (ICLR)}, 2015.

\bibitem[Kostrikov et~al.(2022)Kostrikov, Nair, and Levine]{iql_kostrikov2022}
Ilya Kostrikov, Ashvin Nair, and Sergey Levine.
\newblock Offline reinforcement learning with implicit q-learning.
\newblock In \emph{International Conference on Learning Representations
  (ICLR)}, 2022.

\bibitem[Kumar et~al.(2020)Kumar, Zhou, Tucker, and Levine]{cql_kumar2020}
Aviral Kumar, Aurick Zhou, G.~Tucker, and Sergey Levine.
\newblock Conservative q-learning for offline reinforcement learning.
\newblock In \emph{Neural Information Processing Systems (NeurIPS)}, 2020.

\bibitem[Kumar et~al.(2023)Kumar, Agarwal, Geng, Tucker, and
  Levine]{scaledql_kumar2023}
Aviral Kumar, Rishabh Agarwal, Xinyang Geng, George Tucker, and Sergey Levine.
\newblock Offline q-learning on diverse multi-task data both scales and
  generalizes.
\newblock In \emph{International Conference on Learning Representations
  (ICLR)}, 2023.

\bibitem[Laidlaw et~al.(2023)Laidlaw, Russell, and Dragan]{bridge_laidlaw2023}
Cassidy Laidlaw, Stuart~J. Russell, and Anca~D. Dragan.
\newblock Bridging rl theory and practice with the effective horizon.
\newblock In \emph{Neural Information Processing Systems (NeurIPS)}, 2023.

\bibitem[Lee et~al.(2025)Lee, Hwang, Kim, Kim, Tai, Subramanian, Wurman, Choo,
  Stone, and Seno]{simba_lee2025}
Hojoon Lee, Dongyoon Hwang, Donghu Kim, Hyunseung Kim, Jun~Jet Tai, Kaushik
  Subramanian, Peter~R Wurman, Jaegul Choo, Peter Stone, and Takuma Seno.
\newblock Simba: Simplicity bias for scaling up parameters in deep
  reinforcement learning.
\newblock In \emph{International Conference on Learning Representations
  (ICLR)}, 2025.

\bibitem[Lee(2012)]{smooth_lee2002}
John~M Lee.
\newblock \emph{Introduction to Smooth Manifolds}.
\newblock Springer, 2012.

\bibitem[Lee et~al.(2021)Lee, Jeon, Lee, Pineau, and Kim]{optidice_lee2021}
Jongmin Lee, Wonseok Jeon, Byung-Jun Lee, Jo{\"e}lle Pineau, and Kee-Eung Kim.
\newblock Optidice: Offline policy optimization via stationary distribution
  correction estimation.
\newblock In \emph{International Conference on Machine Learning (ICML)}, 2021.

\bibitem[Lee et~al.(2022)Lee, Nachum, Yang, Lee, Freeman, Xu, Guadarrama,
  Fischer, Jang, Michalewski, and Mordatch]{mgdt_lee2022}
Kuang-Huei Lee, Ofir Nachum, Mengjiao Yang, L.~Y. Lee, Daniel Freeman, Winnie
  Xu, Sergio Guadarrama, Ian~S. Fischer, Eric Jang, Henryk Michalewski, and
  Igor Mordatch.
\newblock Multi-game decision transformers.
\newblock In \emph{Neural Information Processing Systems (NeurIPS)}, 2022.

\bibitem[Levine et~al.(2020)Levine, Kumar, Tucker, and Fu]{offline_levine2020}
Sergey Levine, Aviral Kumar, G.~Tucker, and Justin Fu.
\newblock Offline reinforcement learning: Tutorial, review, and perspectives on
  open problems.
\newblock \emph{ArXiv}, abs/2005.01643, 2020.

\bibitem[Levy et~al.(2019)Levy, Konidaris, Platt, and Saenko]{hac_levy2019}
Andrew Levy, George~Dimitri Konidaris, Robert~W. Platt, and Kate Saenko.
\newblock Learning multi-level hierarchies with hindsight.
\newblock In \emph{International Conference on Learning Representations
  (ICLR)}, 2019.

\bibitem[Li et~al.(2022)Li, Tang, Tomizuka, and Zhan]{higoc_li2022}
Jinning Li, Chen Tang, Masayoshi Tomizuka, and Wei Zhan.
\newblock Hierarchical planning through goal-conditioned offline reinforcement
  learning.
\newblock \emph{IEEE Robotics and Automation Letters (RA-L)}, 7\penalty0
  (4):\penalty0 10216--10223, 2022.

\bibitem[Li et~al.(2023)Li, Chen, Hong, Ajay, and Agrawal]{pql_li2023}
Zechu Li, Tao Chen, Zhang-Wei Hong, Anurag Ajay, and Pulkit Agrawal.
\newblock Parallel q-learning: Scaling off-policy reinforcement learning under
  massively parallel simulation.
\newblock In \emph{International Conference on Machine Learning (ICML)}, 2023.

\bibitem[Lipman et~al.(2023)Lipman, Chen, Ben-Hamu, Nickel, and
  Le]{flow_lipman2023}
Yaron Lipman, Ricky~TQ Chen, Heli Ben-Hamu, Maximilian Nickel, and Matt Le.
\newblock Flow matching for generative modeling.
\newblock In \emph{International Conference on Learning Representations
  (ICLR)}, 2023.

\bibitem[Lipman et~al.(2024)Lipman, Havasi, Holderrieth, Shaul, Le, Karrer,
  Chen, Lopez-Paz, Ben-Hamu, and Gat]{flow_lipman2024}
Yaron Lipman, Marton Havasi, Peter Holderrieth, Neta Shaul, Matt Le, Brian
  Karrer, Ricky T.~Q. Chen, David Lopez-Paz, Heli Ben-Hamu, and Itai Gat.
\newblock Flow matching guide and code.
\newblock \emph{ArXiv}, abs/2412.06264, 2024.

\bibitem[Liu et~al.(2023)Liu, Gong, and Liu]{flow_liu2023}
Xingchao Liu, Chengyue Gong, and Qiang Liu.
\newblock Flow straight and fast: Learning to generate and transfer data with
  rectified flow.
\newblock In \emph{International Conference on Learning Representations
  (ICLR)}, 2023.

\bibitem[Lynch et~al.(2019)Lynch, Khansari, Xiao, Kumar, Tompson, Levine, and
  Sermanet]{play_lynch2019}
Corey Lynch, Mohi Khansari, Ted Xiao, Vikash Kumar, Jonathan Tompson, Sergey
  Levine, and Pierre Sermanet.
\newblock Learning latent plans from play.
\newblock In \emph{Conference on Robot Learning (CoRL)}, 2019.

\bibitem[Ma and Collins(2018)]{nce_ma2018}
Zhuang Ma and Michael Collins.
\newblock Noise contrastive estimation and negative sampling for conditional
  models: Consistency and statistical efficiency.
\newblock In \emph{Conference on Empirical Methods in Natural Language
  Processing (EMNLP)}, 2018.

\bibitem[Mandlekar et~al.(2020)Mandlekar, Ramos, Boots, Fei-Fei, Garg, and
  Fox]{iris_mandlekar2020}
Ajay Mandlekar, Fabio Ramos, Byron Boots, Li~Fei-Fei, Animesh Garg, and Dieter
  Fox.
\newblock Iris: Implicit reinforcement without interaction at scale for
  learning control from offline robot manipulation data.
\newblock In \emph{IEEE International Conference on Robotics and Automation
  (ICRA)}, 2020.

\bibitem[Mnih et~al.(2013)Mnih, Kavukcuoglu, Silver, Graves, Antonoglou,
  Wierstra, and Riedmiller]{dqn_mnih2013}
Volodymyr Mnih, Koray Kavukcuoglu, David Silver, Alex Graves, Ioannis
  Antonoglou, Daan Wierstra, and Martin~A. Riedmiller.
\newblock Playing atari with deep reinforcement learning.
\newblock \emph{ArXiv}, abs/1312.5602, 2013.

\bibitem[Nachum et~al.(2018)Nachum, Gu, Lee, and Levine]{hiro_nachum2018}
Ofir Nachum, Shixiang~Shane Gu, Honglak Lee, and Sergey Levine.
\newblock Data-efficient hierarchical reinforcement learning.
\newblock In \emph{Neural Information Processing Systems (NeurIPS)}, 2018.

\bibitem[Nachum et~al.(2019{\natexlab{a}})Nachum, Gu, Lee, and
  Levine]{nearoptimal_nachum2019}
Ofir Nachum, Shixiang~Shane Gu, Honglak Lee, and Sergey Levine.
\newblock Near-optimal representation learning for hierarchical reinforcement
  learning.
\newblock In \emph{International Conference on Learning Representations
  (ICLR)}, 2019{\natexlab{a}}.

\bibitem[Nachum et~al.(2019{\natexlab{b}})Nachum, Tang, Lu, Gu, Lee, and
  Levine]{why_nachum2019}
Ofir Nachum, Haoran Tang, Xingyu Lu, Shixiang~Shane Gu, Honglak Lee, and Sergey
  Levine.
\newblock Why does hierarchy (sometimes) work so well in reinforcement
  learning?
\newblock \emph{ArXiv}, abs/1909.10618, 2019{\natexlab{b}}.

\bibitem[Nasiriany et~al.(2019)Nasiriany, Pong, Lin, and
  Levine]{leap_nasiriany2019}
Soroush Nasiriany, Vitchyr~H. Pong, Steven Lin, and Sergey Levine.
\newblock Planning with goal-conditioned policies.
\newblock In \emph{Neural Information Processing Systems (NeurIPS)}, 2019.

\bibitem[Nauman et~al.(2024)Nauman, Ostaszewski, Jankowski, Mi{\l}o{\'s}, and
  Cygan]{bro_nauman2024}
Michal Nauman, Mateusz Ostaszewski, Krzysztof Jankowski, Piotr Mi{\l}o{\'s},
  and Marek Cygan.
\newblock Bigger, regularized, optimistic: scaling for compute and
  sample-efficient continuous control.
\newblock In \emph{Neural Information Processing Systems (NeurIPS)}, 2024.

\bibitem[Nikulin et~al.(2023)Nikulin, Kurenkov, Tarasov, and
  Kolesnikov]{sacrnd_nikulin2023}
Alexander Nikulin, Vladislav Kurenkov, Denis Tarasov, and Sergey Kolesnikov.
\newblock Anti-exploration by random network distillation.
\newblock In \emph{International Conference on Machine Learning (ICML)}, 2023.

\bibitem[Park et~al.(2023)Park, Ghosh, Eysenbach, and Levine]{hiql_park2023}
Seohong Park, Dibya Ghosh, Benjamin Eysenbach, and Sergey Levine.
\newblock Hiql: Offline goal-conditioned rl with latent states as actions.
\newblock In \emph{Neural Information Processing Systems (NeurIPS)}, 2023.

\bibitem[Park et~al.(2024{\natexlab{a}})Park, Frans, Levine, and
  Kumar]{bottleneck_park2024}
Seohong Park, Kevin Frans, Sergey Levine, and Aviral Kumar.
\newblock Is value learning really the main bottleneck in offline rl?
\newblock In \emph{Neural Information Processing Systems (NeurIPS)},
  2024{\natexlab{a}}.

\bibitem[Park et~al.(2024{\natexlab{b}})Park, Kreiman, and
  Levine]{hilp_park2024}
Seohong Park, Tobias Kreiman, and Sergey Levine.
\newblock Foundation policies with hilbert representations.
\newblock In \emph{International Conference on Machine Learning (ICML)},
  2024{\natexlab{b}}.

\bibitem[Park et~al.(2025{\natexlab{a}})Park, Frans, Eysenbach, and
  Levine]{ogbench_park2025}
Seohong Park, Kevin Frans, Benjamin Eysenbach, and Sergey Levine.
\newblock Ogbench: Benchmarking offline goal-conditioned rl.
\newblock In \emph{International Conference on Learning Representations
  (ICLR)}, 2025{\natexlab{a}}.

\bibitem[Park et~al.(2025{\natexlab{b}})Park, Li, and Levine]{fql_park2025}
Seohong Park, Qiyang Li, and Sergey Levine.
\newblock Flow q-learning.
\newblock In \emph{International Conference on Machine Learning (ICML)},
  2025{\natexlab{b}}.

\bibitem[Peng et~al.(2019)Peng, Kumar, Zhang, and Levine]{awr_peng2019}
Xue~Bin Peng, Aviral Kumar, Grace Zhang, and Sergey Levine.
\newblock Advantage-weighted regression: Simple and scalable off-policy
  reinforcement learning.
\newblock \emph{ArXiv}, abs/1910.00177, 2019.

\bibitem[Peters and Schaal(2007)]{rwr_peters2007}
Jan Peters and Stefan Schaal.
\newblock Reinforcement learning by reward-weighted regression for operational
  space control.
\newblock In \emph{International Conference on Machine Learning (ICML)}, 2007.

\bibitem[Puterman(2014)]{mdp_puterman2014}
Martin~L Puterman.
\newblock \emph{Markov decision processes: discrete stochastic dynamic
  programming}.
\newblock John Wiley \& Sons, 2014.

\bibitem[Reed et~al.(2022)Reed, Zolna, Parisotto, Colmenarejo, Novikov,
  Barth-maron, Gim{\'e}nez, Sulsky, Kay, Springenberg, et~al.]{gato_reed2022}
Scott Reed, Konrad Zolna, Emilio Parisotto, Sergio~G{\'o}mez Colmenarejo,
  Alexander Novikov, Gabriel Barth-maron, Mai Gim{\'e}nez, Yury Sulsky, Jackie
  Kay, Jost~Tobias Springenberg, et~al.
\newblock A generalist agent.
\newblock \emph{Transactions on Machine Learning Research (TMLR)}, 2022.

\bibitem[Rybkin et~al.(2025)Rybkin, Nauman, Fu, Snell, Abbeel, Levine, and
  Kumar]{rlscale_rybkin2025}
Oleh Rybkin, Michal Nauman, Preston Fu, Charlie Snell, Pieter Abbeel, Sergey
  Levine, and Aviral Kumar.
\newblock Value-based deep rl scales predictably.
\newblock In \emph{International Conference on Machine Learning (ICML)}, 2025.

\bibitem[Savinov et~al.(2018)Savinov, Dosovitskiy, and
  Koltun]{sptm_savinov2018}
Nikolay Savinov, Alexey Dosovitskiy, and Vladlen Koltun.
\newblock Semi-parametric topological memory for navigation.
\newblock In \emph{International Conference on Learning Representations
  (ICLR)}, 2018.

\bibitem[Schweitzer and Seidmann(1985)]{residual_schweitzer1985}
Paul~J Schweitzer and Abraham Seidmann.
\newblock Generalized polynomial approximations in markovian decision
  processes.
\newblock \emph{Journal of mathematical analysis and applications},
  110\penalty0 (2):\penalty0 568--582, 1985.

\bibitem[Sikchi et~al.(2024)Sikchi, Zheng, Zhang, and
  Niekum]{dualrl_sikchi2024}
Harshit~S. Sikchi, Qinqing Zheng, Amy Zhang, and Scott Niekum.
\newblock Dual rl: Unification and new methods for reinforcement and imitation
  learning.
\newblock In \emph{International Conference on Learning Representations
  (ICLR)}, 2024.

\bibitem[Singla et~al.(2024)Singla, Agarwal, and Pathak]{sapg_singla2024}
Jayesh Singla, Ananye Agarwal, and Deepak Pathak.
\newblock Sapg: Split and aggregate policy gradients.
\newblock In \emph{International Conference on Machine Learning (ICML)}, 2024.

\bibitem[Springenberg et~al.(2024)Springenberg, Abdolmaleki, Zhang, Groth,
  Bloesch, Lampe, Brakel, Bechtle, Kapturowski, Hafner, Heess, and
  Riedmiller]{pac_springenberg2024}
Jost~Tobias Springenberg, Abbas Abdolmaleki, Jingwei Zhang, Oliver Groth,
  Michael Bloesch, Thomas Lampe, Philemon Brakel, Sarah Bechtle, Steven
  Kapturowski, Roland Hafner, Nicolas Manfred~Otto Heess, and Martin~A.
  Riedmiller.
\newblock Offline actor-critic reinforcement learning scales to large models.
\newblock In \emph{International Conference on Machine Learning (ICML)}, 2024.

\bibitem[Stolle and Precup(2002)]{option_stolle2002}
Martin Stolle and Doina Precup.
\newblock Learning options in reinforcement learning.
\newblock In \emph{Symposium on Abstraction, Reformulation and Approximation},
  2002.

\bibitem[Sutton(2019)]{bitter_sutton2019}
Richard Sutton.
\newblock The bitter lesson, 2019.
\newblock URL \url{http://www.incompleteideas.net/IncIdeas/BitterLesson.html}.

\bibitem[Sutton and Barto(2005)]{rl_sutton2005}
Richard~S. Sutton and Andrew~G. Barto.
\newblock Reinforcement learning: An introduction.
\newblock \emph{IEEE Transactions on Neural Networks}, 16:\penalty0 285--286,
  2005.

\bibitem[Sutton et~al.(1999)Sutton, Precup, and Singh]{option_sutton1999}
Richard~S Sutton, Doina Precup, and Satinder Singh.
\newblock Between mdps and semi-mdps: A framework for temporal abstraction in
  reinforcement learning.
\newblock \emph{Artificial intelligence}, 112\penalty0 (1-2):\penalty0
  181--211, 1999.

\bibitem[Tarasov et~al.(2023{\natexlab{a}})Tarasov, Kurenkov, Nikulin, and
  Kolesnikov]{rebrac_tarasov2023}
Denis Tarasov, Vladislav Kurenkov, Alexander Nikulin, and Sergey Kolesnikov.
\newblock Revisiting the minimalist approach to offline reinforcement learning.
\newblock In \emph{Neural Information Processing Systems (NeurIPS)},
  2023{\natexlab{a}}.

\bibitem[Tarasov et~al.(2023{\natexlab{b}})Tarasov, Nikulin, Akimov, Kurenkov,
  and Kolesnikov]{corl_tarasov2023}
Denis Tarasov, Alexander Nikulin, Dmitry Akimov, Vladislav Kurenkov, and Sergey
  Kolesnikov.
\newblock Corl: Research-oriented deep offline reinforcement learning library.
\newblock In \emph{Neural Information Processing Systems (NeurIPS)},
  2023{\natexlab{b}}.

\bibitem[Tay et~al.(2023)Tay, Dehghani, Abnar, Chung, Fedus, Rao, Narang, Tran,
  Yogatama, and Metzler]{scaling_tay2023}
Yi~Tay, Mostafa Dehghani, Samira Abnar, Hyung~Won Chung, William Fedus, Jinfeng
  Rao, Sharan Narang, Vinh~Q Tran, Dani Yogatama, and Donald Metzler.
\newblock Scaling laws vs model architectures: How does inductive bias
  influence scaling?
\newblock In \emph{Findings of the Association for Computational Linguistics:
  EMNLP 2023}, 2023.

\bibitem[Team et~al.(2025)Team, Du, Gao, Xing, Jiang, Chen, Li, Xiao, Du, Liao,
  et~al.]{kimi_team2025}
Kimi Team, Angang Du, Bofei Gao, Bowei Xing, Changjiu Jiang, Cheng Chen, Cheng
  Li, Chenjun Xiao, Chenzhuang Du, Chonghua Liao, et~al.
\newblock Kimi k1. 5: Scaling reinforcement learning with llms.
\newblock \emph{ArXiv}, abs/2501.12599, 2025.

\bibitem[Vaswani et~al.(2017)Vaswani, Shazeer, Parmar, Uszkoreit, Jones, Gomez,
  Kaiser, and Polosukhin]{transformer_vaswani2017}
Ashish Vaswani, Noam~M. Shazeer, Niki Parmar, Jakob Uszkoreit, Llion Jones,
  Aidan~N. Gomez, Lukasz Kaiser, and Illia Polosukhin.
\newblock Attention is all you need.
\newblock In \emph{Neural Information Processing Systems (NeurIPS)}, 2017.

\bibitem[Wang et~al.(2023)Wang, Torralba, Isola, and Zhang]{qrl_wang2023}
Tongzhou Wang, Antonio Torralba, Phillip Isola, and Amy Zhang.
\newblock Optimal goal-reaching reinforcement learning via quasimetric
  learning.
\newblock In \emph{International Conference on Machine Learning (ICML)}, 2023.

\bibitem[Wang et~al.(2020)Wang, Novikov, Zolna, Springenberg, Reed, Shahriari,
  Siegel, Merel, Gulcehre, Heess, and de~Freitas]{crr_wang2020}
Ziyun Wang, Alexander Novikov, Konrad Zolna, Jost~Tobias Springenberg, Scott~E.
  Reed, Bobak Shahriari, Noah Siegel, Josh Merel, Caglar Gulcehre, Nicolas
  Manfred~Otto Heess, and Nando de~Freitas.
\newblock Critic regularized regression.
\newblock In \emph{Neural Information Processing Systems (NeurIPS)}, 2020.

\bibitem[Wei et~al.(2022)Wei, Wang, Schuurmans, Bosma, Xia, Chi, Le, Zhou,
  et~al.]{cot_wei2022}
Jason Wei, Xuezhi Wang, Dale Schuurmans, Maarten Bosma, Fei Xia, Ed~Chi, Quoc~V
  Le, Denny Zhou, et~al.
\newblock Chain-of-thought prompting elicits reasoning in large language
  models.
\newblock In \emph{Neural Information Processing Systems (NeurIPS)}, 2022.

\bibitem[Wu et~al.(2019)Wu, Tucker, and Nachum]{brac_wu2019}
Yifan Wu, G.~Tucker, and Ofir Nachum.
\newblock Behavior regularized offline reinforcement learning.
\newblock \emph{ArXiv}, abs/1911.11361, 2019.

\bibitem[Xu et~al.(2023)Xu, Jiang, Li, Yang, Wang, Chan, and Zhan]{sql_xu2023}
Haoran Xu, Li~Jiang, Jianxiong Li, Zhuoran Yang, Zhaoran Wang, Victor Chan, and
  Xianyuan Zhan.
\newblock Offline rl with no ood actions: In-sample learning via implicit value
  regularization.
\newblock In \emph{International Conference on Learning Representations
  (ICLR)}, 2023.

\bibitem[Yang et~al.(2021)Yang, Hu, Babuschkin, Sidor, Liu, Farhi, Ryder,
  Pachocki, Chen, and Gao]{mup_yang2021}
Greg Yang, Edward~J Hu, Igor Babuschkin, Szymon Sidor, Xiaodong Liu, David
  Farhi, Nick Ryder, Jakub Pachocki, Weizhu Chen, and Jianfeng Gao.
\newblock Tensor programs v: Tuning large neural networks via zero-shot
  hyperparameter transfer.
\newblock In \emph{Neural Information Processing Systems (NeurIPS)}, 2021.

\bibitem[Yu et~al.(2020)Yu, Thomas, Yu, Ermon, Zou, Levine, Finn, and
  Ma]{mopo_yu2020}
Tianhe Yu, Garrett Thomas, Lantao Yu, Stefano Ermon, James~Y. Zou, Sergey
  Levine, Chelsea Finn, and Tengyu Ma.
\newblock Mopo: Model-based offline policy optimization.
\newblock In \emph{Neural Information Processing Systems (NeurIPS)}, 2020.

\bibitem[Yu et~al.(2021)Yu, Kumar, Rafailov, Rajeswaran, Levine, and
  Finn]{combo_yu2021}
Tianhe Yu, Aviral Kumar, Rafael Rafailov, Aravind Rajeswaran, Sergey Levine,
  and Chelsea Finn.
\newblock Combo: Conservative offline model-based policy optimization.
\newblock In \emph{Neural Information Processing Systems (NeurIPS)}, 2021.

\bibitem[Zhang et~al.(2021)Zhang, Ji, and Du]{mvp_zhang2021}
Zihan Zhang, Xiangyang Ji, and Simon Du.
\newblock Is reinforcement learning more difficult than bandits? a near-optimal
  algorithm escaping the curse of horizon.
\newblock In \emph{Conference on Learning Theory (COLT)}, 2021.

\bibitem[Zhang et~al.(2024)Zhang, Chen, Lee, and Du]{mvp_zhang2024}
Zihan Zhang, Yuxin Chen, Jason~D Lee, and Simon~S Du.
\newblock Settling the sample complexity of online reinforcement learning.
\newblock In \emph{Conference on Learning Theory (COLT)}, 2024.

\bibitem[Zhou et~al.(2024)Zhou, Zanette, Pan, Levine, and
  Kumar]{archer_zhou2024}
Yifei Zhou, Andrea Zanette, Jiayi Pan, Sergey Levine, and Aviral Kumar.
\newblock Archer: Training language model agents via hierarchical multi-turn
  rl.
\newblock In \emph{International Conference on Machine Learning (ICML)}, 2024.

\end{thebibliography}
}

\clearpage
\appendix

\section{Offline RL scales well on short-horizon tasks}
\label{sec:flat_easy}

\begin{figure}[h!]
    \centering
    \begin{subfigure}[t]{1.0\textwidth}
    \includegraphics[width=1.0\textwidth]{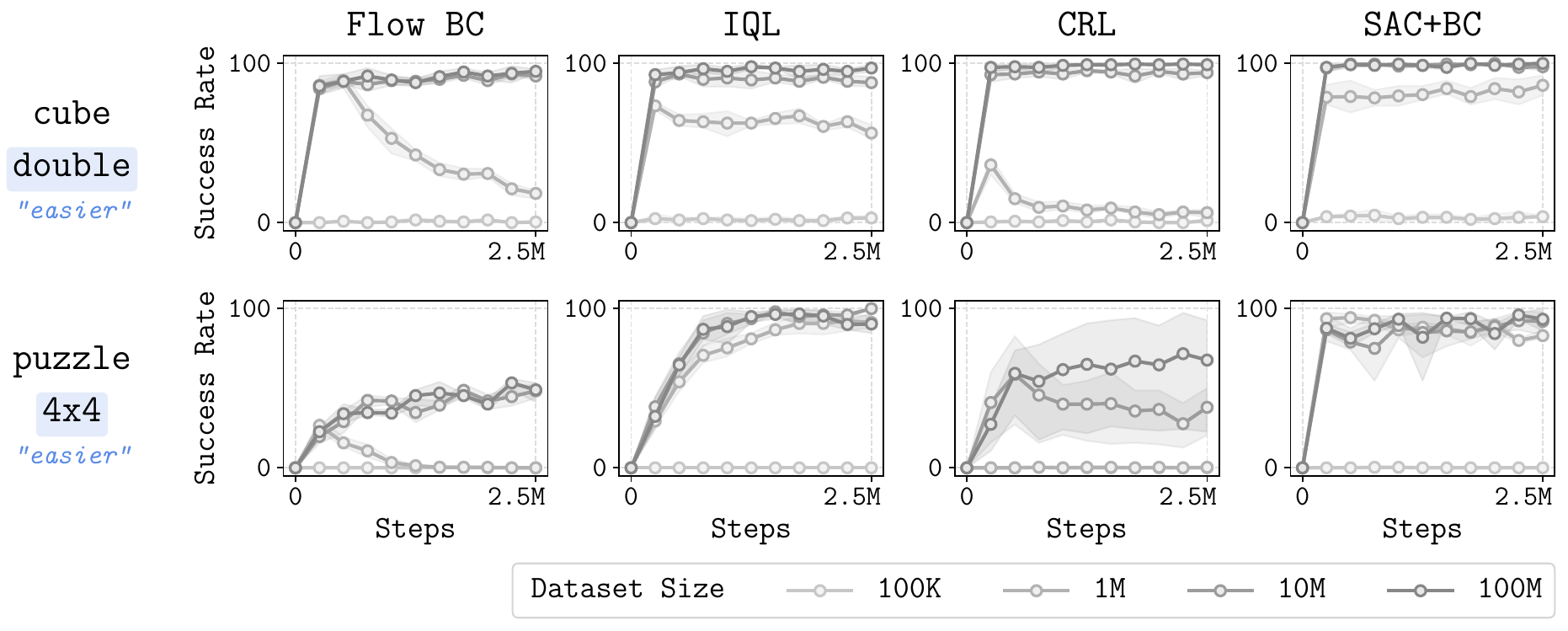}
    \caption{
    \footnotesize
    \textbf{Training curves.}
    }
    \end{subfigure}
    \begin{subfigure}[t]{1.0\textwidth}
    \vspace{10pt}
    \includegraphics[width=1.0\textwidth]{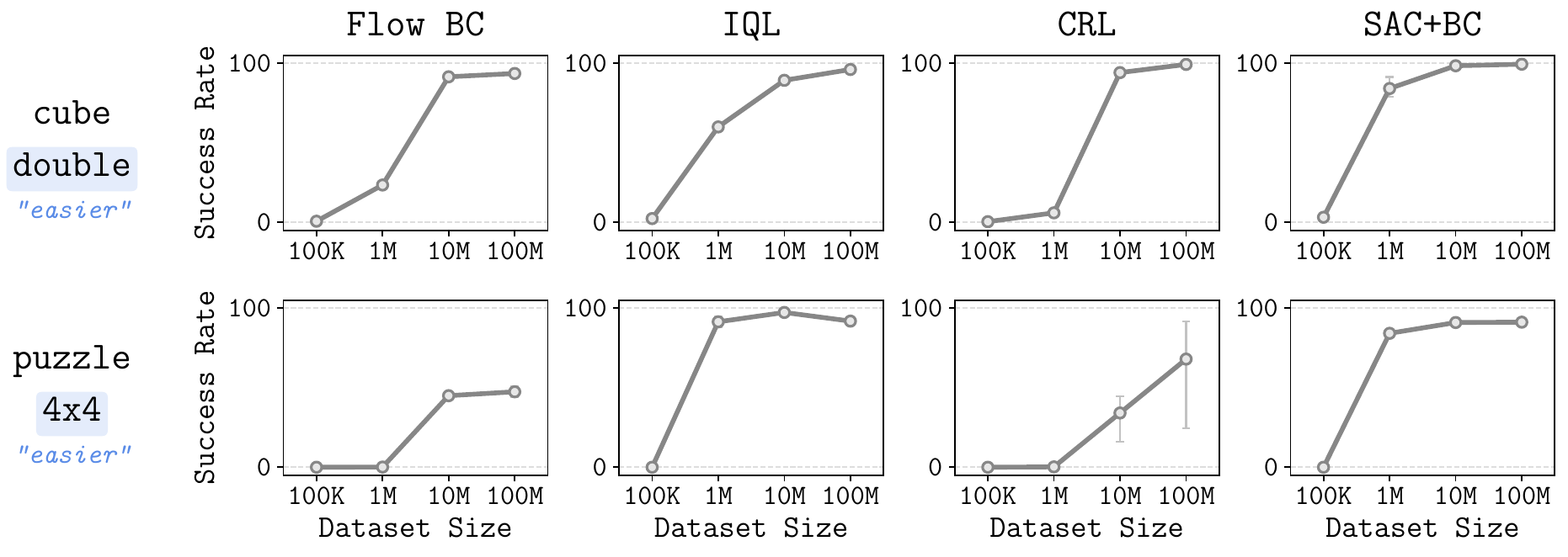}
    \caption{
    \footnotesize
    \textbf{Data-scaling curves.}
    }
    \end{subfigure}
    \vspace{5pt}
    \caption{
    \footnotesize
    \textbf{Offline RL scales well on easier, shorter-horizon tasks.}
    We evaluate flow BC, IQL, CRL, and SAC+BC on the same tasks with fewer objects,
    and show that they generally scale well on these simpler tasks.
    }
    \label{fig:flat_easy}
\end{figure}

To further verify the validity of our benchmark tasks as well as the offline RL algorithms considered in \Cref{sec:flat},
we evaluate these methods on the same tasks with fewer objects:
\tt{cube-double} with $2$ cubes (as opposed to \tt{cube-octuple} with $8$ cubes)
and \tt{puzzle-4x4} with $16$ buttons (as opposed to \tt{puzzle-4x5} with $20$ buttons).
\Cref{fig:flat_easy} shows the training and data-scaling curves of flow BC, IQL, CRL, and SAC+BC on the two tasks.
The results suggest that current offline RL algorithms generally scale well on these easier, shorter-horizon tasks.
This confirms that our dataset \emph{distribution} provides sufficient coverage to learn a near-optimal policy,
and serves as a sanity check for our implementations of these offline RL algorithms

\section{Other attempts to fix the scalability of offline RL}
\label{sec:flat_abl}

\begin{figure}[h!]
    \centering
    \includegraphics[width=1.0\textwidth]{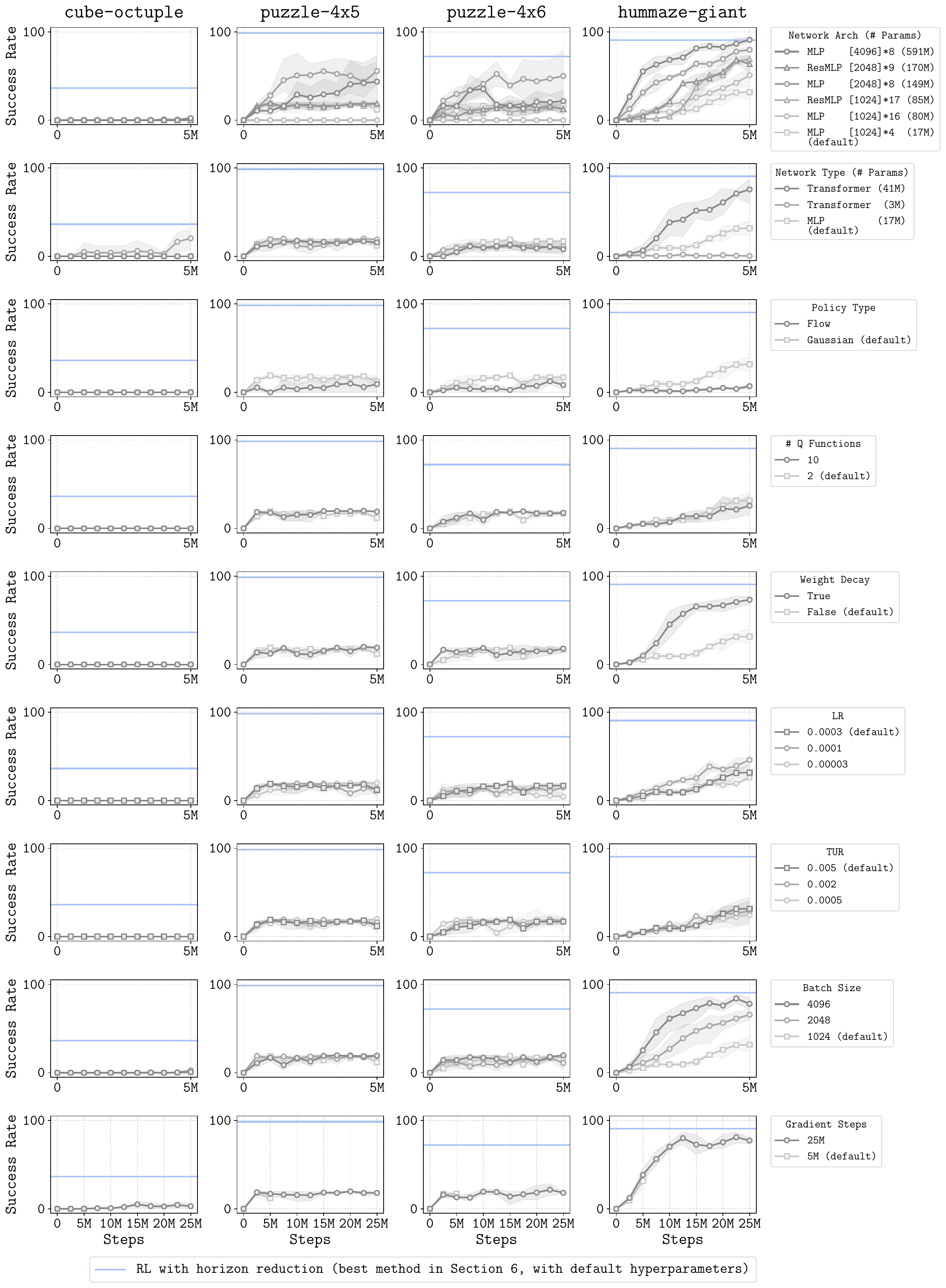}
    \vspace{-10pt}
    \caption{
    \footnotesize
    \textbf{Nine different attempts to fix the scalability of offline RL \emph{without} horizon reductions.}
    The results show that \textbf{none} of these fixes are as effective as horizon reduction (denoted by the {\color{myblue}blue} line) in general.
    }
    \label{fig:flat_abl}
\end{figure}

In the main paper, we showed that standard (flat) offline RL methods struggle to scale on complex, long-horizon tasks,
and that horizon reduction techniques can effectively address this scalability issue.
Are there other solutions to fix scalability without reducing the horizon?
We were unable to find any techniques that are as effective as horizon reduction,
and we describe failed attempts in this section.
Unless otherwise mentioned, we employ SAC+BC and the largest $1$B datasets
in the experiments below.
We note that SAC+BC is the best method in \Cref{fig:flat_easy},
and that behavior-regularized methods of this sort achieve state-of-the-art performance on standard benchmarks~\citep{rebrac_tarasov2023}.

\textbf{Larger networks.}
The first row of \Cref{fig:flat_abl} shows the results with larger networks with MLPs and residual MLPs (ResMLPs)~\citep{bro_nauman2024, simba_lee2025},
up to $591$M-sized models.
To stabilize training, we reduce the learning rate of the largest $591$M model from $0.0003$ to $0.0001$,
conceptually following the suggestion by \citet{mup_yang2021}.
The results suggest that while larger networks can improve performance to some degree,
simply increasing the capacity is not sufficient to master the tasks.
On the other hand, horizon reduction enables significantly better asymptotic performance (denoted in {\color{myblue}blue})
even with the default-sized models.
We refer to the main paper (\Cref{sec:flat}) for further discussion.

\textbf{Transformers.}
We investigate whether replacing MLPs with Transformers~\citep{transformer_vaswani2017} can improve performance.
To handle vector-valued inputs with a Transformer, we first map the input to a $T_r$-dimensional vector using a dense layer,
reshape it into a length-$T_\ell$ sequence of $T_k$-dimensional vectors,
pass it through $T_n$ self-attention blocks (with $T_m$ MLP units) with four independent heads,
and concatenate the outputs for the final dense layer.
We employ Transformers of two different sizes with $(T_r, T_\ell, T_k, T_n, T_m) = (2048, 16, 128, 4, 128)$ 
and $(2048, 8, 256, 10, 1024)$.
The former network has $3$M total parameters
and the latter has $41$M total parameters.
Due to the significantly higher computational cost, we use a smaller batch size ($256$ instead of $1024$)
for runs with the larger Transformer, so that each run completes within three days.
The second row of \Cref{fig:flat_abl} shows the results with Transformers.
These results suggest that while using Transformers improves performance on some tasks,
it still often falls significantly short of horizon reduction techniques.

\textbf{More expressive policies.}
To understand whether a more expressive policy can improve performance,
we train (goal-conditioned) FQL~\citep{fql_park2025},
one of the closest methods to SAC+BC that use expressive flow policies~\citep{flow_lipman2023, flow_liu2023, flow_albergo2023}.
The third row of \Cref{fig:flat_abl} presents the results,
which suggest that simply changing the policy class does not improve performance on the four benchmark tasks.

\textbf{Larger Q ensembles.}
The fourth row of \Cref{fig:flat_abl} compares the results with $2$ (default) and $10$ Q networks.
The results show that their performances are nearly identical.

\textbf{Regularization.}
To understand whether additional regularization can address the scalability issue,
we evaluate performance with weight decay (with a coefficient of $0.01$, selected from $\{0.0001, 0.001, 0.01, 0.1\}$).
We note that we use layer normalization~\citep{ln_ba2016} by default for all networks.
The fifth row of \Cref{fig:flat_abl} shows the results.
While weight decay yields a non-trivial improvement on one task (\tt{humanoidmaze-giant}),
it does not improve performance on the other three, more challenging tasks.

\textbf{Smaller learning rates (LRs) and target network update rates (TURs).}
The sixth and seventh rows of \Cref{fig:flat_abl} show the results with different learning rates and target network update rates.
These results indicate that simply adjusting these hyperparameters does not substantially improve performance on the benchmark tasks.

\textbf{Larger batch sizes.}
The eighth row of \Cref{fig:flat_abl} shows the results with larger batch sizes.
While larger batches help on \tt{humanoidmaze-giant},
they do not improve performance on the other three tasks.

\textbf{Longer training.}
The ninth row of \Cref{fig:flat_abl} shows the results with $5\times$ longer training ($25$M gradient steps in total).
While extended training improves performance on \tt{humanoidmaze-giant}, it does not yield significant improvements on the other three tasks.

\textbf{Other attempts.}
In the earlier stages of this research,
we tried a classification-based loss with HL-Gauss~\citep{hlgauss_imani2018, hlgauss_farebrother2024},
but it did not lead to a significant improvement in performance.
We also tried residual TD error minimization~\citep{residual_schweitzer1985}
(\ie, removing the stop-gradient in the TD target),
but we were unable to achieve non-trivial performance with the residual loss.

\section{Ablation studies of SHARSA}
\label{sec:sharsa_abl}

\begin{figure}[h!]
    \centering
    \includegraphics[width=1.0\textwidth]{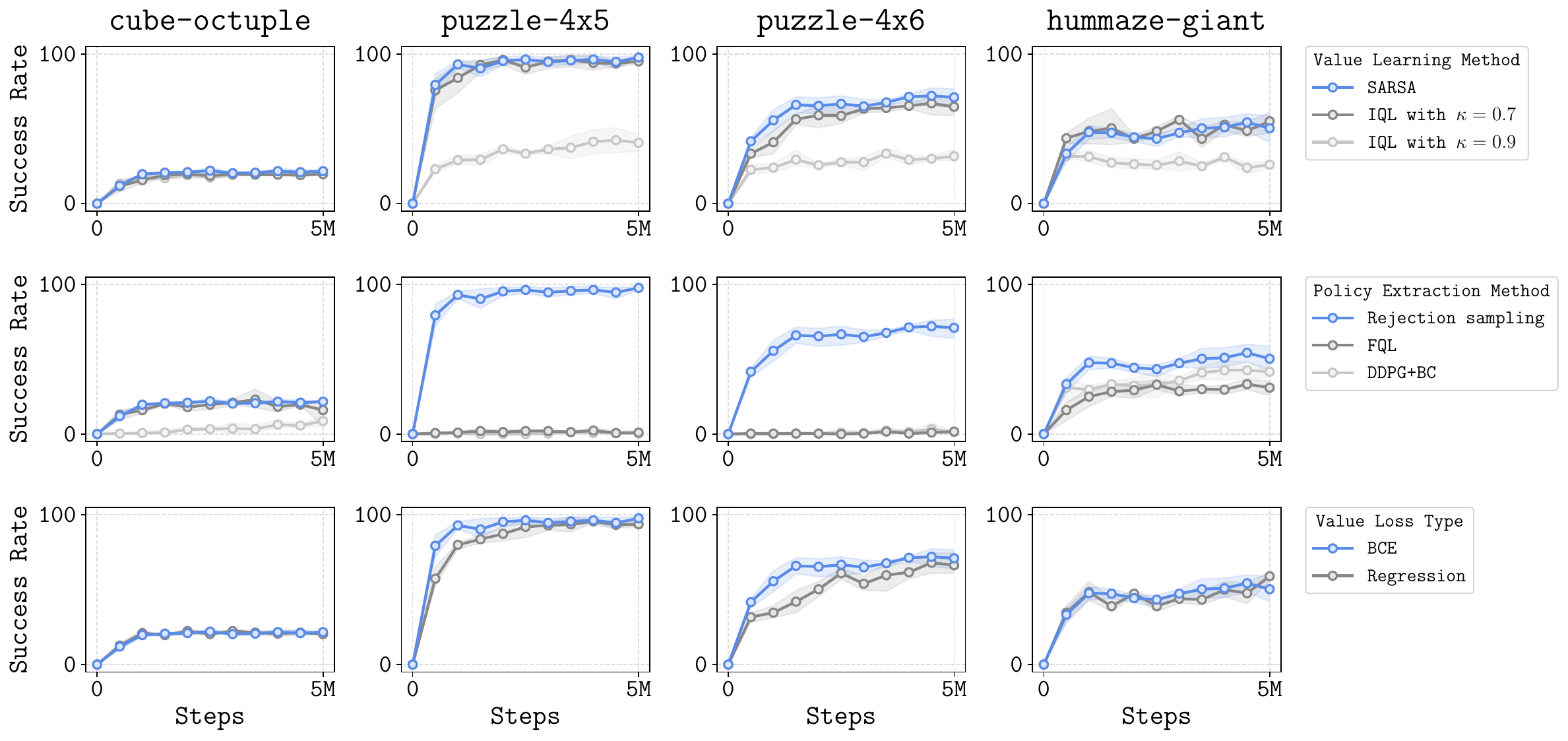}
    \vspace{-10pt}
    \caption{
    \footnotesize
    \textbf{Ablation studies of SHARSA.}
    }
    \label{fig:sharsa_abl}
\end{figure}

In this section, we present three ablation studies on the design choices of SHARSA.
All results are evaluated on the largest $1$B datasets.

\textbf{Value learning methods.}
While SHARSA uses SARSA for the value learning algorithm,
we can in principle use any decoupled value learning algorithm (\ie, one that does not involve policy learning) in place of SARSA,
such as IQL~\citep{iql_kostrikov2022} or its variants~\citep{sql_xu2023, xql_garg2023}.
The first row of \Cref{fig:sharsa_abl} compares the performance of three different value learning methods within the SHARSA framework:
SARSA, IQL with $\kappa = 0.7$, and IQL with $\kappa = 0.9$,
where $\kappa$ is the expectile hyperparameter in IQL (\Cref{sec:algo_flat}).
The results suggest that the simplest SARSA algorithm is sufficient to achieve the best performance on our benchmark tasks,
which partly aligns with recent findings~\citep{onestep_brandfonbrener2021, crl_eysenbach2022, bridge_laidlaw2023}.

\textbf{Policy extraction methods.}
SHARSA uses rejection sampling for high-level policy extraction.
In the main paper, we discussed how reparameterized gradient-based approaches
may not be suitable for \emph{high-level} policy extraction,
due to potentially ill-defined first-order gradient information in the state space.
To empirically confirm this, we replace rejection sampling in SHARSA with two alternative policy extraction methods based on reparameterized gradients:
DDPG+BC~\citep{td3bc_fujimoto2021, bottleneck_park2024} and FQL~\citep{fql_park2025}.
The former extracts a (high-level) Gaussian policy and the latter extracts a (high-level) flow policy.
We recall that SHARSA uses goal-conditioned BC for the low-level policy.
The second row of \Cref{fig:sharsa_abl} presents the results.
As expected, the results show that these reparameterized gradient-based methods perform worse than rejection sampling,
especially on the \tt{puzzle} tasks, which contain discrete information (\eg, button states) in the state space.

\textbf{Value losses.}
As explained in \Cref{sec:algo_sharsa}, we employ the binary cross-entropy (BCE) loss (instead of the more commonly used regression loss)
for the value losses in SHARSA (\Cref{eq:sharsa_v_high,eq:sharsa_q_high}).
The third row of \Cref{fig:sharsa_abl} compares these two choices,
showing that the BCE loss leads to better performance and faster convergence.
While we do not provide separate plots,
we found that the BCE loss generally results in better performance,
regardless of the underlying algorithms.

\section{Additional results}
\label{sec:add_results}

\begin{table}[h!]
\caption{
\footnotesize
\textbf{Horizon reduction improves performance in reward-based (non-goal-conditioned) RL too.}
}
\vspace{5pt}
\label{table:singletask}
\centering
\scalebox{0.65}
{

\begin{tabular}{lcccccc}
\toprule
\tt{Task} & \tt{SARSA} & \tt{IQL} & \tt{SAC+BC} & \tt{n-SAC+BC} & \tt{SHARSA} $(\kappa=0.5)$ & \tt{SHARSA} $(\kappa=0.7)$ \\
\midrule
\tt{Horizon Reduction Type} & - & - & - & \tt{Value} & \tt{Value \& policy} & \tt{Value \& Policy} \\
\midrule
$\texttt{cube-quadruple-play-singletask-task1-v0}$ & $0$ {\tiny $\pm 0$} & $7$ {\tiny $\pm 6$} & $0$ {\tiny $\pm 0$} & $0$ {\tiny $\pm 0$} & $50$ {\tiny $\pm 9$} & $\mathbf{60}$ {\tiny $\pm 8$} \\
$\texttt{puzzle-4x5-play-singletask-task1-v0}$ & $4$ {\tiny $\pm 3$} & $24$ {\tiny $\pm 7$} & $0$ {\tiny $\pm 0$} & $0$ {\tiny $\pm 0$} & $\mathbf{94}$ {\tiny $\pm 3$} & $\mathbf{94}$ {\tiny $\pm 2$} \\
$\texttt{puzzle-4x6-play-singletask-task1-v0}$ & $2$ {\tiny $\pm 2$} & $5$ {\tiny $\pm 2$} & $0$ {\tiny $\pm 0$} & $0$ {\tiny $\pm 0$} & $\mathbf{14}$ {\tiny $\pm 6$} & $\mathbf{14}$ {\tiny $\pm 7$} \\
$\texttt{humanoidmaze-giant-navigate-singletask-task1-v0}$ & $0$ {\tiny $\pm 0$} & $2$ {\tiny $\pm 2$} & $44$ {\tiny $\pm 35$} & $\mathbf{88}$ {\tiny $\pm 3$} & $26$ {\tiny $\pm 4$} & $\mathbf{87}$ {\tiny $\pm 3$} \\
\bottomrule
\end{tabular}

}
\end{table}

\textbf{Results on reward-based tasks.}
While we focus on goal-conditioned tasks in this work,
the benefits of horizon reduction are not limited to goal-conditioned RL.
To empirically demonstrate this,
we additionally evaluate two horizon reduction techniques,
$n$-step SAC+BC (which reduces the value horizon) and SHARSA (which reduces both the value and policy horizons),
on four reward-based \tt{singletask} tasks from OGBench~\citep{ogbench_park2025}.
We employ $100$M-sized (\tt{cube}) and $1$B-sized (others) datasets.

On these tasks, we evaluate SARSA,
IQL (with $\kappa = 0.7$),
SAC+BC,
$n$-step SAC+BC, and SHARSA.
Additionally, we consider an IQL variant of SHARSA (with $\kappa = 0.7$, \Cref{sec:sharsa_abl}),
which can be helpful as these \tt{singletask} tasks have higher suboptimality due to the absence of hindsight relabeling.
We use AWR with $\alpha = 10$ for SARSA and IQL,
and BC regularization with $\alpha = 0.01$ (\tt{humanoidmaze}) or $0.1$ (others) for SAC+BC and $n$-step SAC+BC.

\Cref{table:singletask} shows the performance measured at the $1$M epoch.
The results suggest that these horizon reduction techniques significantly improve performance in reward-based offline RL as well. 

\begin{figure}[t!]
    \centering
    \includegraphics[width=1.0\textwidth]{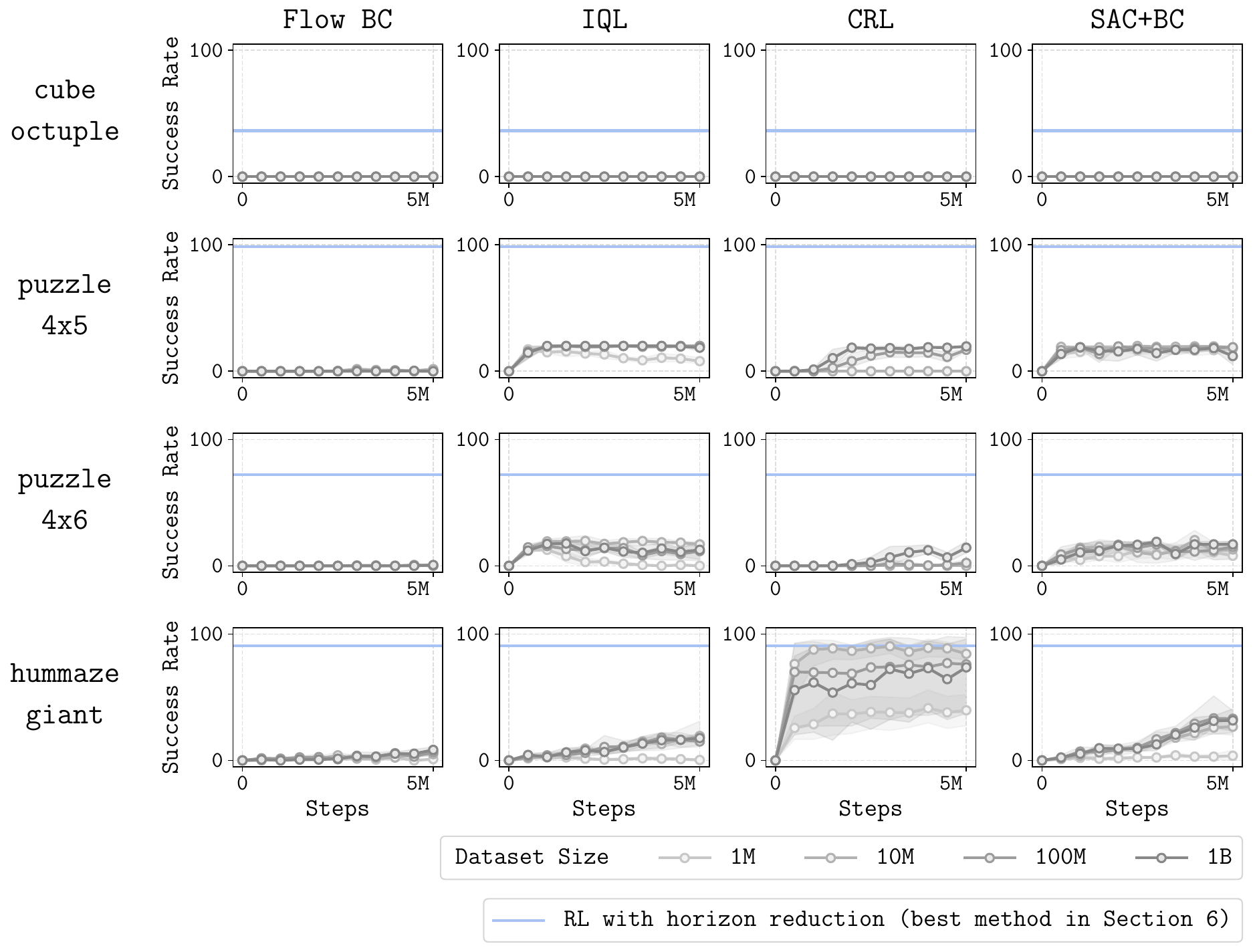}
    \vspace{-10pt}
    \caption{
    \footnotesize
    \textbf{Training curves of standard offline RL methods.}
    }
    \label{fig:flat_main}
\end{figure}

\begin{figure}[t!]
    \centering
    \includegraphics[width=1.0\textwidth]{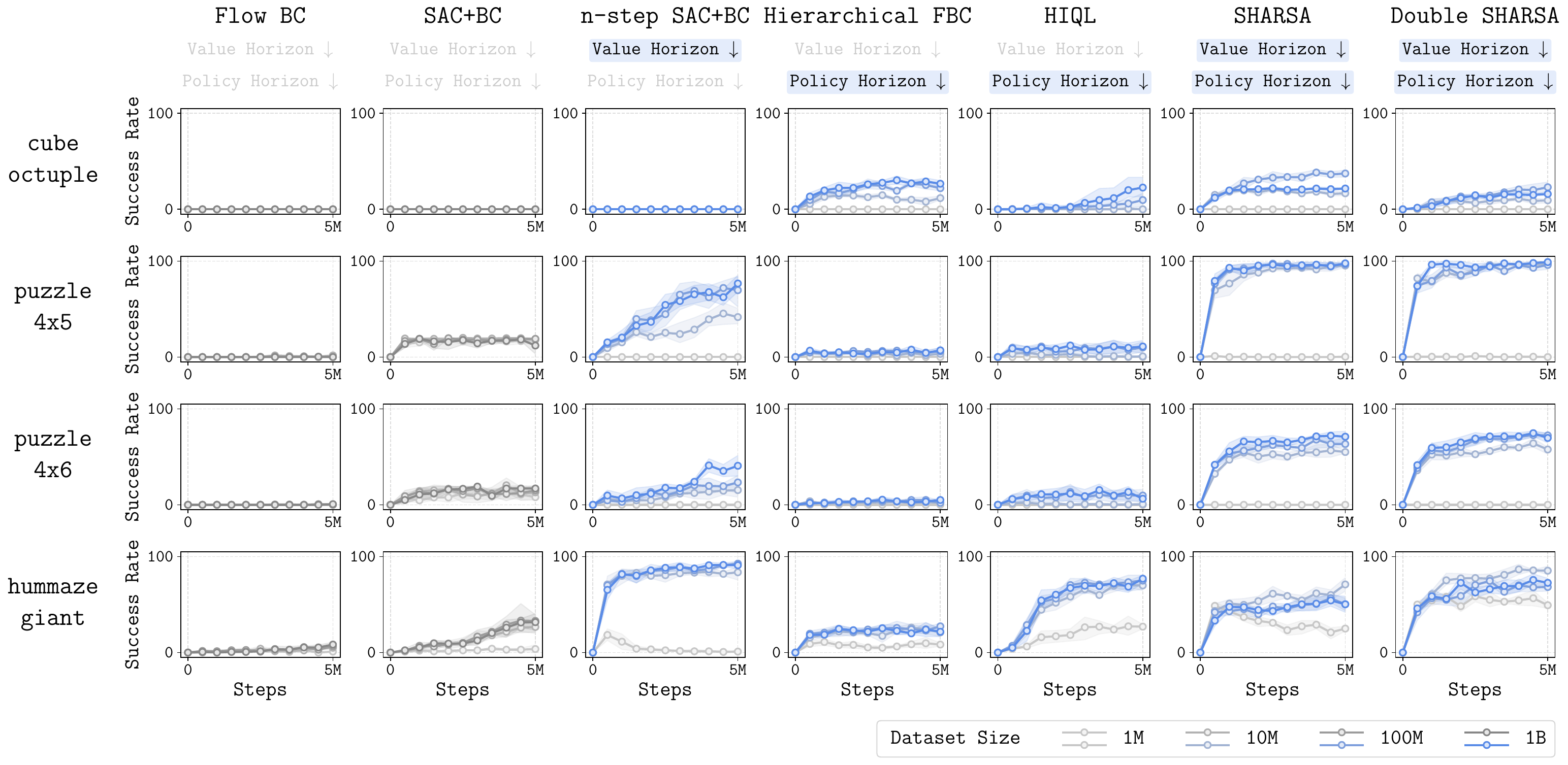}
    \vspace{-10pt}
    \caption{
    \footnotesize
    \textbf{Training curves of horizon reduction techniques.}
    }
    \label{fig:hrl_main}
\end{figure}

\textbf{Full training curves.}
\Cref{fig:flat_main,fig:hrl_main} provide the full training curves of the methods considered in \Cref{fig:flat_data,fig:hrl_data}, respectively.

\clearpage

\section{Offline RL algorithms}
\label{sec:algo}

In this section, we describe the offline (goal-conditioned) RL algorithms considered in this work.
In the below, $\gamma \in [0, 1]$ denotes the discount factor
and $\gG$ denotes the goal space, which is the domain of a goal specification function $\phi_g(\pl{s}): \gS \to \gG$.
For example, in \tt{humanoidmaze}, $\phi_g$ is a function that outputs only the $x$-$y$ coordinates of the state.
We also assume that the action space is a $d$-dimensional Euclidean space (\ie, $\gA = \sR^d$), unless otherwise mentioned.
We denote network parameters as $\theta$ (with corresponding subscripts when there are multiple networks).
The goal-conditioned reward function $r(\pl{s}, \pl{g}): \gS \times \gG \to \sR$ is given as
either $[s=g]$ (the $0$-$1$ sparse reward function) or $[s=g] - 1$ (the $-1$-$0$ sparse reward function),
where $[\cdot]$ is the Iverson bracket (\ie, the indicator function for propositions).
We use the former for classification-based methods and the latter for regression-based methods.

We denote the state-action-goal sampling distribution as $p^\gD$.
In general, $p^\gD(\pl{s}, \pl{a})$ is the uniform distribution over the dataset state-action pairs
and $p^\gD(\pl{g} \mid s, a)$ is a mixture of the four distributions:
the Dirac delta distribution at the current state ($p^\gD_\mathrm{cur}$),
a geometric distribution over the future states ($p^\gD_\mathrm{geom}$),
the uniform distribution over the future states ($p^\gD_\mathrm{traj}$),
and the uniform distribution over the dataset states ($p^\gD_\mathrm{rand}$).
We refer to \citet{ogbench_park2025} for the full details.
The ratios of these four distributions are tunable hyperparameters, which we specify in \Cref{table:hyp}.

\subsection{Flat offline RL algorithms}
\label{sec:algo_flat}

\textbf{Flow behavioral cloning (flow BC)~\citep{diffusionpolicy_chi2023, pi0_black2024, fql_park2025}.}
Goal-conditioned flow behavioral cloning
trains a vector field $v_\theta(\pl{t}, \pl{s}, \pl{z}, \pl{g}): [0, 1] \times \gS \times \sR^d \times \gG \to \sR^d$
that generates behavioral action distributions.
It minimizes the following objective:
\begin{align}
    L(\theta) = \E_{\substack{s,a \sim p^\gD(\pl{s}, \pl{a}),\ g \sim p^\gD(\pl{g} \mid s, a), \\ z \sim \gN(0, I_d), \ t \sim \mathrm{Unif}([0, 1]), \\ a^t = (1-t)z + ta}}
    \left[\left\|v_\theta(t, s, a^t, g) - (a - z)\right\|_2^2 \right],
\end{align}
where $\mathrm{Unif}([0, 1])$ denotes the uniform distribution over the interval $[0, 1]$.

After training the vector field $v_\theta$, actions are obtained by solving the ordinary differential equation (ODE)
corresponding to the \emph{flow}~\citep{smooth_lee2002} generated by the vector field.
We use the Euler method with a step count of $10$, following prior work~\citep{fql_park2025}.
See \citet{flow_lipman2024, fql_park2025} for further discussions about flow matching and flow policies.

\textbf{Implicit Q-learning (IQL)~\citep{iql_kostrikov2022, hiql_park2023}.}
Goal-conditioned IQL trains a state value function $V_{\theta_V}(\pl{s}, \pl{g}): \gS \times \gG \to \sR$
and a state-action value function $Q_{\theta_Q}(\pl{s}, \pl{a}, \pl{g}): \gS \times \gA \times \gG \to \sR$
with the following losses:
\begin{align}
    L^V(\theta_V) &= \E_{(s, a) \sim p^\gD(\pl{s}, \pl{a}),\  g \sim p^\gD(\pl{g} \mid s, a)}
    \left[\ell_\kappa^2 \left(V_{\theta_V}(s, g) - Q_{\bar \theta_Q}(s, a, g)\right)\right], \\
    L^Q(\theta_Q) &= \E_{(s, a, s') \sim p^\gD(\pl{s}, \pl{a}, \pl{s'}),\  g \sim p^\gD(\pl{g} \mid s, a)}
    \left[\left(Q_{\theta_Q}(s, a, g) - r(s, g) - \gamma V_{\theta_V}(s', g)\right)^2\right],
\end{align}
where $\ell_\kappa^2$ denotes the expectile loss, $\ell_\kappa^2(x) = |\kappa - [x < 0]|x^2$,
and $\bar \theta_Q$ denotes the parameters of the target Q network~\citep{dqn_mnih2013}.

From the learned Q function, it extracts a (Gaussian) policy $\pi_{\theta_\pi}(\pl{a} \mid \pl{s}, \pl{g}): \gS \times \gG \to \Delta(\gA)$
by maximizing the following DDPG+BC objective~\citep{bottleneck_park2024}:
\begin{align}
J^\pi(\theta_\pi) = \E_{(s, a) \sim p^\gD(\pl{s}, \pl{a}),\ g \sim p^\gD(\pl{g} \mid s, a)}
\left[ Q_{\theta_Q}(s, \mu_{\theta_\pi}(s, g), g) + \alpha \log \pi_{\theta_\pi}(a \mid s, g) \right],
\end{align}
where $\mu_{\theta_\pi}$ denotes the mean of the Gaussian policy $\pi_{\theta_\pi}$
and $\alpha$ denotes the hyperparameter that controls the strength of the BC regularizer.
While the original IQL method uses the AWR objective~\citep{awr_peng2019},
we use DDPG+BC as \citet{bottleneck_park2024} found it to scale better than AWR.

\textbf{Contrastive reinforcement learning (CRL)~\citep{crl_eysenbach2022}.}
CRL trains a logarithmic goal-conditioned value function $f_{\theta_f}(\pl{s}, \pl{a}, \pl{g}): \gS \times \gA \times \gG \to \sR$ with
the following binary noise contrastive estimation objective~\citep{nce_ma2018}:
\begin{align}
    J^f(\theta_f) = \E_{\substack{(s, a) \sim p^\gD(\pl{s}, \pl{a}), g \sim p^\gD_+(\pl{g} \mid s, a), \\ g^- \sim p^\gD_-(\pl{g})}}
    \left[\log \sigma(f_{\theta_f}(s, a, g)) + \log (1 - \sigma(f_{\theta_f}(s, a, g^-)))\right],
\end{align}
where $\sigma \colon \sR \to (0, 1)$ is the sigmoid function,
$p_+^\gD$ is a geometric future goal sampling distribution,
and $p_-^\gD$ is the uniform goal sampling distribution.
\citet{crl_eysenbach2022} show that the optimal solution $f^*$ to the above objective is given by
$f^*(s, a, g) = \log Q^\mathrm{MC}(s, a, g) + C(g)$,
where $Q^\mathrm{MC}$ is the Monte-Carlo value function and $C$ is a function that does not depend on $s$ and $a$.
As in \citet{crl_eysenbach2022}, we employ an inner product parameterization to model $f_{\theta_f}$ as $f(s, a, g) = \psi_1(s, a)^\top \psi_2(g)$
(where we omit the parameter dependencies for simplicity)
with $k$-dimensional representations, $\psi_1: \gS \times \gA \to \sR^k$ and $\psi_2: \gG \to \sR^k$.

Similarly to IQL, CRL extracts a policy with the following DDPG+BC objective:
\begin{align}
    J^\pi(\theta_\pi) = \E_{(s, a) \sim p^\gD(\pl{s}, \pl{a}),\ g \sim p^\gD(\pl{g} \mid s, a)}
    \left[ f_{\theta_f}(s, \mu_{\theta_\pi}(s, g), g) + \alpha \log \pi_{\theta_\pi}(a \mid s, g) \right].
\end{align}

\textbf{Soft actor-critic + behavioral cloning (SAC+BC).}
SAC+BC is the SAC~\citep{sac_haarnoja2018} variant of TD3+BC~\citep{td3bc_fujimoto2021, rebrac_tarasov2023}.
We found SAC+BC to be generally better than both TD3+BC~\citep{td3bc_fujimoto2021} and its successor ReBRAC~\citep{rebrac_tarasov2023}
due to the use of stochastic actions in the actor loss (note that TD3~\citep{td3_fujimoto2018} uses deterministic actions in the actor objective),
which serves as a regularizer in the offline RL setting.
Goal-conditioned SAC+BC minimizes $L^Q$ and maximizes $J^\pi$ below
to train a Q function $Q_{\theta_Q}$ and a policy $\pi_{\theta_\pi}$:
\begin{align}
    L^Q(\theta_Q) &= \E_{\substack{(s, a, s') \sim p^\gD(\pl{s}, \pl{a}, \pl{s'}), \ g \sim p^\gD(\pl{g} \mid s, a),\\ a^\pi \sim \pi_{\theta_\pi}(\pl{a} \mid s', g)}}
    \left[ \left(Q_{\theta_Q}(s, a, g) - r(s, g) - \gamma Q_{\bar \theta_Q}(s', a^\pi, g) \right)^2 \right], \\
    J^\pi(\theta_\pi) &= \E_{\substack{(s, a) \sim p^\gD(\pl{s}, \pl{a}), \\ a^\pi \sim \pi_{\theta_\pi}(\pl{a} \mid s, g)}}
    \left[ Q_{\theta_Q}(s, a^\pi, g) - \alpha \|a^\pi - a\|_2^2 - \lambda \log \pi_{\theta_\pi}(a^\pi \mid s, g) \right], \label{eq:sacbc_actor}
\end{align}
where $\alpha$ is the hyperparameter for the BC strength and $\lambda$ is the entropy regularizer
(which is often automatically adjusted with dual gradient descent to match a target entropy value~\citep{saces_haarnoja2018}).

On goal-conditioned tasks with $0$-$1$ sparse rewards,
since we know that the optimal Q values are always in between $0$ and $1$,
we can instead use the following \emph{binary cross-entropy (BCE)} variant for the Q loss~\citep{qtopt_kalashnikov2018}:
\begin{align}
    L_{\mathrm{BCE}}^Q(\theta_Q) &= \E_{\substack{(s, a, s') \sim p^\gD(\pl{s}, \pl{a}, \pl{s'}), \ g \sim p^\gD(\pl{g} \mid s, a),\\ a^\pi \sim \pi_{\theta_\pi}(\pl{a} \mid s', g)}}
    \left[ \mathrm{BCE} \left(Q_{\theta_Q}(s, a, g), r(s, g) + \gamma Q_{\bar \theta_Q}(s', a^\pi, g) \right) \right],
    \label{eq:bce}
\end{align}
where $\mathrm{BCE}(x, y) = -y \log x - (1-y)\log(1-x)$.
We found this variant to be generally better than the original regression objective,
as it focuses better on small differences in low Q values,
which is crucial for extracting policies on long-horizon tasks.
When using the binary cross-entropy variant,
we model the \emph{logits} of Q values (instead of the raw Q values) with a neural network,
and use the logit values in place of the Q values in the SAC+BC actor objective as follows:
\begin{align}
    J_{\mathrm{BCE}}^\pi(\theta_\pi) &= \E_{\substack{(s, a) \sim p^\gD(\pl{s}, \pl{a}), \\ a^\pi \sim \pi_{\theta_\pi}(\pl{a} \mid s, g)}}
    \left[ \operatorname{logit} Q_{\theta_Q}(s, a^\pi, g) - \alpha \|a^\pi - a\|_2^2 - \lambda \log \pi_{\theta_\pi}(a^\pi \mid s, g) \right].
\end{align}
We found the use of logits in the actor loss to be crucial on long-horizon tasks,
as it applies more uniform behavioral constraints across the state space.

\textbf{Flow Q-learning (FQL)~\citep{fql_park2025}.}
FQL is a behavior-regularized offline RL algorithm that trains a flow policy with one-step distillation.
Goal-conditioned FQL trains a Q function $Q_{\theta_Q}(\pl{s}, \pl{a}, \pl{g}): \gS \times \gA \times \gG \to \sR$,
a BC vector field $v_{\theta_\pi}(\pl{t}, \pl{s}, \pl{z}, \pl{g}): [0, 1] \times \gS \times \sR^d \times \gG \to \sR^d$
that generates a noise-conditioned flow BC policy $\mu_{\theta_\pi}(\pl{s}, \pl{z}, \pl{g}): \gS \times \sR^d \times \gG \to \gA$,
and a noise-conditioned one-step policy $\mu_{\theta_\mu}(\pl{s}, \pl{z}, \pl{g}): \gS \times \sR^d \times \gG \to \gA$,
with the following losses:
\begin{align}
    L^{\pi}(\theta_\pi) &= \E_{\substack{(s,a) \sim p^\gD(\pl{s}, \pl{a}),\ g \sim p^\gD(\pl{g} \mid s, a), \\ z \sim \gN(0, I_d), \ t \sim \mathrm{Unif}([0, 1]), \\ a^t=(1-t)z + ta}}
    \left[\left\|v_{\theta_\pi}(t, s, a^t, g) - (a - z)\right\|_2^2 \right], \\
    L^Q(\theta_Q) &= \E_{\substack{(s, a, s') \sim p^\gD(\pl{s}, \pl{a}, \pl{s'}), \ g \sim p^\gD(\pl{g} \mid s, a),\\z \sim \gN(0, I_d), \ a^\pi = \mu_{\theta_\mu}(s', z, g)}}
    \left[ \left(Q_{\theta_Q}(s, a, g) - r(s, g) - \gamma Q_{\bar \theta_Q}(s', a^\pi, g) \right)^2 \right], \label{eq:fql_regression} \\
    L^{\mu}(\theta_\mu) &= \E_{\substack{(s, a) \sim p^\gD(\pl{s}, \pl{a}), \\ g \sim p^\gD(\pl{g} \mid s, a), \\ z \sim \gN(0, I_d)}}
    \left[ -Q(s, \mu_{\theta_\mu}(s, z, g), g) + \alpha \|\mu_{\theta_\mu}(s, z, g) - \mu_{\theta_\pi}(s, z, g)  \|_2^2 \right],
\end{align}
where $\alpha$ is the BC coefficient.
The output of FQL is the one-step policy $\mu_{\theta_\mu}$.
In our experiments,
we use the binary cross-entropy variant of FQL in our experiments,
which replaces the regression loss in \Cref{eq:fql_regression} with the corresponding binary cross-entropy loss,
as in \Cref{eq:bce}.

\subsection{Hierarchical offline RL algorithms}
\label{sec:algo_hrl}

\textbf{$\bm{n}$-step soft actor-critic + behavioral cloning ($\bm{n}$-step SAC+BC).}
$n$-step SAC+BC is a variant of SAC+BC that employs $n$-step returns.
The only difference from SAC+BC is that it minimizes the following value loss:
\begin{align}
    L^Q(\theta_Q) &= \E_{\substack{(s_h, a_h, \ldots, s_{h+n}) \sim p^\gD,\\ g \sim p^\gD(\pl{g} \mid s_h, a_h),\\ a^\pi \sim \pi_{\theta_\pi}(\pl{a} \mid s_{h+n}, g)}}
    \left[ D \left(Q_{\theta_Q}(s_h, a_h, g), \sum_{i=0}^{n-1} \gamma^i r(s_{h+i}, g) + \gamma^n Q_{\bar \theta_Q}(s_{h+n}, a^\pi, g) \right) \right],
\end{align}
where we omit the arguments in $p^\gD(\pl{s_h}, \pl{a_h}, \ldots, \pl{s_{h+n}})$
and $D$ is either the regression loss $\mathrm{Reg}(x, y) = (x - y)^2$
or the binary cross-entropy loss $\mathrm{BCE}(x, y) = -y \log x - (1-y)\log(1-x)$.
We found that the BCE loss performs and scales better in our experiments.
In practice,
we also need to handle several edge cases involving truncated trajectories and goals in the above loss;
we refer the reader to our implementation for further details.

\textbf{Hierarchical flow BC (hierarchical FBC).}
Hierarchical flow BC trains two policies:
a high-level policy $\pi^h_{\theta_h}(\pl{w} \mid \pl{s}, \pl{g}): \gS \times \gG \to \Delta(\gG)$
and a low-level policy $\pi^\ell_{\theta_\ell}(\pl{a} \mid \pl{s}, \pl{w}): \gS \times \gG \to \Delta(\gA)$,
where we denote subgoals by $w$.
The high-level policy is trained to predict subgoals that are $n$ steps away from the current state,
and the low-level policy is trained to predict actions to reach the given subgoal.
Both policies are modeled by flows,
with vector fields $v^h_{\theta_h}(\pl{t}, \pl{s}, \pl{z}, \pl{g}): [0, 1] \times \gS \times \sR^m \times \gG \to \sR^m$ and
$v^\ell_{\theta_\ell}(\pl{t}, \pl{s}, \pl{z}, \pl{w}): [0, 1] \times \gS \times \sR^d \times \gG \to \sR^d$,
where we assume that the goal space is $\gG = \sR^m$.
These vector fields are trained with the following flow-matching losses:
\begin{align}
    L^h(\theta_h) &= \E_{\substack{(s_h, a_h, \ldots, s_{h+n}) \sim p^\gD, \ 
    g \sim p^\gD(\pl{g} \mid s_h, a_h), \\
    z \sim \gN(0, I_m), \ 
    t \sim \mathrm{Unif}([0, 1]), \\
    w^t = (1-t)z + t\phi_g(s_{t+h})}}
    \left[\left\|v^h_{\theta_h}(t, s_h, w^t, g) - (\phi_g(s_{t+h}) - z)\right\|_2^2 \right], \\
    L^\ell(\theta_\ell) &= \E_{\substack{(s_h, a_h, \ldots, s_{h+n}) \sim p^\gD, \ 
    z \sim \gN(0, I_d), \\
    t \sim \mathrm{Unif}([0, 1]), \ 
    a^t = (1-t)z + ta_h}}
    \left[\left\|v^\ell_{\theta_\ell}(t, s_h, a^t, s_{h+n}) - (a_h - z)\right\|_2^2 \right].
\end{align}

\textbf{Hierarchical implicit Q-learning (HIQL)~\citep{hiql_park2023}.}
HIQL trains a single goal-conditioned value function with implicit V-learning (IVL)~\citep{ogbench_park2025},
and extract hierarchical policies ($\pi^h_{\theta_h}$ and $\pi^\ell_{\theta_\ell}$) with AWR-like objectives~\citep{awr_peng2019}.
It trains a value function $V_{\theta_V}(\pl{s}, \pl{g}): \gS \times \gG \to \sR$
that is parameterized as $V_{\theta_V}(s, g) = \tilde V_{\theta_V}(s, \psi_{\theta_V}(s, g))$ with
a representation function $\psi_{\theta_V}(\pl{s}, \pl{g}): \gS \times \gG \to \sR^k$
and a remainder network $\tilde V_{\theta_V}(\pl{s}, \pl{z}): \gS \times \sR^k \to \sR$,
where we do not distinguish the parameters for $\psi$, $\tilde V$, and $V$
to emphasize that they are part of the value network.
The IVL value loss is as follows:
\begin{align}
    L^V(\theta_V) = \E_{(s, a, s') \sim p^\gD(\pl{s}, \pl{a}, \pl{s'}), \ g \sim p^\gD(\pl{g} \mid s, a)}
    \left[\ell_\kappa^2 \left(V_{\theta_V}(s, g) - r(s, g) - \gamma V_{\bar \theta_V}(s', g)\right)\right],
\end{align}
where $\ell_\kappa^2$ is the expectile loss described in \Cref{sec:algo_flat}.
From the value function, it extracts two policies by maximizing the following AWR objectives:
\begin{align}
    J^h(\theta_h) &= \E_{\substack{(s_h, a_h, \ldots, s_{h+n}) \sim p^\gD,\\ g \sim p^\gD(\pl{g} \mid s_h, a_h)}}
    \left[ e^{\alpha (V(s_{h+n}, g) - V(s_h, g))} \log \pi^h_{\theta_h}(\psi_{\theta_V}(s_h, s_{h+n}) \mid s_h, g) \right], \\
    J^\ell(\theta_\ell) &= \E_{(s_h, a_h, \ldots, s_{h+n}) \sim p^\gD}
    \left[ e^{\alpha (V(s_{h+1}, s_{h+n}) - V(s_h, s_{h+n}))} \log \pi^\ell_{\theta_\ell} (a_h \mid s_h, \psi_{\theta_V}(s_h, s_{h+n})) \right],
\end{align}
where $\alpha$ is the inverse temperature hyperparameter for AWR.
Similar to $n$-step SAC+BC, there are several edge cases with truncated trajectories and goals,
and we refer to our implementation for the full details.

\subsection{SHARSA}
\label{sec:algo_sharsa}

\textbf{SHARSA.}
SHARSA is our newly proposed offline RL algorithm
based on hierarchical flow BC and $n$-step SARSA.
It has the following components:
\begin{itemize}
    \item High-level BC flow policy $\pi^h_{\beta, \theta_h}(\pl{w} \mid \pl{s}, \pl{g}): \gS \times \gG \to \Delta(\gG)$,
    \item Low-level BC flow policy $\pi^\ell_{\beta, \theta_\ell}(\pl{a} \mid \pl{s}, \pl{w}): \gS \times \gG \to \Delta(\gA)$,
    \item $n$-step Q function $Q_{\theta_Q}(\pl{s}, \pl{w}, \pl{g}): \gS \times \gG \times \gG \to \sR$,
    \item $n$-step V function $V_{\theta_V}(\pl{s}, \pl{g}): \gS \times \gG \to \sR$.
\end{itemize}
As in hierarchical FBC, the policies are modeled by vector fields, $v^h_{\theta_h}(\pl{t}, \pl{s}, \pl{z}, \pl{g}): [0, 1] \times \gS \times \sR^m \times \gG \to \sR^m$ and
$v^\ell_{\theta_\ell}(\pl{t}, \pl{s}, \pl{z}, \pl{w}): [0, 1] \times \gS \times \sR^d \times \gG \to \sR^d$, where we recall that $\gG = \sR^m$ and $\gA = \sR^d$.
They are trained via the following flow behavioral cloning losses:
\begin{align}
    L^h(\theta_h) &= \E_{\substack{(s_h, a_h, \ldots, s_{h+n}) \sim p^\gD, \ 
    g \sim p^\gD(\pl{g} \mid s_h, a_h), \\
    z \sim \gN(0, I_m), \ 
    t \sim \mathrm{Unif}([0, 1]), \\
    w^t = (1-t)z + t\phi_g(s_{t+h})}}
    \left[\left\|v^h_{\theta_h}(t, s_h, w^t, g) - (\phi_g(s_{t+h}) - z)\right\|_2^2 \right], \label{eq:sharsa_pi_high} \\
    L^\ell(\theta_\ell) &= \E_{\substack{(s_h, a_h, \ldots, s_{h+n}) \sim p^\gD, \ 
    z \sim \gN(0, I_d), \\
    t \sim \mathrm{Unif}([0, 1]), \ 
    a^t = (1-t)z + ta_h}}
    \left[\left\|v^\ell_{\theta_\ell}(t, s_h, a^t, s_{h+n}) - (a_h - z)\right\|_2^2 \right], \label{eq:sharsa_pi_low}
\end{align}
where we recall that $\phi_g$ is the goal specification function defined in the first paragraph of \Cref{sec:algo}.
The value functions are trained with the following SARSA losses:
\begin{align}
    L^V(\theta_V) &= \E_{\substack{(s_h, a_h, \ldots, s_{h+n}) \sim p^\gD,\\ g \sim p^\gD(\pl{g} \mid s_h, a_h)}}
    \left[ D\left( V^h_{\theta_V}(s_h, g), Q^h_{\bar \theta_Q}(s_h, s_{h+n}, g)\right)\right], \label{eq:sharsa_v_high} \\
    L^Q(\theta_Q) &= \E_{\substack{(s_h, a_h, \ldots, s_{h+n}) \sim p^\gD,\\ g \sim p^\gD(\pl{g} \mid s_h, a_h)}}
    \left[ D\left(Q^h_{\theta_Q}(s_h, s_{h+n}, g), \sum_{i=0}^{n-1} \gamma^i r(s_{h+i}, g) + \gamma^n V^h_{\theta_V}(s_{h+n}, g) \right) \right], \label{eq:sharsa_q_high}
\end{align}
where $D$ is either the regression loss $\mathrm{Reg}(x, y) = (x - y)^2$
or the binary cross-entropy loss $\mathrm{BCE}(x, y) = -y \log x - (1-y)\log(1-x)$.
As before, we found that the BCE variant works better on long-horizon tasks.

At test time, we employ rejection sampling for the high-level policy. Specifically, it defines the distribution of the high-level policy $\pi^h_{\theta_h}(\pl{w} \mid \pl{s}, \pl{g}): \gS \times \gG \to \Delta(\gG)$ as follows:
\begin{align}
    \pi^h_{\theta_h}(s, g) \stackrel{d}{=} \argmax_{w_1, \ldots, w_N: w_i \sim \pi_{\beta, \theta_h}^h(\pl{w} \mid s, g)} Q^h_{\theta_Q}(s, w_i, g),
\end{align}
where $N$ is the number of samples.
SHARSA simply uses the behavioral low-level policy; \ie, $\pi^\ell_{\theta_\ell}(s, w) \stackrel{d}{=} \pi^\ell_{\beta,\theta_\ell}(s, w)$.
We provide the pseudocode in \Cref{alg:sharsa}.

\textbf{Double SHARSA.}
Double SHARSA employs an additional round of rejection sampling in the low-level policy
to further enhance optimality.
To do this, it defines additional low-level value networks:
\begin{itemize}
    \item Low-level Q function $Q^\ell_{\theta_q}(\pl{s}, \pl{a}, \pl{w}): \gS \times \gA \times \gG \to \sR$,
    \item Low-level V function $V^\ell_{\theta_v}(\pl{s}, \pl{w}): \gS \times \gG \to \sR$.
\end{itemize}
They are trained with the following SARSA losses:
\begin{align}
    L^v(\theta_v) &= \E_{\substack{(s_h, a_h, \ldots, s_{h+n}) \sim p^\gD}}
    \left[ D\left( V^\ell_{\theta_v}(s_h, s_{h+n}), Q^\ell_{\bar \theta_q}(s_h, a_h, s_{h+n})\right)\right], \\
    L^q(\theta_q) &= \E_{\substack{(s_h, a_h, \ldots, s_{h+n}) \sim p^\gD}}
    \left[ D\left(Q^\ell_{\theta_q}(s_h, a_h, s_{h+n}), r(s_h, s_{h+n}) + \tilde\gamma V^\ell_{\theta_v}(s_{h+1}, s_{h+n}) \right) \right],
\end{align}
where $\tilde \gamma$ is the low-level discount factor defined as $\tilde \gamma = 1 - 1/n$.
At test time, double SHARSA defines the distribution of the low-level policy $\pi^\ell_{\theta_\ell}(\pl{a} \mid \pl{s}, \pl{w}): \gS \times \gG \to \Delta(\gA)$ with rejection sampling:
\begin{align}
    \pi^\ell_{\theta_\ell}(s, w) \stackrel{d}{=} \argmax_{a_1, \ldots, a_N: a_i \sim \pi_{\beta, \theta_\ell}^\ell(\pl{a} \mid s, w)} Q^\ell_{\theta_q}(s, a_i, w).
\end{align}
We provide the pseudocode in \Cref{alg:double_sharsa}.

\begin{algorithm}[t!]
\caption{Double SHARSA}
\label{alg:double_sharsa}
\begin{algorithmic}
\footnotesize

\LComment{\color{myblue} Training loop}
\While{not converged}
\State Sample batch $\{(s_h, a_h, \ldots, s_{h+n}, g)\}$ from $\gD$
\BeginBox[fill=white]
\LComment{Hierarchical flow BC}
\State Update high-level flow BC policy $\pi_\beta^h(\pl{s_{h+n}} \mid \pl{s_h}, \pl{g})$ with flow-matching loss (\Cref{eq:sharsa_pi_high})
\State Update low-level flow BC policy $\pi_\beta^\ell(\pl{a_h} \mid \pl{s_h}, \pl{s_{h+n}})$ with flow-matching loss (\Cref{eq:sharsa_pi_low})
\EndBox
\BeginBox[fill=white]
\LComment{High-level ($n$-step) SARSA value learning}
\State Update $V^h$ to minimize $\E\left[D\left(V^h(s_h, g), \bar Q^h(s_h, s_{h+n}, g)\right)\right]$
\State Update $Q^h$ to minimize $\E\left[D\left(Q^h(s_h, s_{h+n}, g), \sum_{i=0}^{n-1} \gamma^i r(s_{h+i}, g) + \gamma^n V^h(s_{h+n}, g)\right)\right]$
\EndBox
\BeginBox[fill=white]
\LComment{Low-level SARSA value learning}
\State Update $V^\ell$ to minimize $\E\left[D\left(V^\ell(s_h, s_{h+n}), \bar Q^\ell(s_h, a_h, s_{h+n})\right)\right]$
\State Update $Q^\ell$ to minimize $\E\left[D\left(Q^\ell(s_h, a_h, s_{h+n}), r(s_h, s_{h+n}) + \gamma V^\ell(s_{h+1}, s_{h+n})\right)\right]$
\EndBox
\EndWhile
\Return $\pi(\pl{s}, \pl{g})$ defined below

\vspace{5pt}

\LComment{\color{myblue} Resulting policy}
\Function{$\pi(s, g)$}{}
\BeginBox[fill=white]
\LComment{High-level: rejection sampling}
\State Sample $w_1, \ldots, w_N \sim \pi_\beta^h(s, g)$
\State Set $w \gets \argmax_{w_1, \ldots, w_N} Q^h(s, w_i, g)$
\EndBox
\BeginBox[fill=white]
\LComment{Low-level: rejection sampling} 
\State Sample $a_1, \ldots, a_N \sim \pi_\beta^\ell(s, w)$\vskip 30pt
\State Set $a \gets \argmax_{a_1, \ldots, a_N} Q^\ell(s, a_i, w)$
\EndBox
\State \Return $a$
\EndFunction

\end{algorithmic}
\end{algorithm}

\clearpage

\section{Experimental details}
\label{sec:exp_details}

We implement all methods used in this work on top of the reference implementations of OGBench~\citep{ogbench_park2025}.
Each run in this work takes no more than three days on a single A5000 GPU.
We provide our implementations and datasets at \url{https://github.com/seohongpark/horizon-reduction}.

\subsection{Didactic experiments}
\label{sec:exp_details_didactic}

In this section, we describe additional experimental details for the experiments in \Cref{sec:horizon_value}.

\textbf{Task.}
The \tt{combination-lock} task consists of $H$ states numbered from $0$ to $H-1$ and two discrete actions.
Each state is represented by $\lceil \log_2 H \rceil$-dimensional binary vector.
The ordering of the states is randomly determined by a fixed random seed,
which ensures that all runs in our experiments share the same environment dynamics.

\textbf{Algorithms.}
We consider $1$-step DQN and $n$-step DQN (with $n=64$) in \Cref{sec:horizon_value}.
The value losses for $1$-step DQN and $n$-step DQN are as follows:
\begin{align}
L_{1\mathrm{-step}}(\theta) &= \E %
\left[\left(Q_\theta(s_h, a_h) - r(s_h, a_h) - \max_{a_{h+1} \in \gA} Q_{\bar \theta}(s_{h+1}, a_{h+1}) \right)^2\right], \label{eq:1dqn} \\
L_{n\mathrm{-step}}(\theta) &= \E
\left[\left(Q_\theta(s_h, a_h) - \sum_{i=0}^{n-1} r(s_{h+i}, a_{h+i}) - \max_{a_{h+n} \in \gA} Q_{\bar \theta}(s_{h+n}, a_{h+n}) \right)^2\right], \label{eq:ndqn}
\end{align}
where the expectations are taken over consecutive state-action trajectories
uniformly sampled from the dataset.
We also employ double Q-learning~\citep{ddqn_hasselt2016} to stabilize training.

\begin{wrapfigure}{r}{0.38\textwidth}
    \centering
    \vspace{-3pt}
    \raisebox{0pt}[\dimexpr\height-1.0\baselineskip\relax]{
        \begin{subfigure}[t]{1.0\linewidth}
        \includegraphics[width=\linewidth]{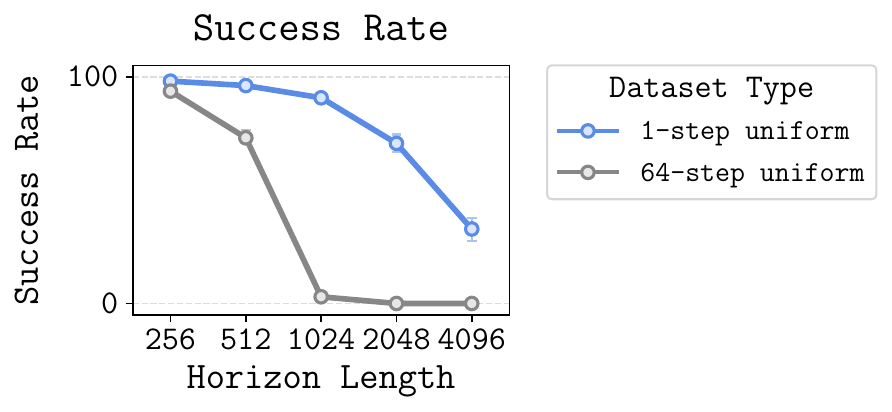}
        \end{subfigure}
    }
    \vspace{-15pt}
    \caption{
    \footnotesize
    \textbf{$1$-step DQN on two datasets.}
    }
    \vspace{-7pt}
    \label{fig:toy_data_abl}
\end{wrapfigure}
\textbf{Datasets.}
We generate two types of datasets.
The first is a $1$-step uniform-coverage dataset, collected by sampling state-action pairs uniformly from all possible $2H$ tuples.
The second is a $64$-step uniform-coverage dataset, collected by the following procedure:
first sample a state uniformly from $H$ states, and then perform either $64$ consecutive correct actions (with probability $0.5$)
or $64$ consecutive incorrect actions (with probability $0.5$).
Note that the former provides uniform state-action coverage for $1$-step DQN,
and the latter provides uniform state-action coverage for $64$-step DQN.
We use the $1$-step uniform dataset for $1$-step DQN and the $64$-step uniform dataset for $64$-step DQN.
Since $1$-step DQN works worse on the $64$-step uniform dataset (\Cref{fig:toy_data_abl})
and $64$-step DQN is incompatible with the $1$-step uniform dataset,
this setup provides a fair comparison of the maximum possible performance of the two algorithms,
with the dataset factor marginalized out.

\textbf{Metrics.}
We train each agent for $5$M gradient steps and evaluate every $100$K steps.
We measure three metrics: success rate, TD error, and Q error.
The success rate is measured by rolling out the deterministic policy induced by the learned Q function, averaged over all evaluation epochs.
The TD error is measured by the critic loss (\Cref{eq:1dqn,eq:ndqn}),
averaged over steps on and after $4$M.
The Q error is measured by the difference between the predicted Q values and the ground-truth Q values (\ie, the negative of the remaining steps to the goal),
evaluated at the final epoch.

We provide the full list of hyperparameters in \Cref{table:hyp_didactic}.

\subsection{OGBench experiments}
\label{sec:exp_details_ogbench}

\begin{figure}[h!]
    \begin{minipage}{\linewidth}
        \centering
        \begin{minipage}{0.18\linewidth}
            \centering
            \includegraphics[width=\linewidth]{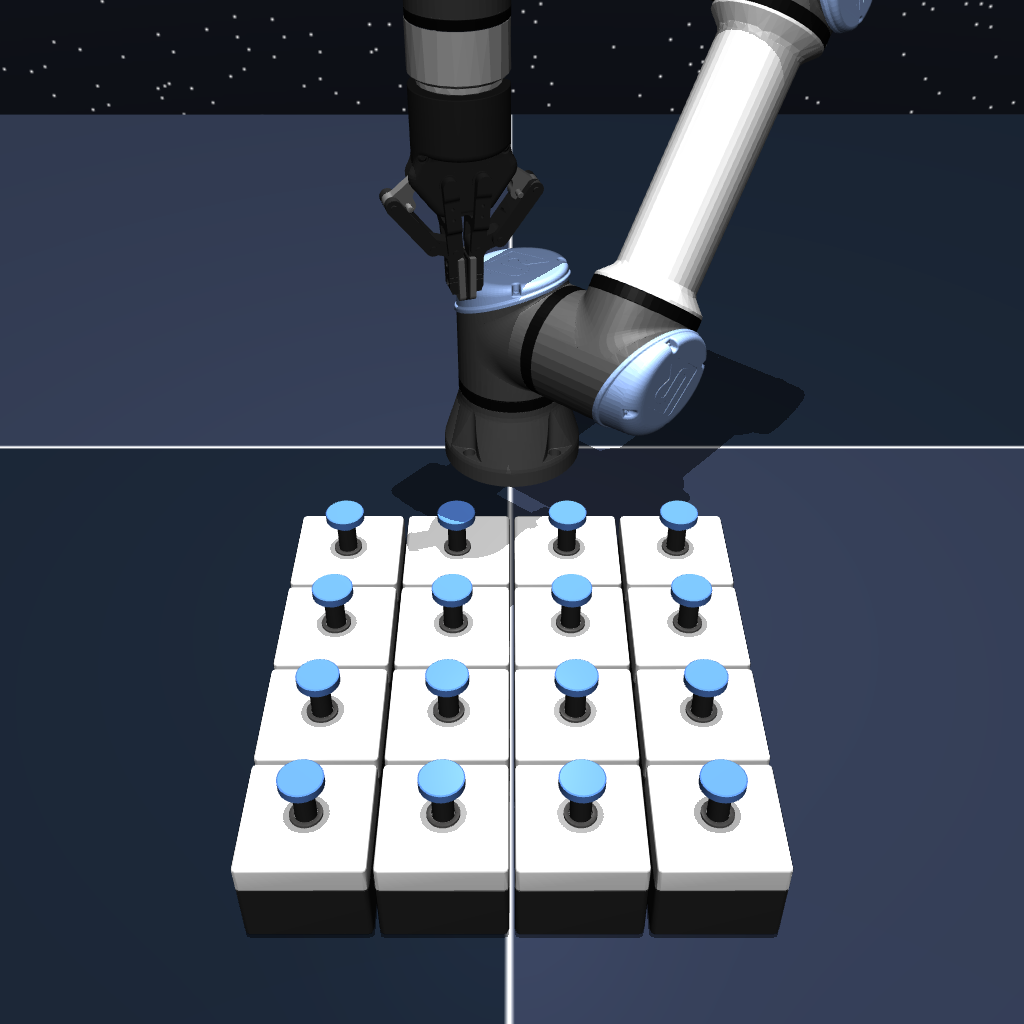}
            \begin{tikzpicture}
                \node at (0, 0.58) {};
                \draw[Triangle-, thick] (0,0.5) -- (0,0);
                \node at (0, -0.1) {};
            \end{tikzpicture}
            \includegraphics[width=\linewidth]{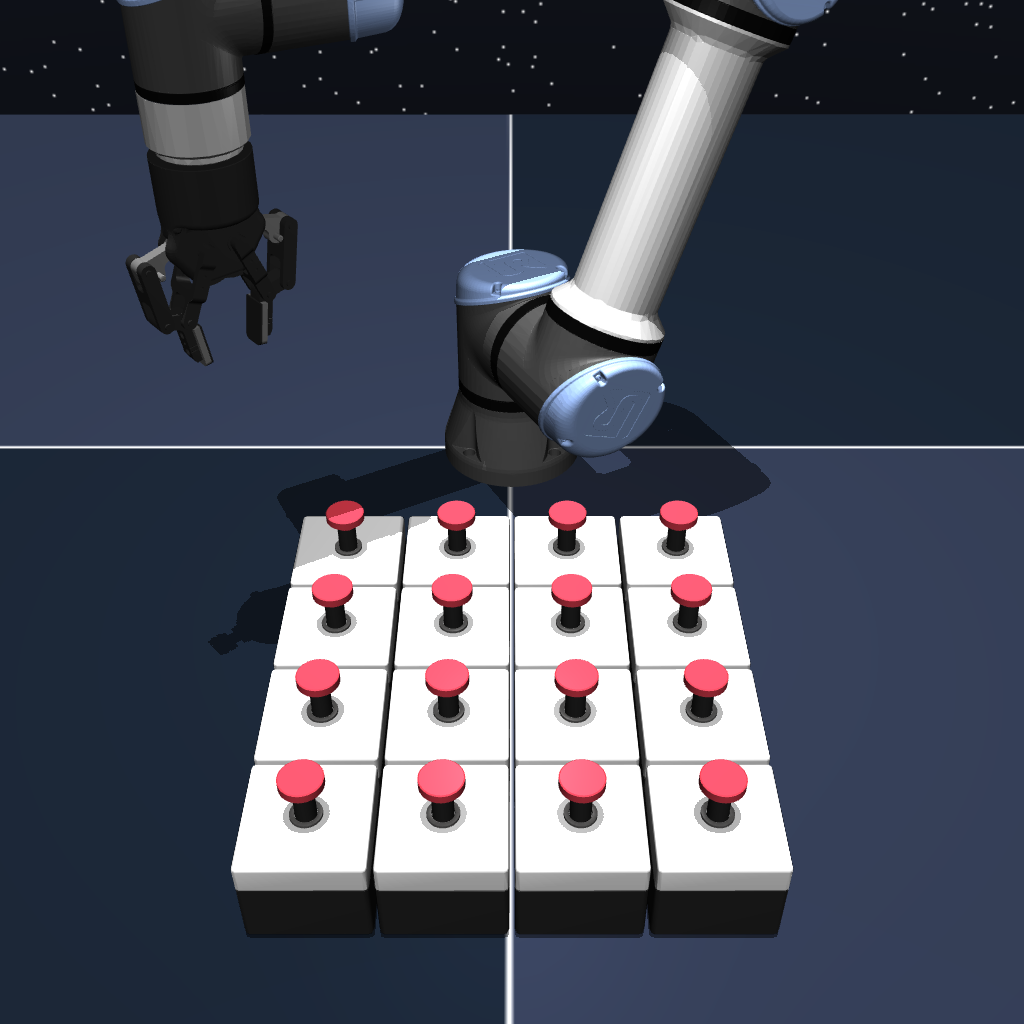}
            \vspace{-15pt}
            \captionsetup{font={stretch=0.7}}
            \caption*{\centering \texttt{task1}}
        \end{minipage}
        \hfill
        \begin{minipage}{0.18\linewidth}
            \centering
            \includegraphics[width=\linewidth]{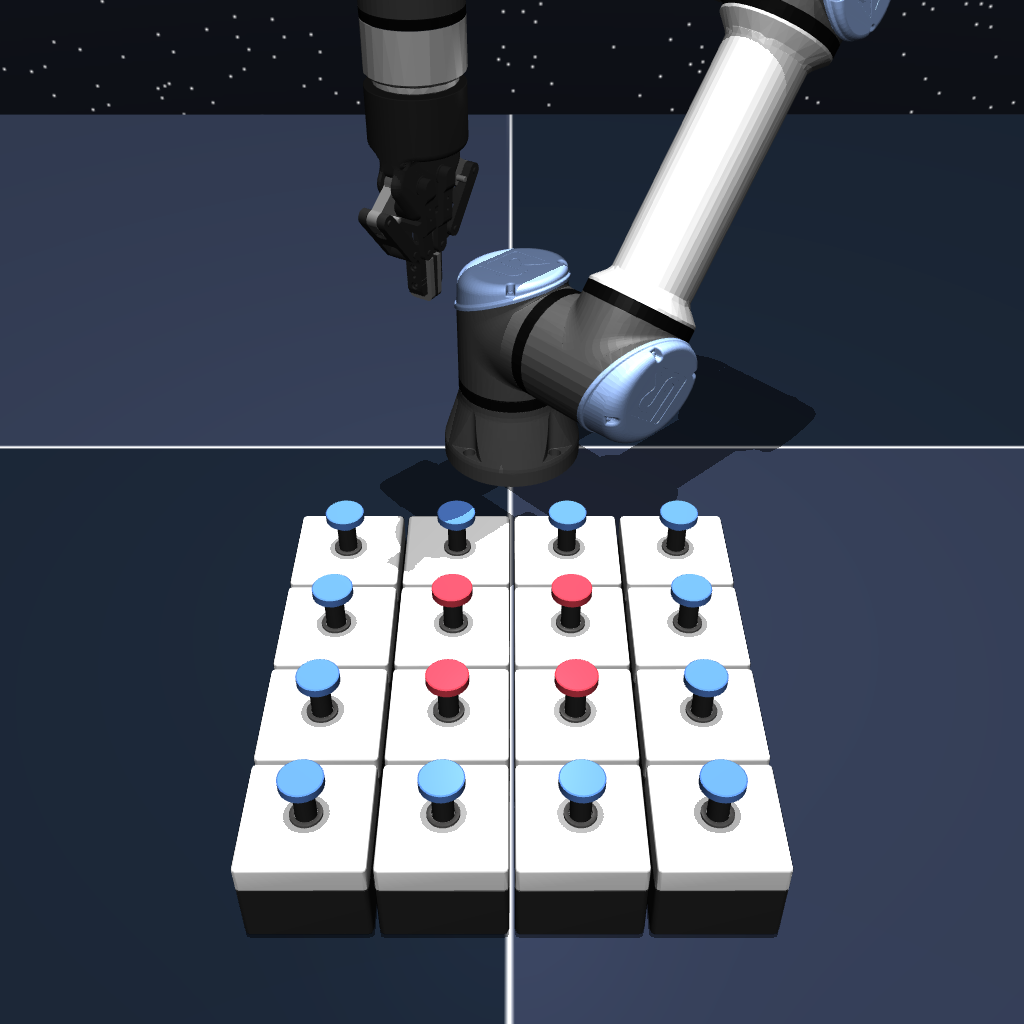}
            \begin{tikzpicture}
                \node at (0, 0.58) {};
                \draw[Triangle-, thick] (0,0.5) -- (0,0);
                \node at (0, -0.1) {};
            \end{tikzpicture}
            \includegraphics[width=\linewidth]{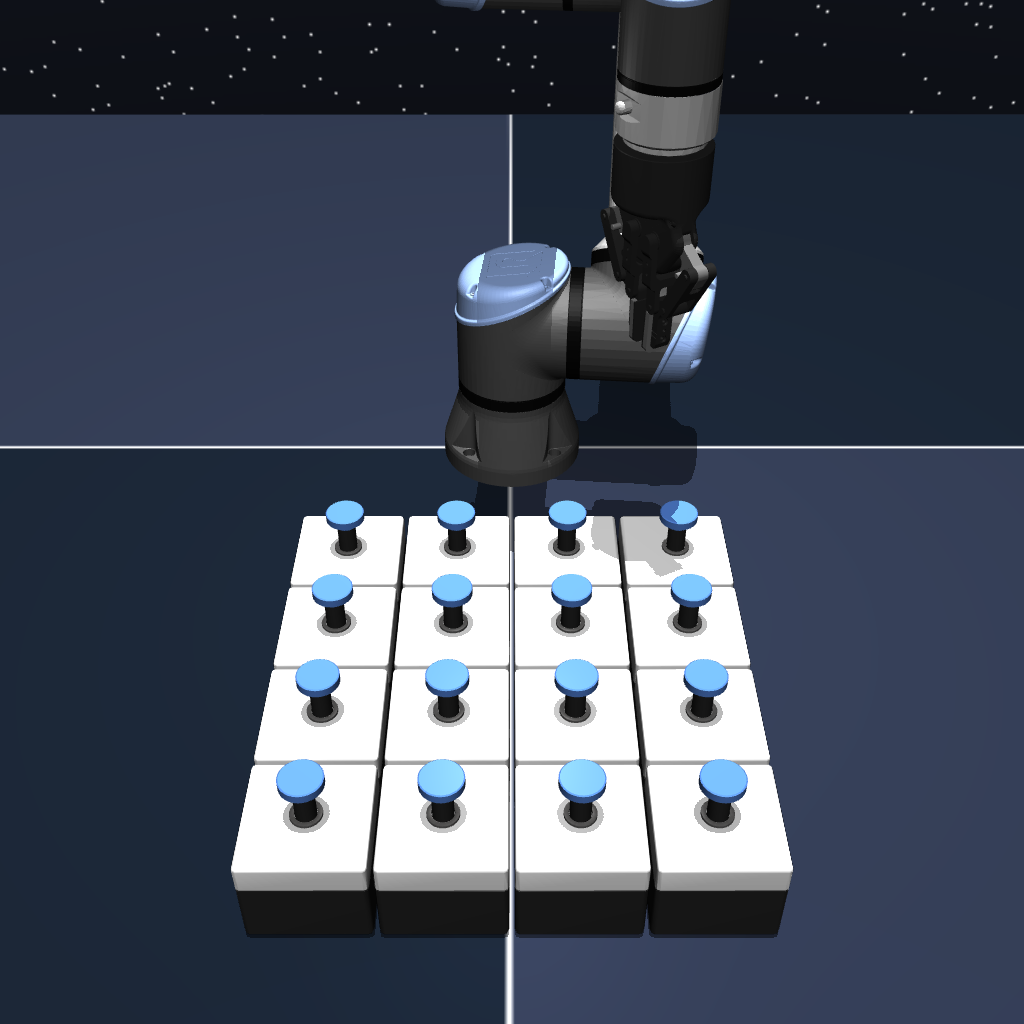}
            \vspace{-15pt}
            \captionsetup{font={stretch=0.7}}
            \caption*{\centering \texttt{task2}}
        \end{minipage}
        \hfill
        \begin{minipage}{0.18\linewidth}
            \centering
            \includegraphics[width=\linewidth]{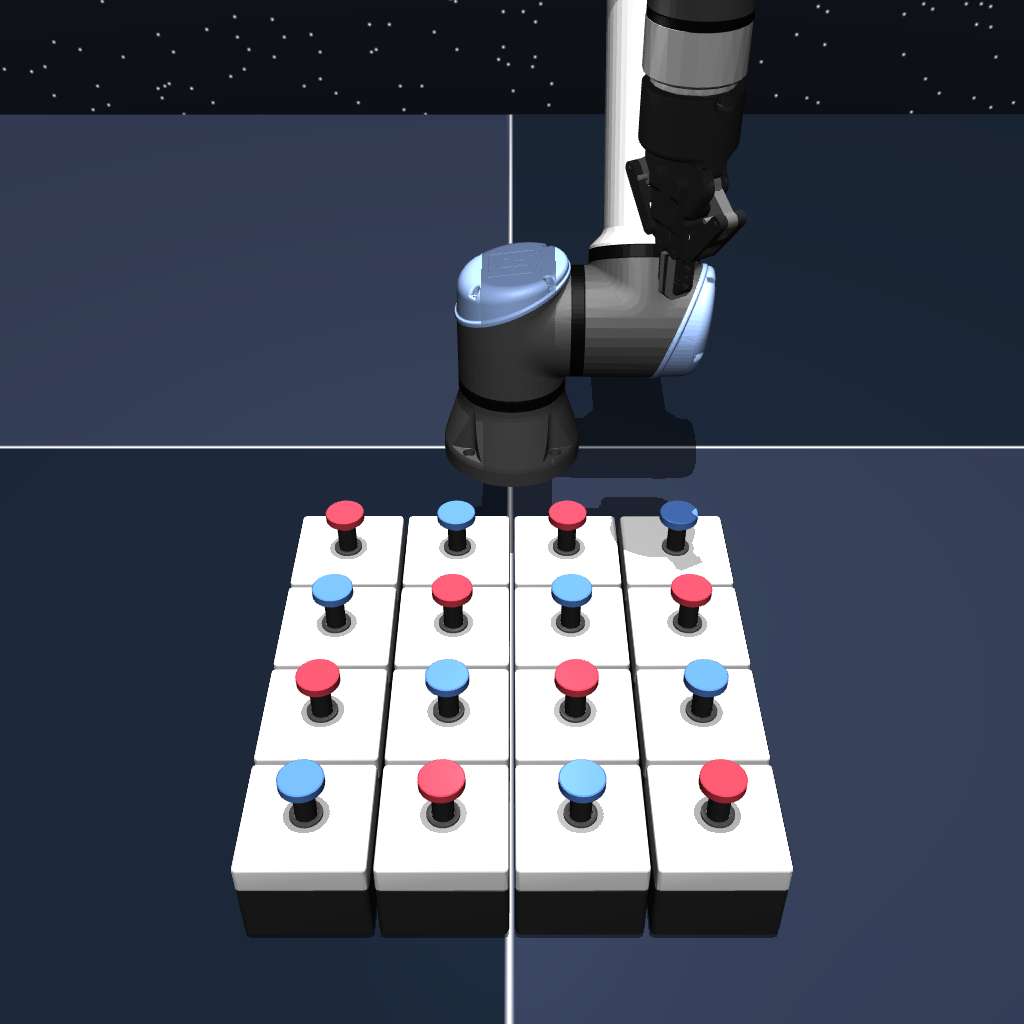}
            \begin{tikzpicture}
                \node at (0, 0.58) {};
                \draw[Triangle-, thick] (0,0.5) -- (0,0);
                \node at (0, -0.1) {};
            \end{tikzpicture}
            \includegraphics[width=\linewidth]{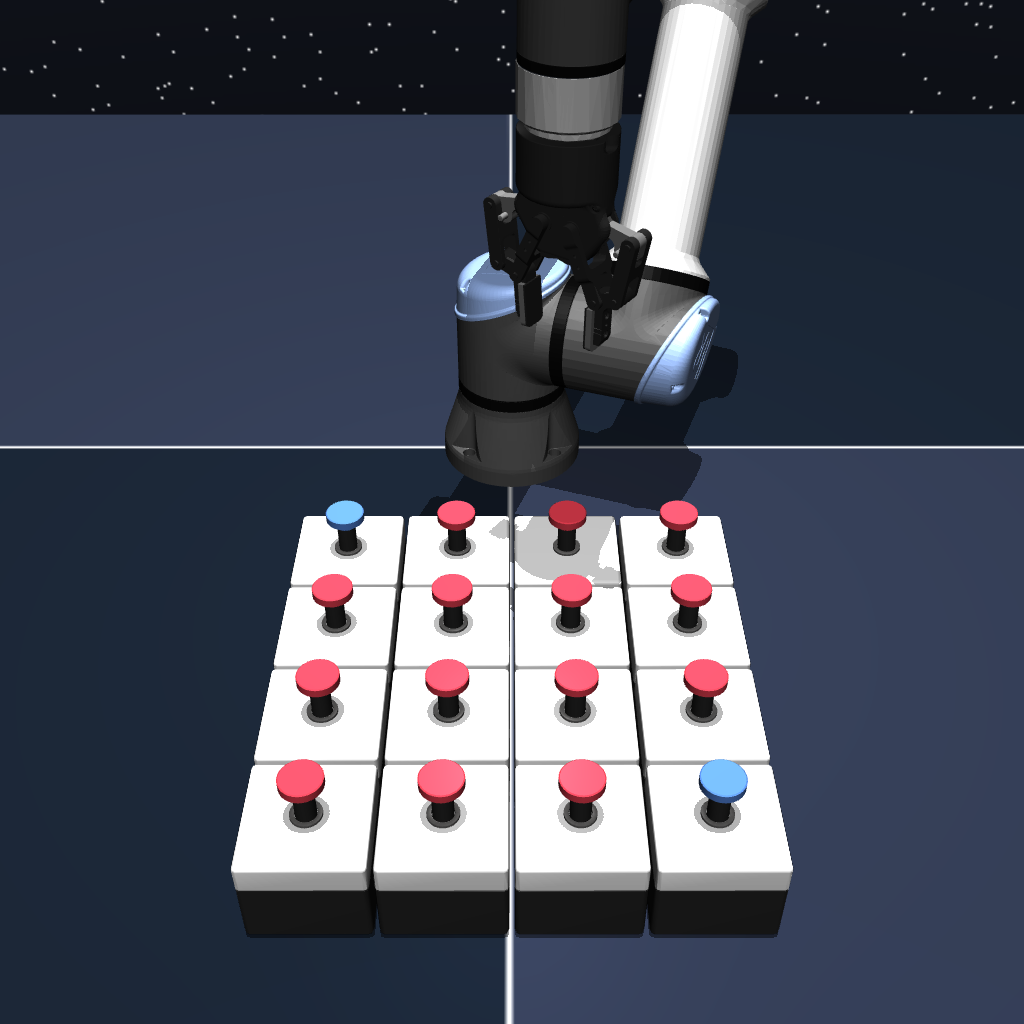}
            \vspace{-15pt}
            \captionsetup{font={stretch=0.7}}
            \caption*{\centering \texttt{task3}}
        \end{minipage}
        \hfill
        \begin{minipage}{0.18\linewidth}
            \centering
            \includegraphics[width=\linewidth]{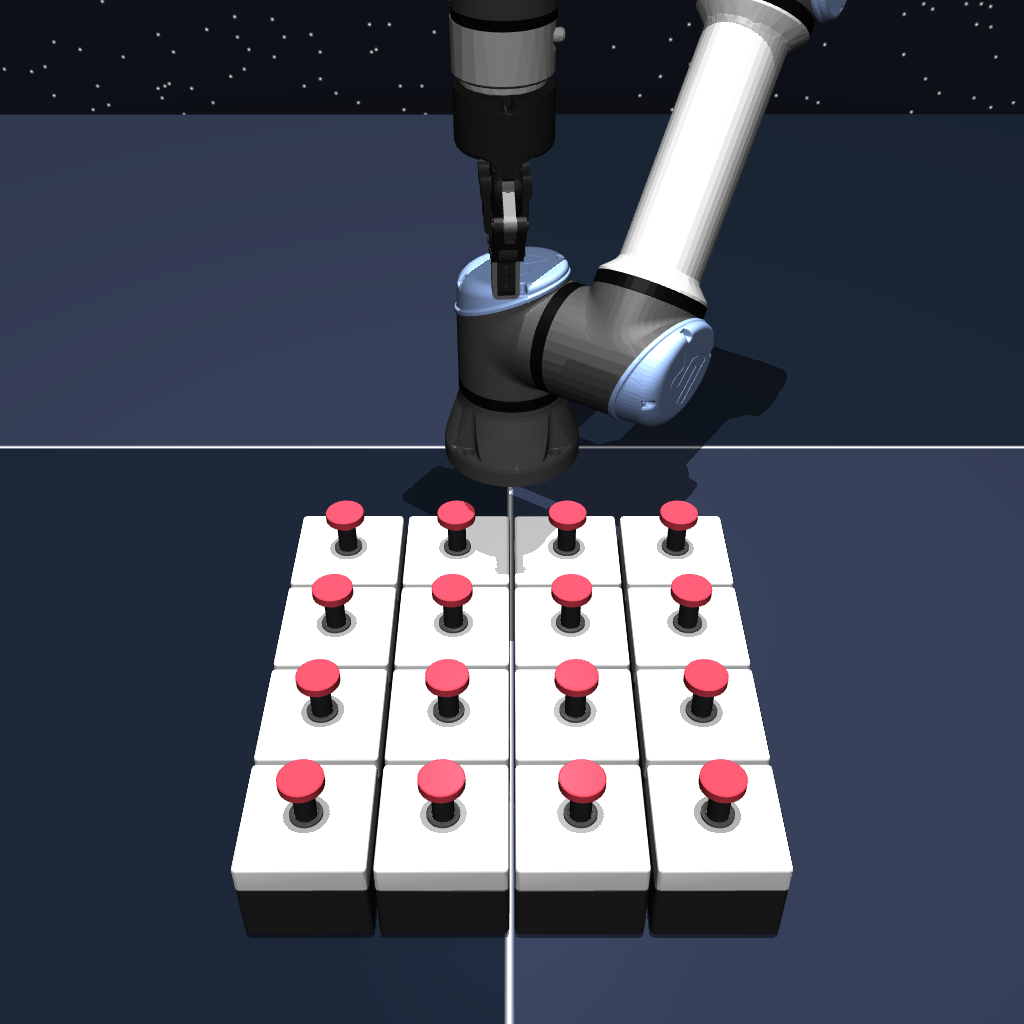}
            \begin{tikzpicture}
                \node at (0, 0.58) {};
                \draw[Triangle-, thick] (0,0.5) -- (0,0);
                \node at (0, -0.1) {};
            \end{tikzpicture}
            \includegraphics[width=\linewidth]{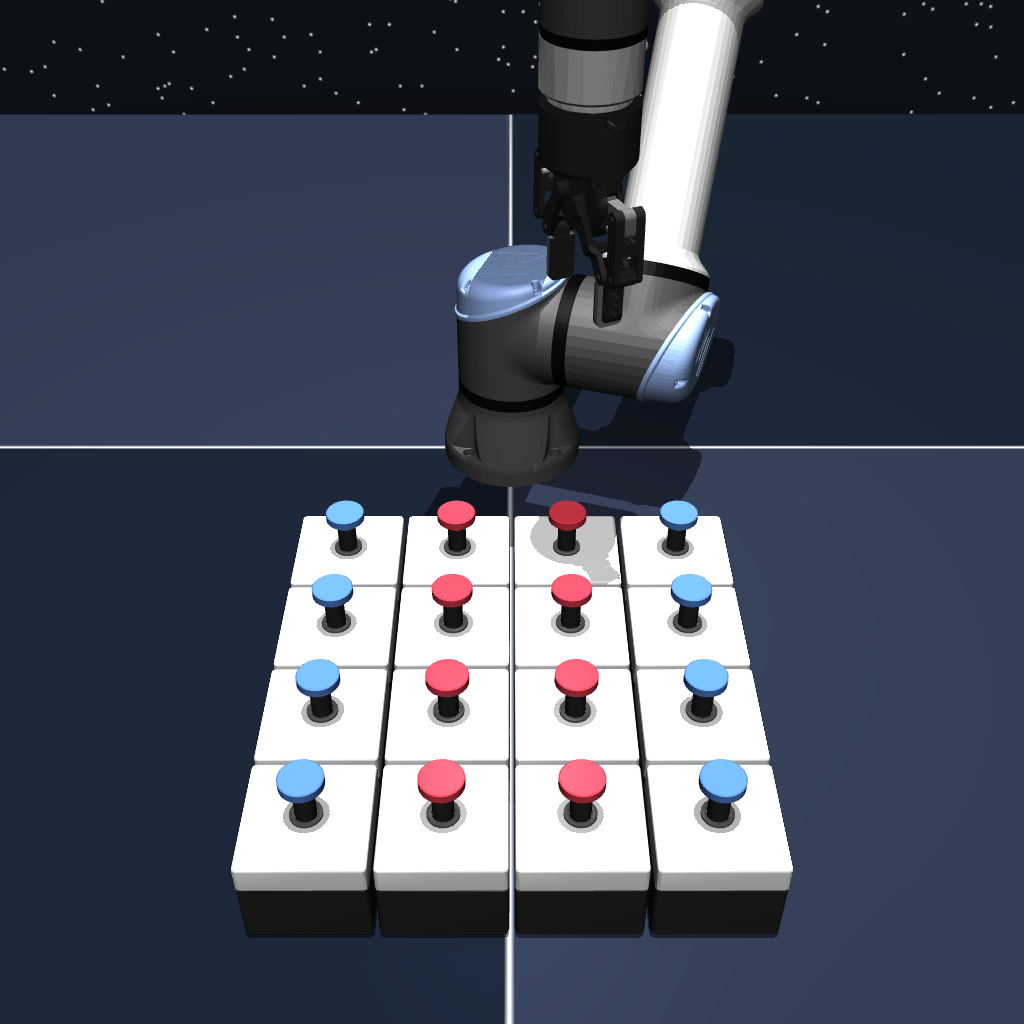}
            \vspace{-15pt}
            \captionsetup{font={stretch=0.7}}
            \caption*{\centering \texttt{task4}}
        \end{minipage}
        \hfill
        \begin{minipage}{0.18\linewidth}
            \centering
            \includegraphics[width=\linewidth]{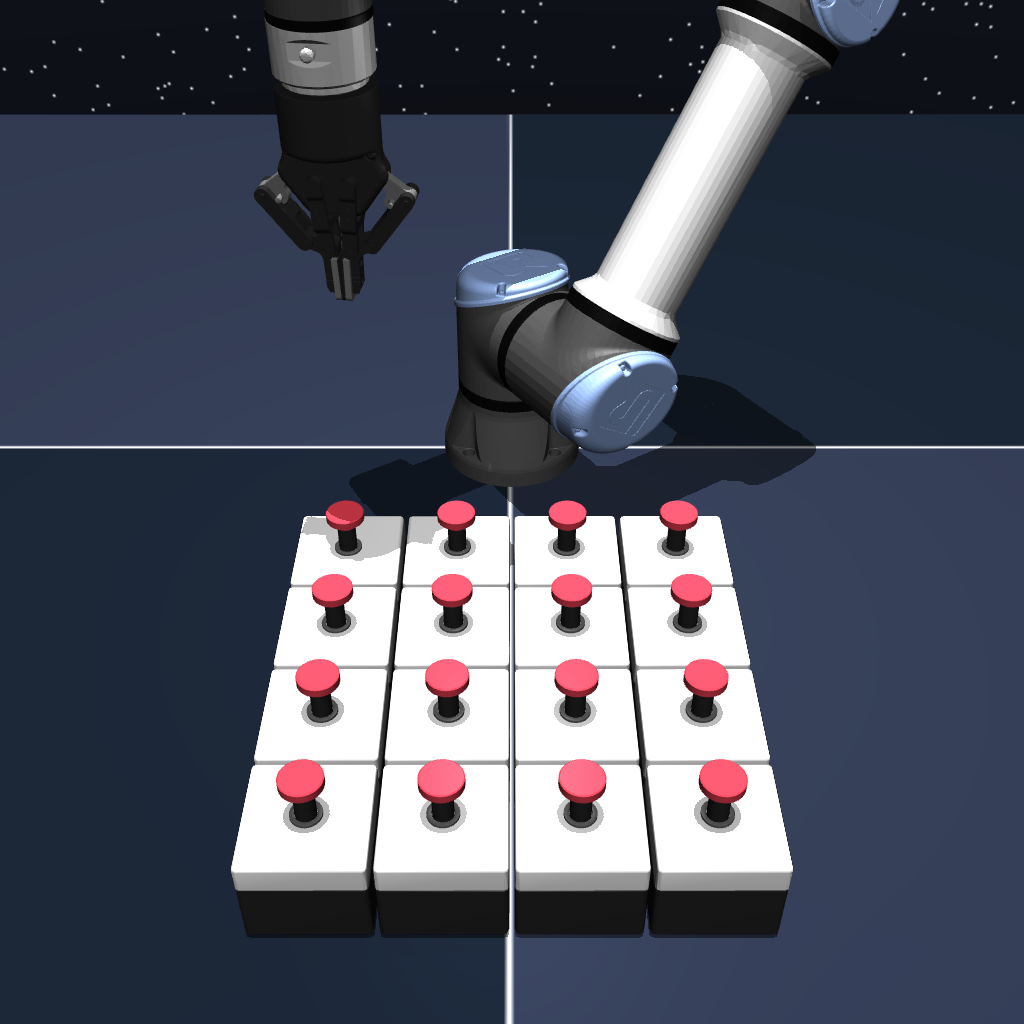}
            \begin{tikzpicture}
                \node at (0, 0.58) {};
                \draw[Triangle-, thick] (0,0.5) -- (0,0);
                \node at (0, -0.1) {};
            \end{tikzpicture}
            \includegraphics[width=\linewidth]{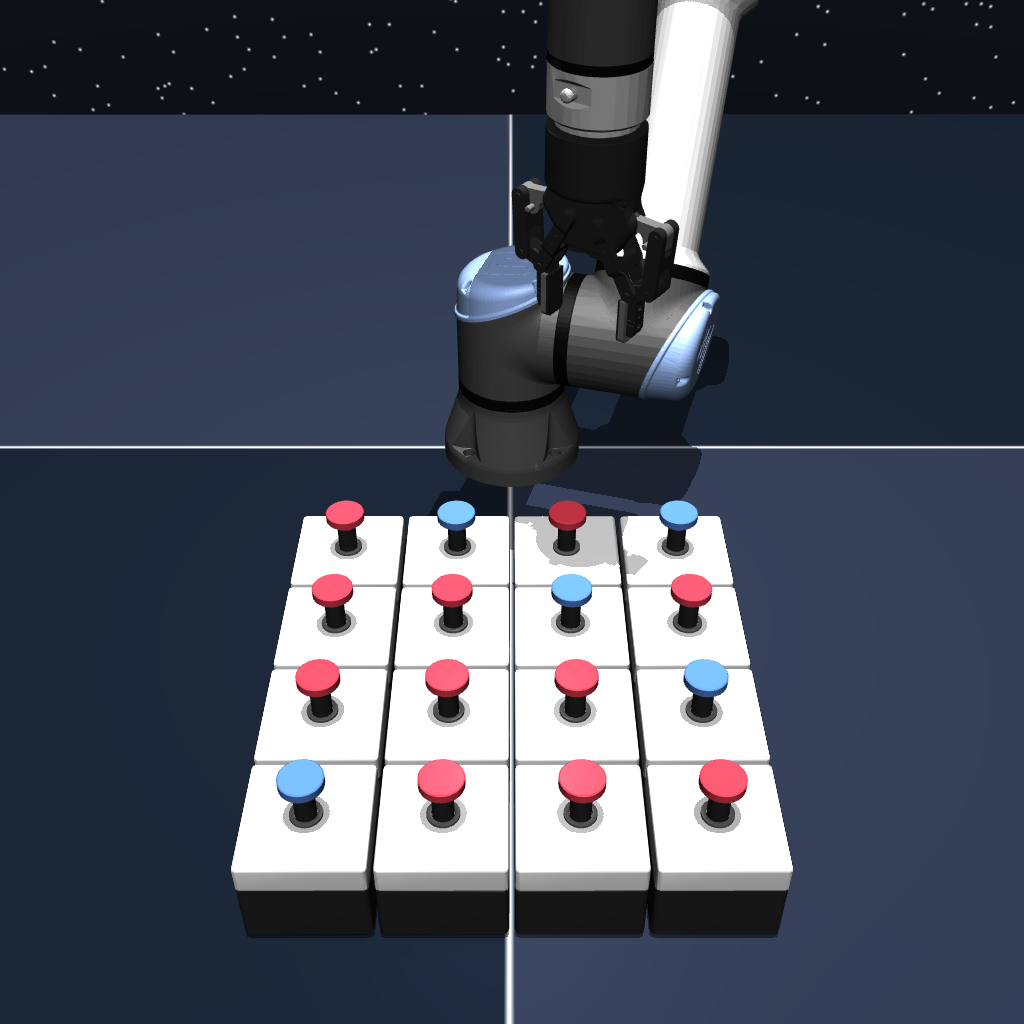}
            \vspace{-15pt}
            \captionsetup{font={stretch=0.7}}
            \caption*{\centering \texttt{task5}}
        \end{minipage}
    \end{minipage}
    \vspace{-3pt}
    \caption{\footnotesize \textbf{Evaluation goals for \tt{puzzle-4x4}.}}
    \label{fig:goals_1}
\end{figure}

\begin{figure}[h!]
    \begin{minipage}{\linewidth}
        \centering
        \begin{minipage}{0.18\linewidth}
            \centering
            \includegraphics[width=\linewidth]{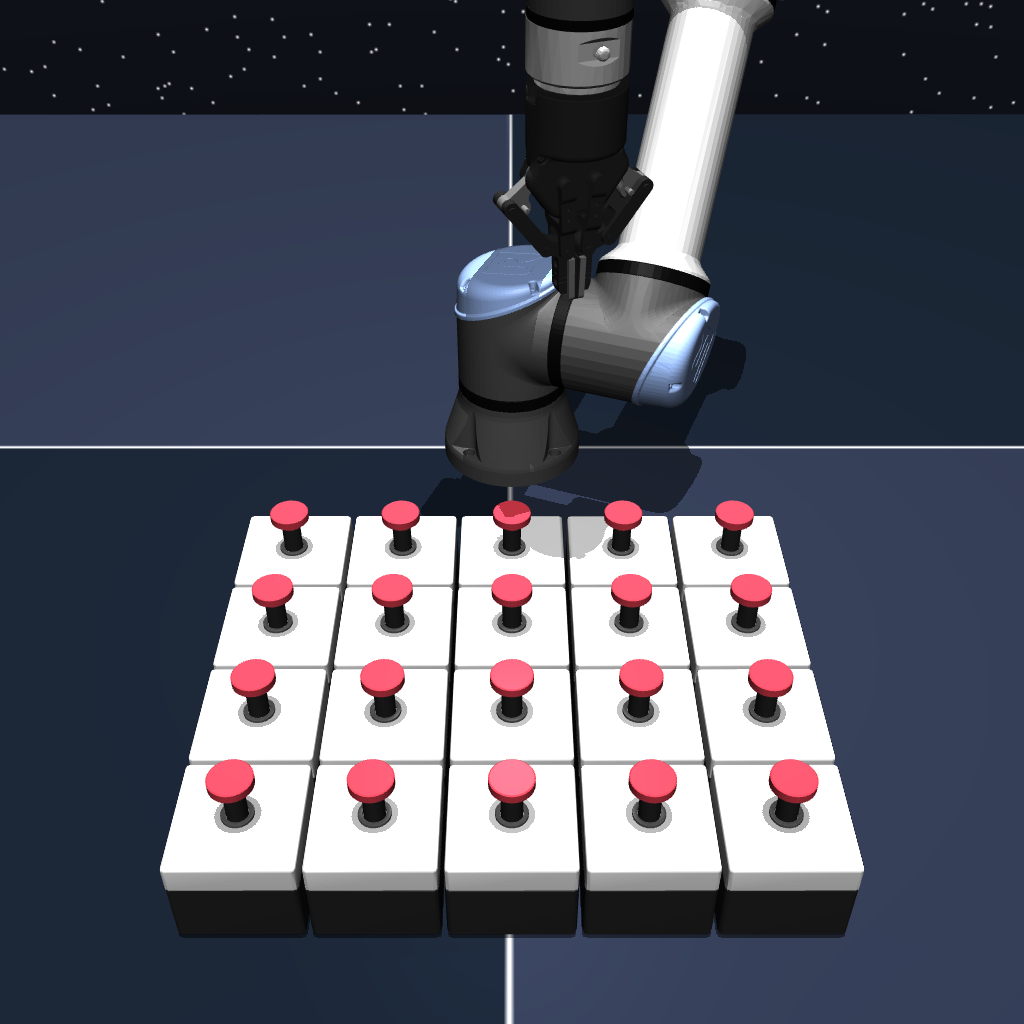}
            \begin{tikzpicture}
                \node at (0, 0.58) {};
                \draw[Triangle-, thick] (0,0.5) -- (0,0);
                \node at (0, -0.1) {};
            \end{tikzpicture}
            \includegraphics[width=\linewidth]{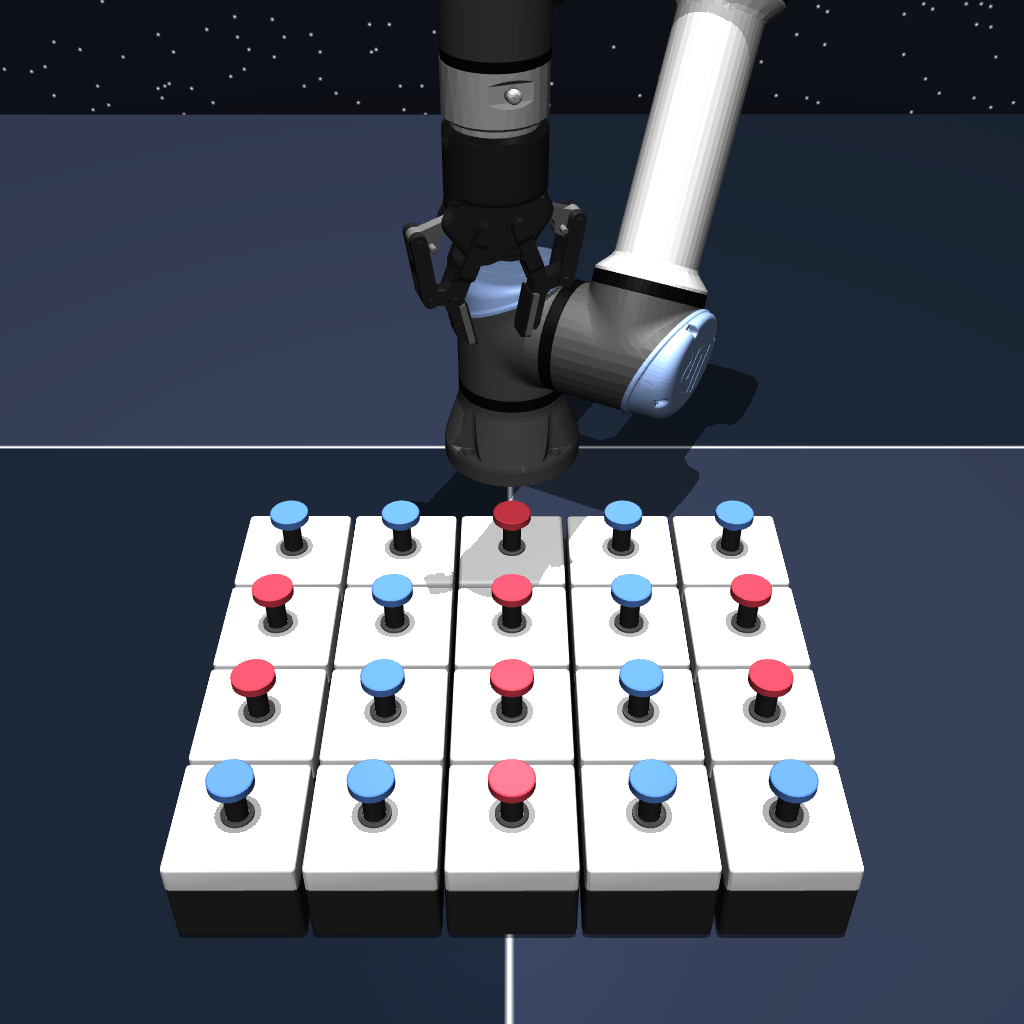}
            \vspace{-15pt}
            \captionsetup{font={stretch=0.7}}
            \caption*{\centering \texttt{task1}}
        \end{minipage}
        \hfill
        \begin{minipage}{0.18\linewidth}
            \centering
            \includegraphics[width=\linewidth]{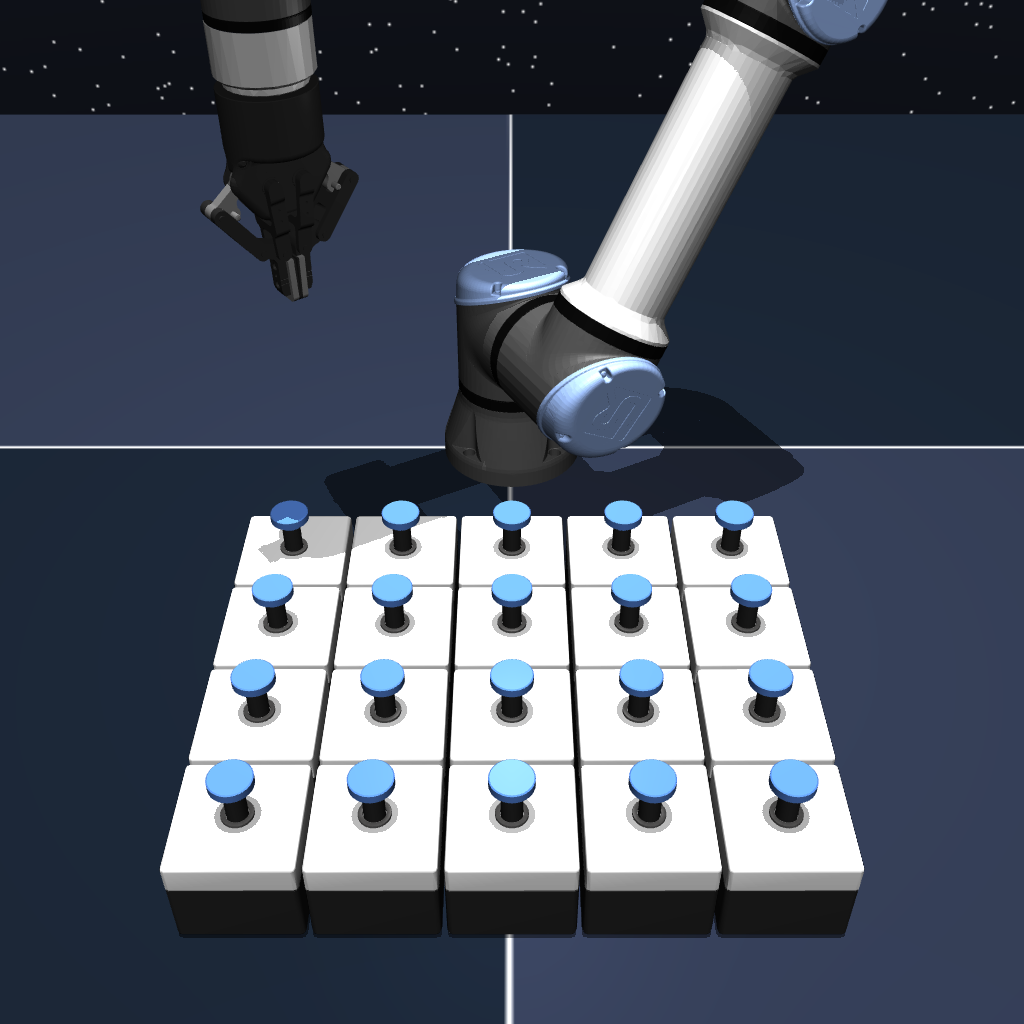}
            \begin{tikzpicture}
                \node at (0, 0.58) {};
                \draw[Triangle-, thick] (0,0.5) -- (0,0);
                \node at (0, -0.1) {};
            \end{tikzpicture}
            \includegraphics[width=\linewidth]{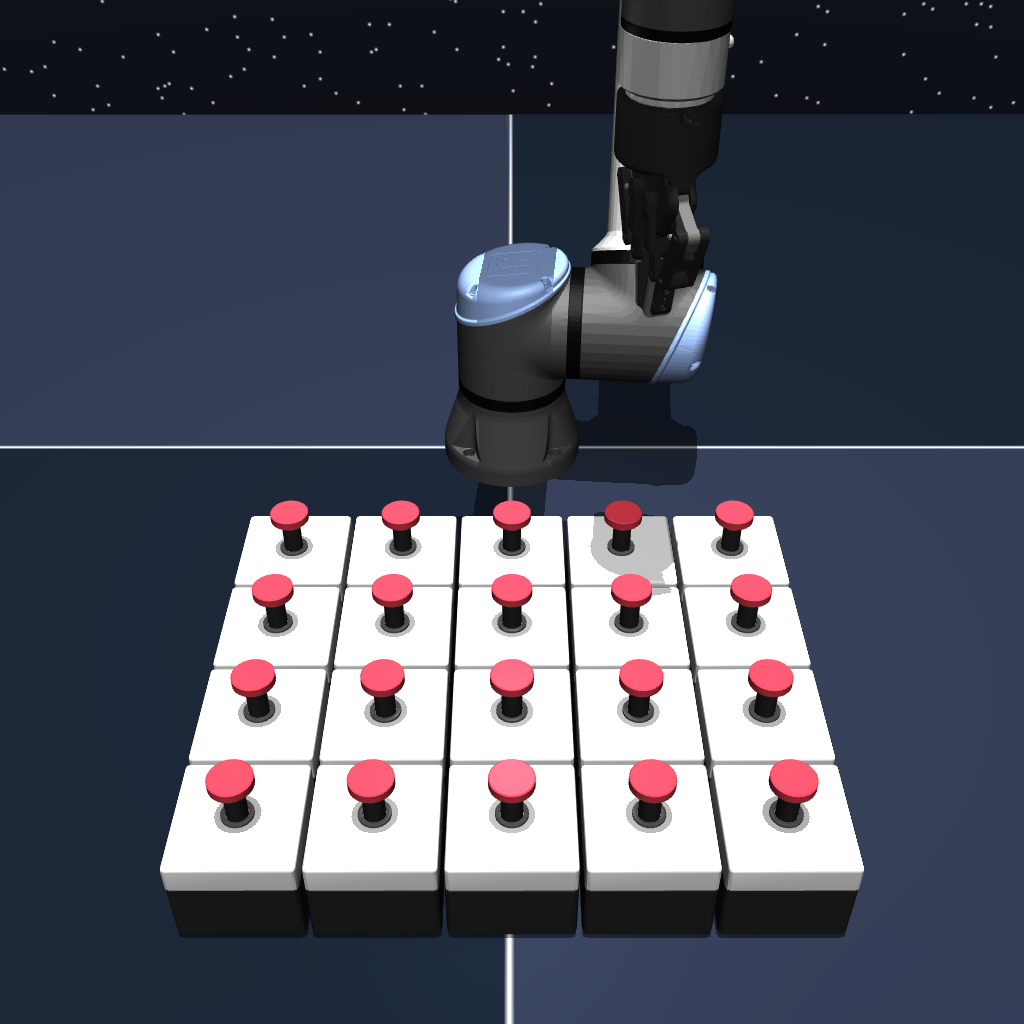}
            \vspace{-15pt}
            \captionsetup{font={stretch=0.7}}
            \caption*{\centering \texttt{task2}}
        \end{minipage}
        \hfill
        \begin{minipage}{0.18\linewidth}
            \centering
            \includegraphics[width=\linewidth]{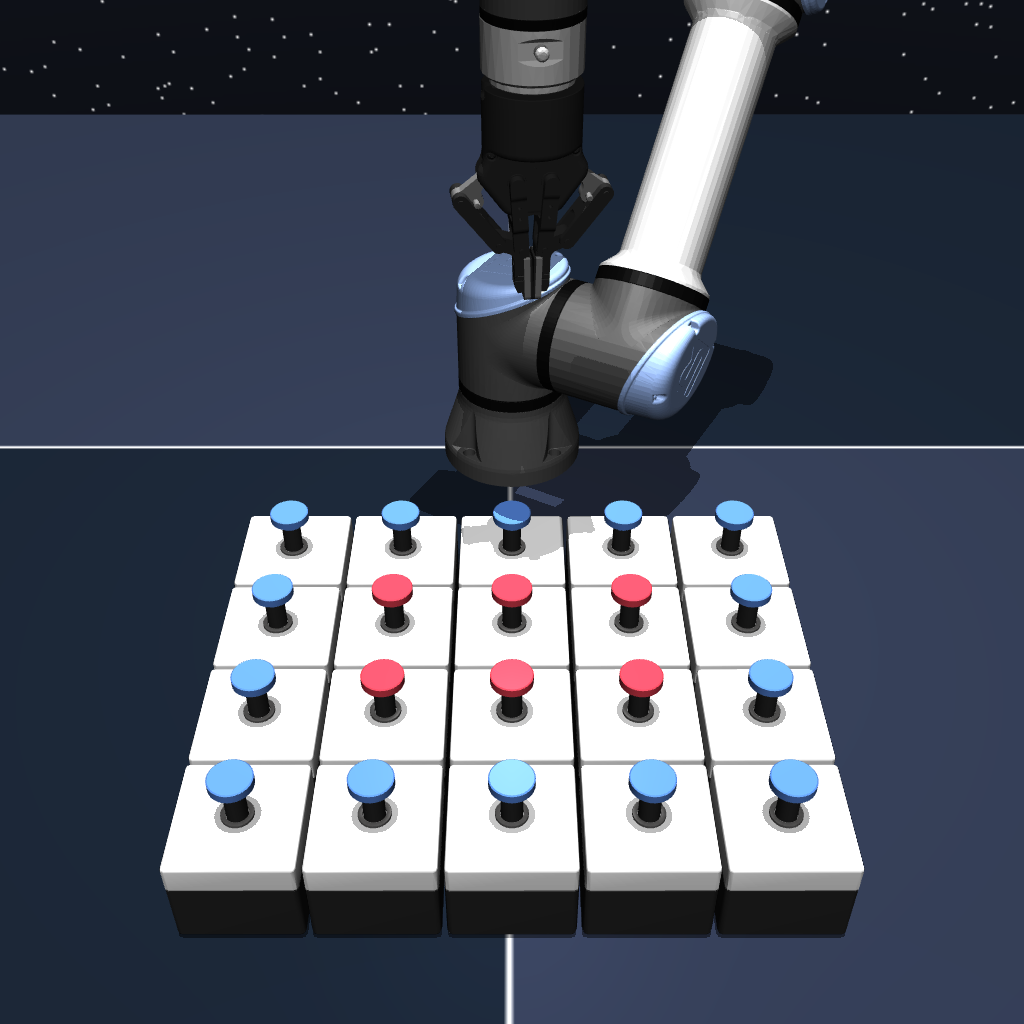}
            \begin{tikzpicture}
                \node at (0, 0.58) {};
                \draw[Triangle-, thick] (0,0.5) -- (0,0);
                \node at (0, -0.1) {};
            \end{tikzpicture}
            \includegraphics[width=\linewidth]{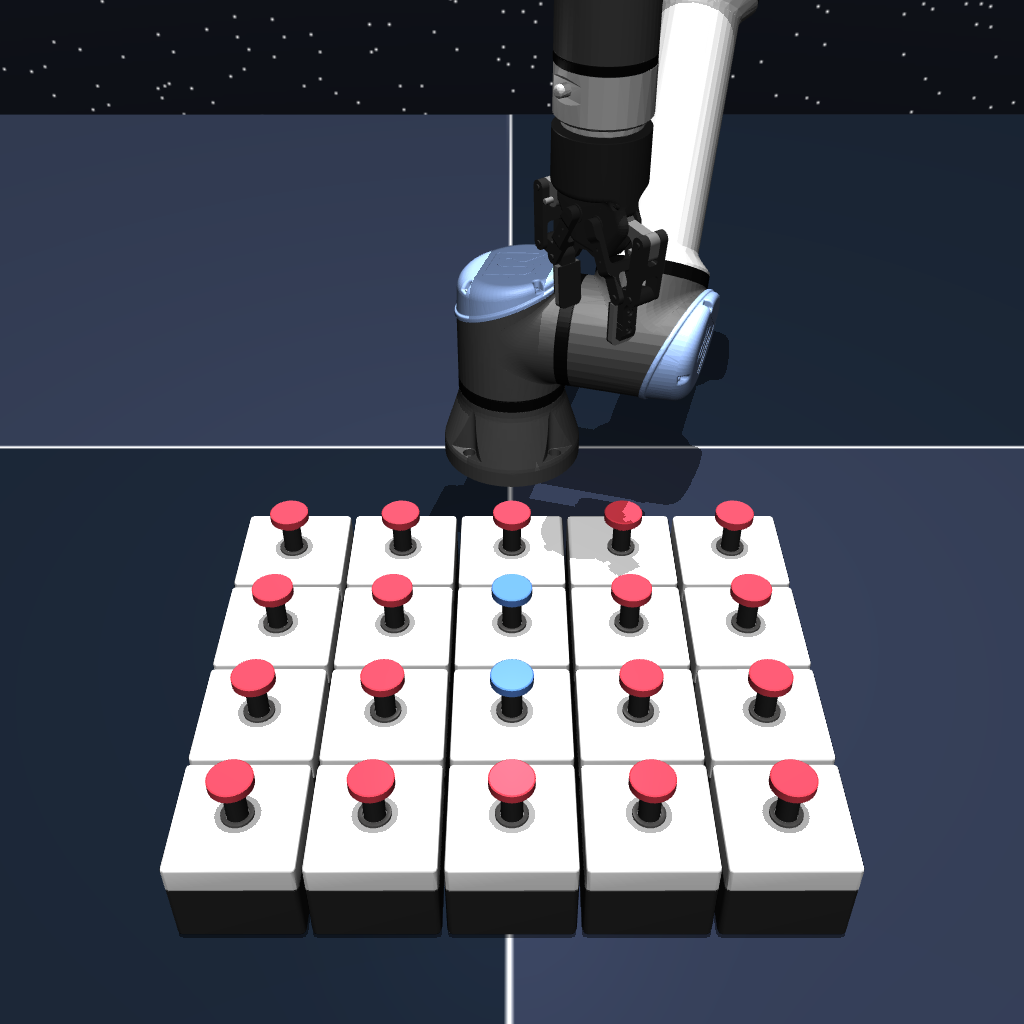}
            \vspace{-15pt}
            \captionsetup{font={stretch=0.7}}
            \caption*{\centering \texttt{task3}}
        \end{minipage}
        \hfill
        \begin{minipage}{0.18\linewidth}
            \centering
            \includegraphics[width=\linewidth]{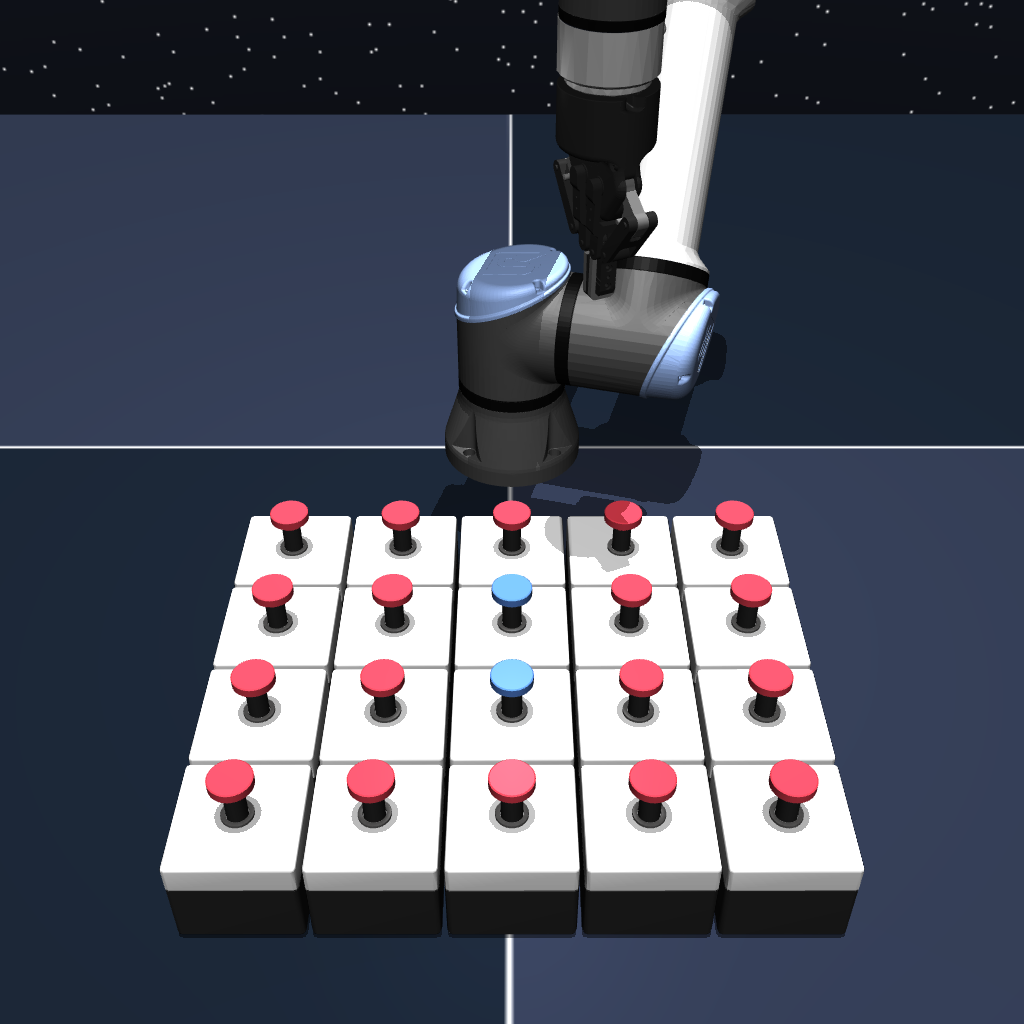}
            \begin{tikzpicture}
                \node at (0, 0.58) {};
                \draw[Triangle-, thick] (0,0.5) -- (0,0);
                \node at (0, -0.1) {};
            \end{tikzpicture}
            \includegraphics[width=\linewidth]{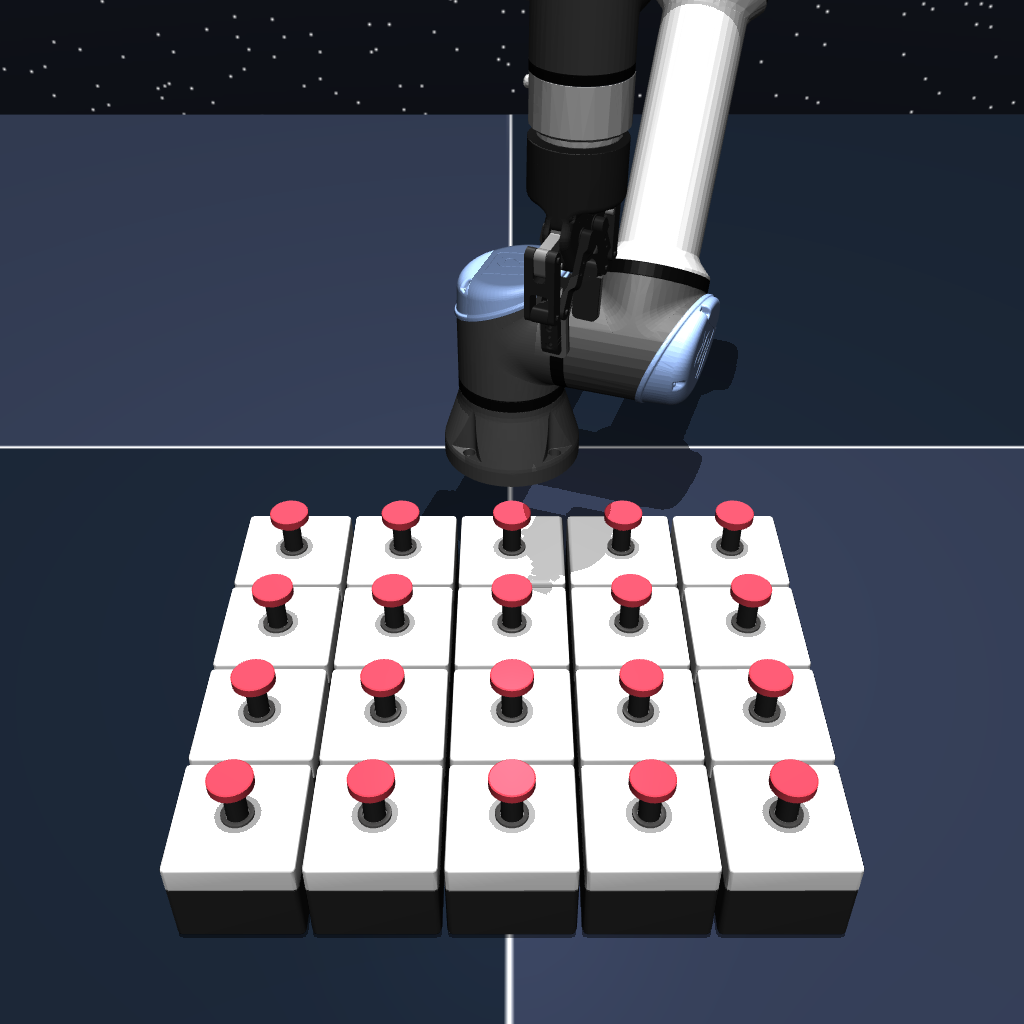}
            \vspace{-15pt}
            \captionsetup{font={stretch=0.7}}
            \caption*{\centering \texttt{task4}}
        \end{minipage}
        \hfill
        \begin{minipage}{0.18\linewidth}
            \centering
            \includegraphics[width=\linewidth]{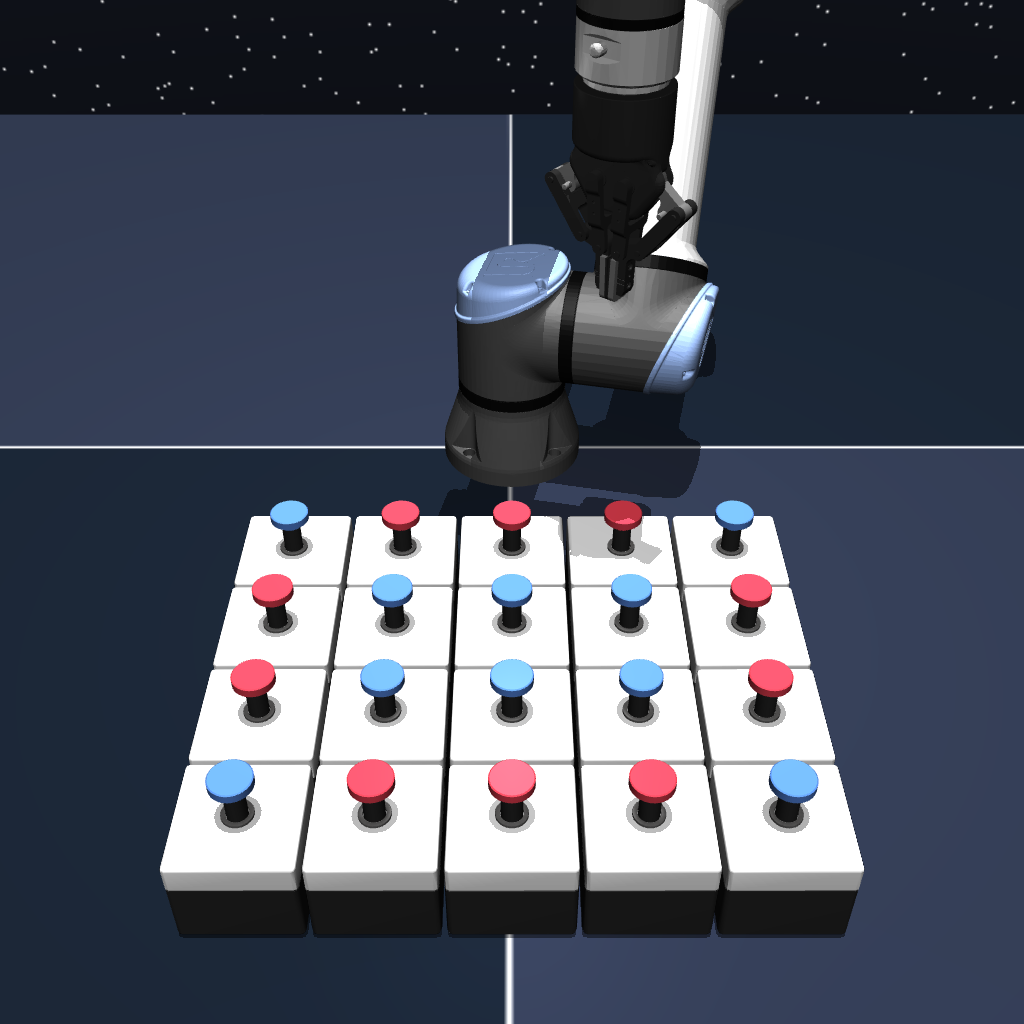}
            \begin{tikzpicture}
                \node at (0, 0.58) {};
                \draw[Triangle-, thick] (0,0.5) -- (0,0);
                \node at (0, -0.1) {};
            \end{tikzpicture}
            \includegraphics[width=\linewidth]{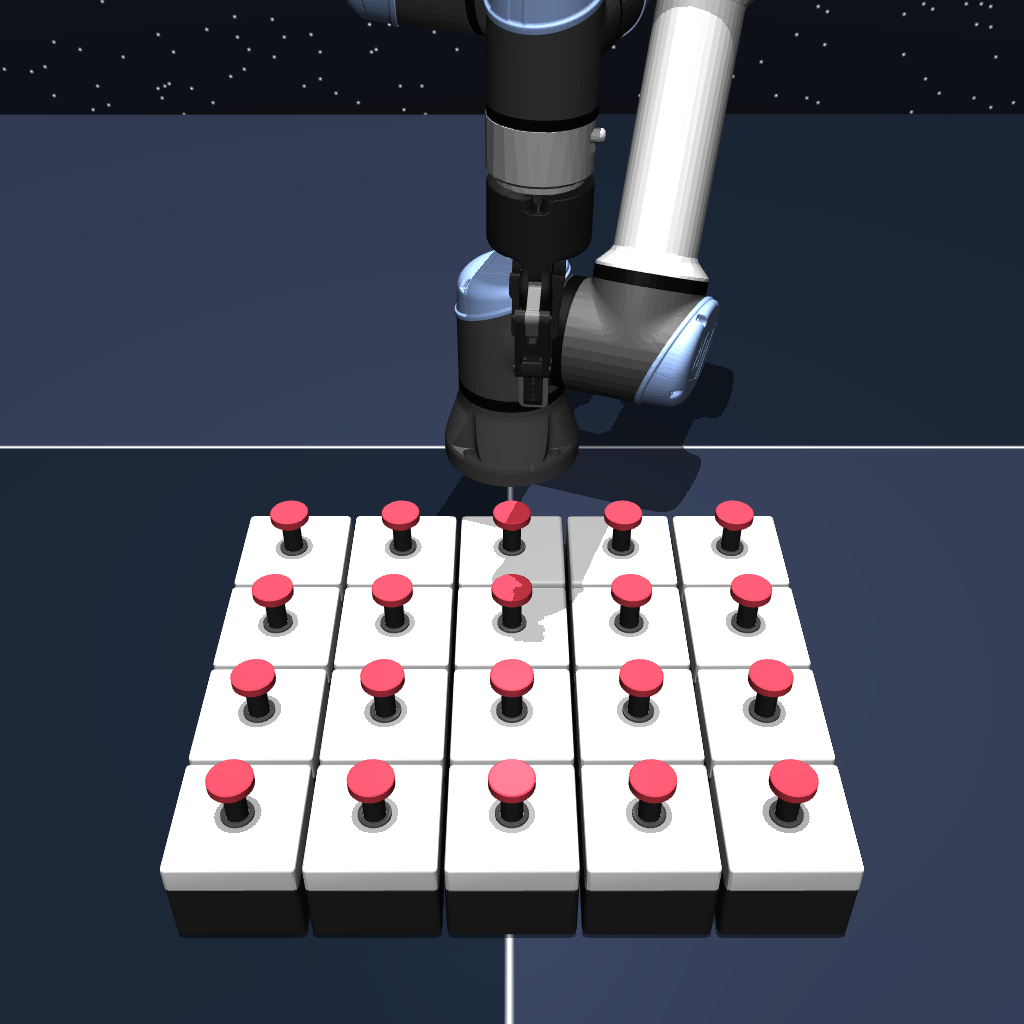}
            \vspace{-15pt}
            \captionsetup{font={stretch=0.7}}
            \caption*{\centering \texttt{task5}}
        \end{minipage}
    \end{minipage}
    \vspace{-3pt}
    \caption{\footnotesize \textbf{Evaluation goals for \tt{puzzle-4x5}.}}
    \label{fig:goals_2}
\end{figure}

\begin{figure}[h!]
    \begin{minipage}{\linewidth}
        \centering
        \begin{minipage}{0.18\linewidth}
            \centering
            \includegraphics[width=\linewidth]{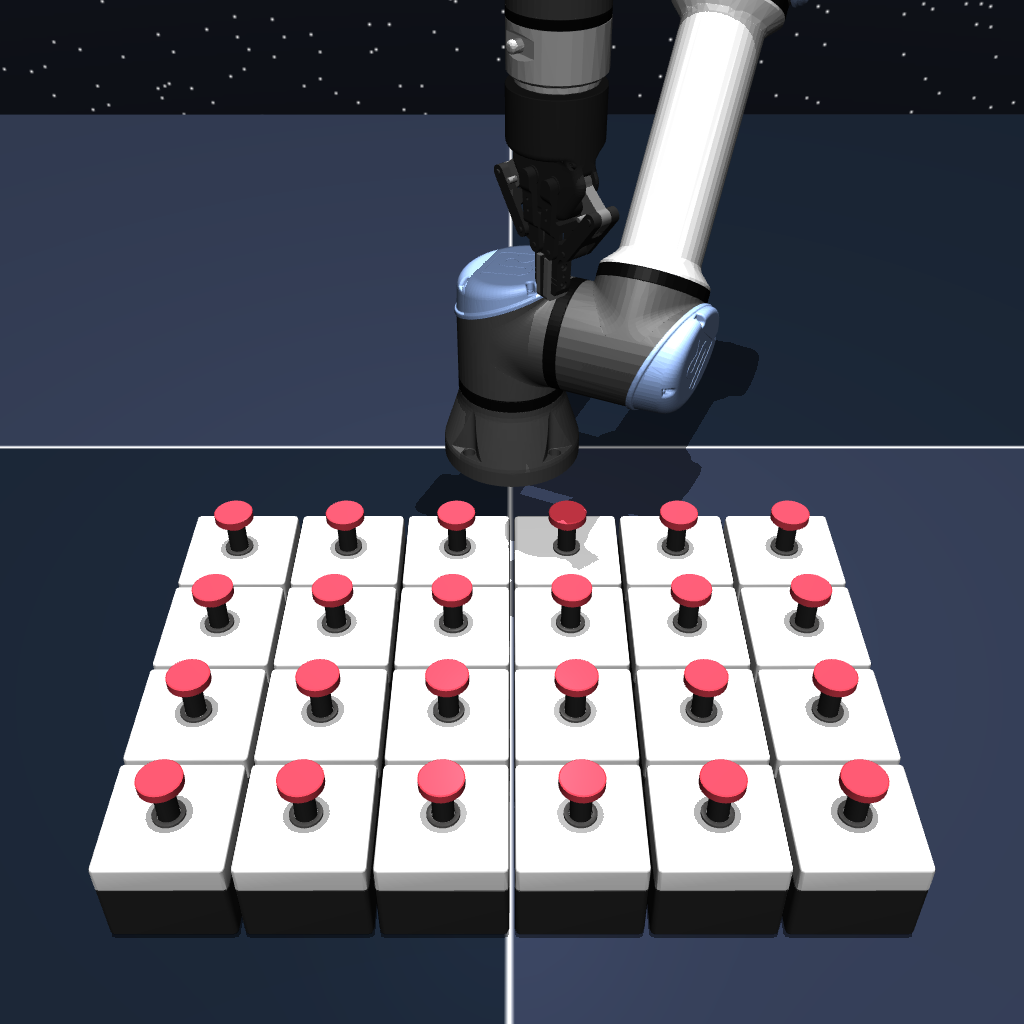}
            \begin{tikzpicture}
                \node at (0, 0.58) {};
                \draw[Triangle-, thick] (0,0.5) -- (0,0);
                \node at (0, -0.1) {};
            \end{tikzpicture}
            \includegraphics[width=\linewidth]{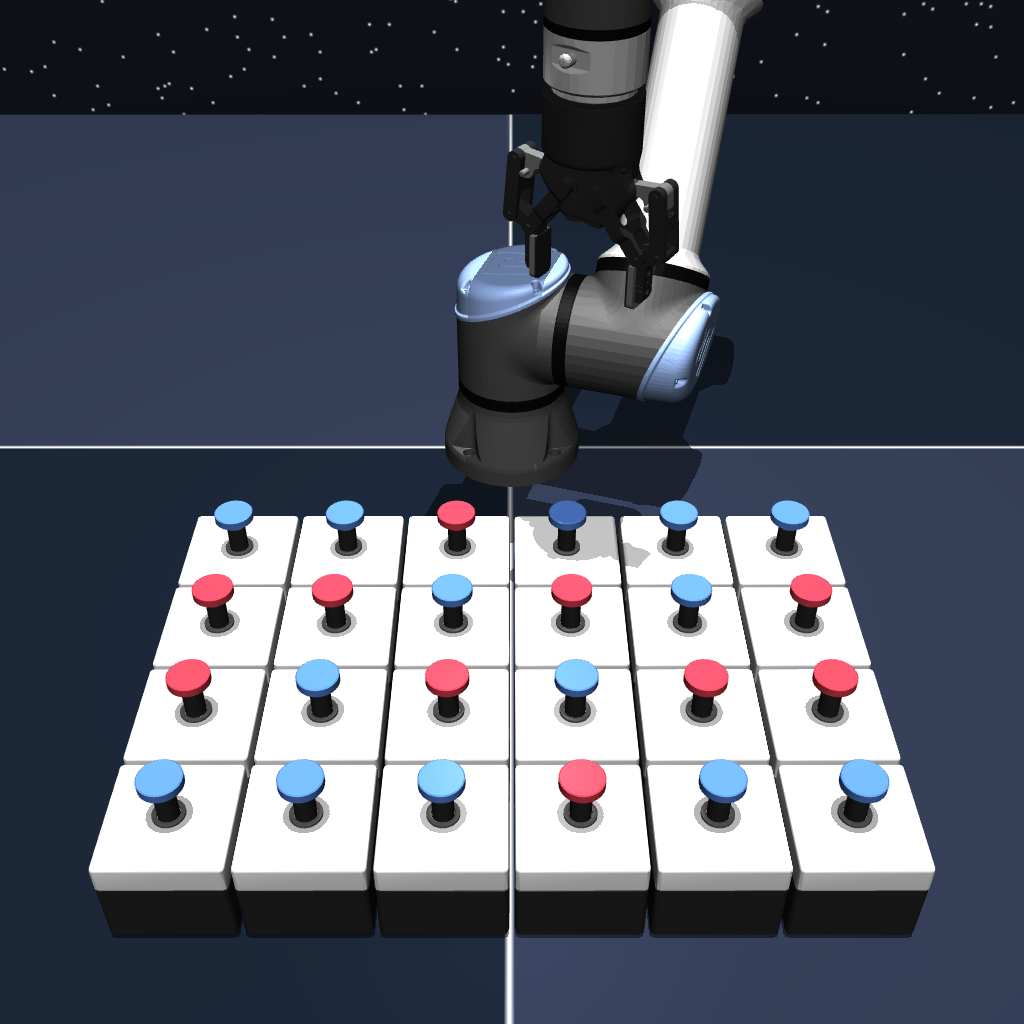}
            \vspace{-15pt}
            \captionsetup{font={stretch=0.7}}
            \caption*{\centering \texttt{task1}}
        \end{minipage}
        \hfill
        \begin{minipage}{0.18\linewidth}
            \centering
            \includegraphics[width=\linewidth]{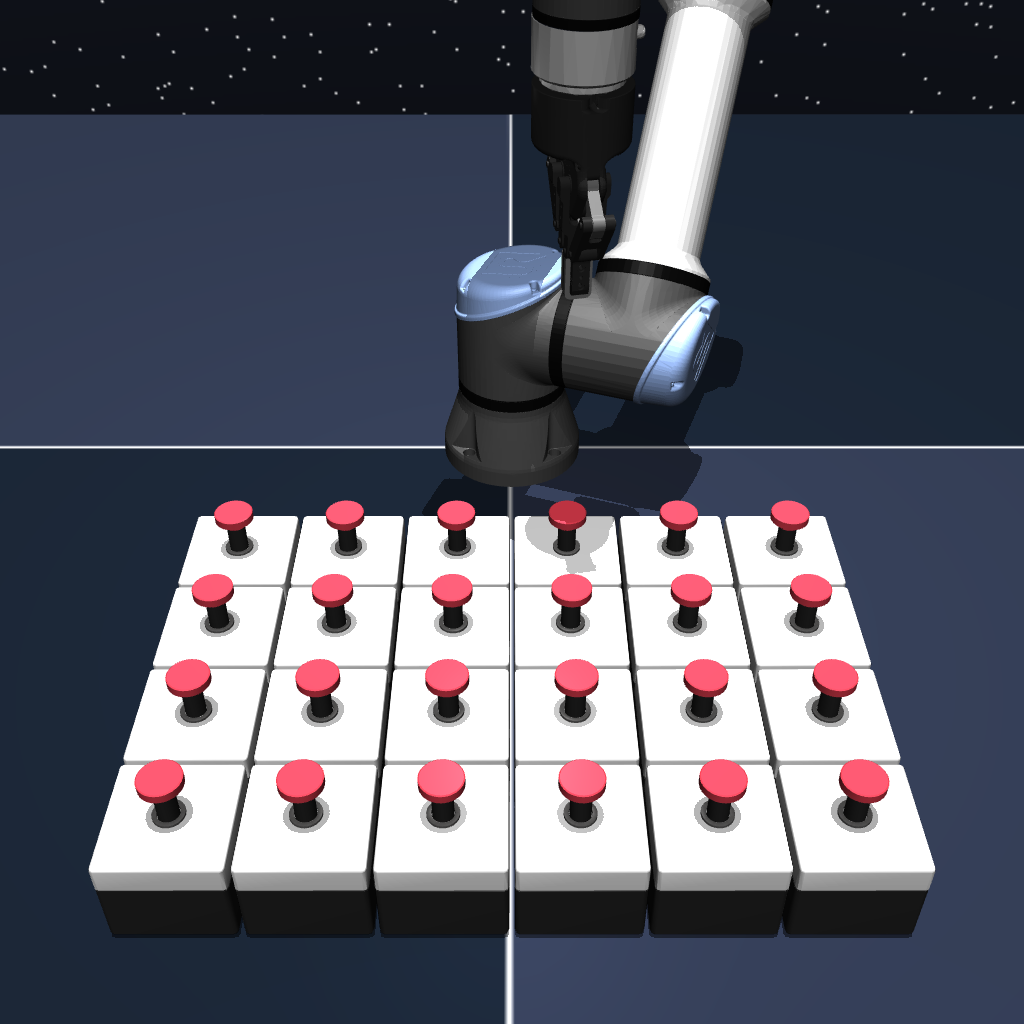}
            \begin{tikzpicture}
                \node at (0, 0.58) {};
                \draw[Triangle-, thick] (0,0.5) -- (0,0);
                \node at (0, -0.1) {};
            \end{tikzpicture}
            \includegraphics[width=\linewidth]{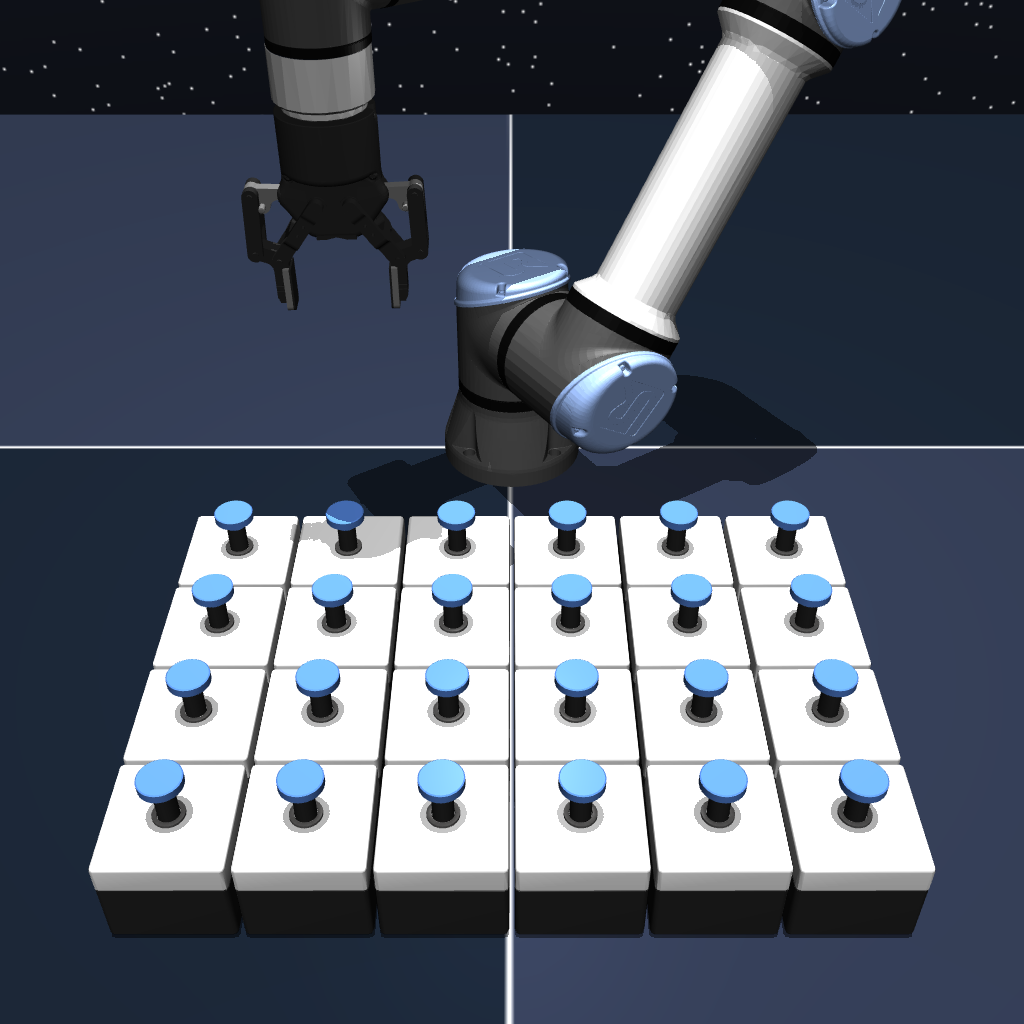}
            \vspace{-15pt}
            \captionsetup{font={stretch=0.7}}
            \caption*{\centering \texttt{task2}}
        \end{minipage}
        \hfill
        \begin{minipage}{0.18\linewidth}
            \centering
            \includegraphics[width=\linewidth]{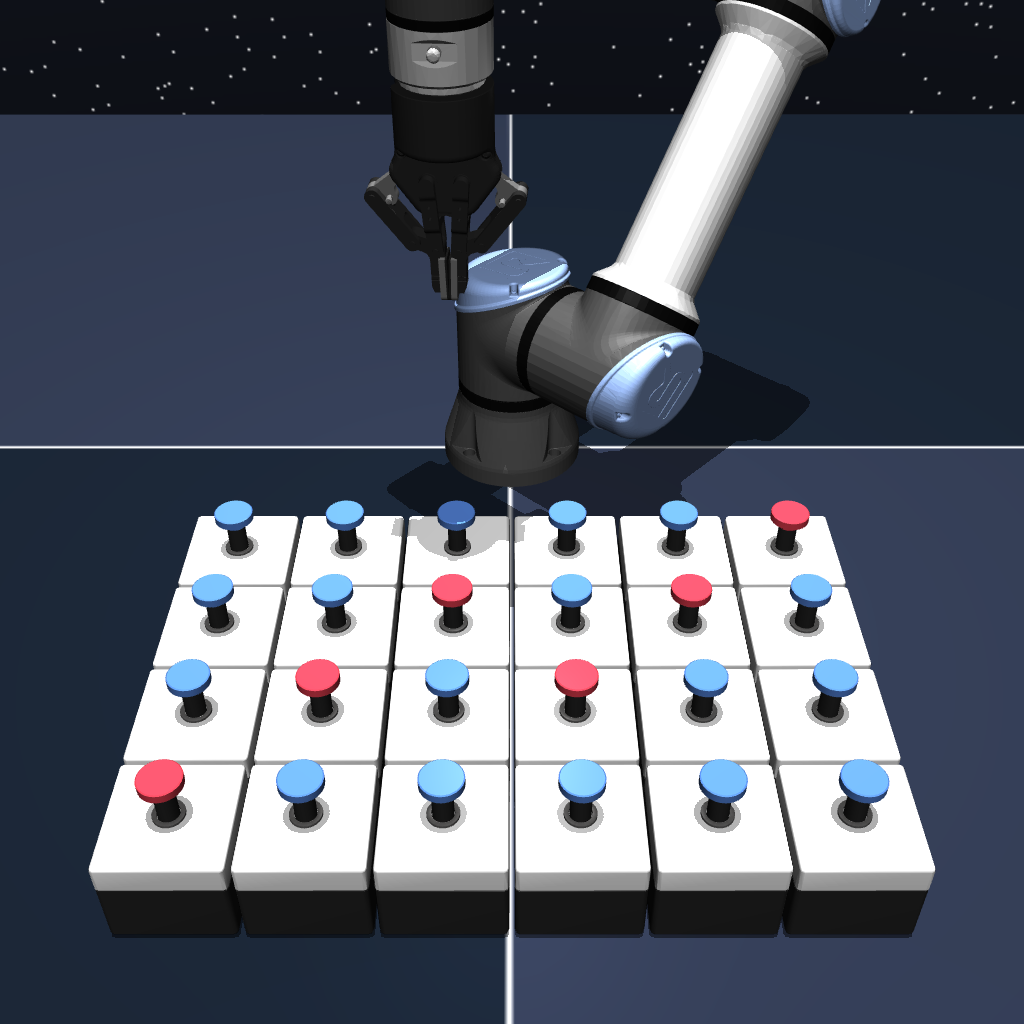}
            \begin{tikzpicture}
                \node at (0, 0.58) {};
                \draw[Triangle-, thick] (0,0.5) -- (0,0);
                \node at (0, -0.1) {};
            \end{tikzpicture}
            \includegraphics[width=\linewidth]{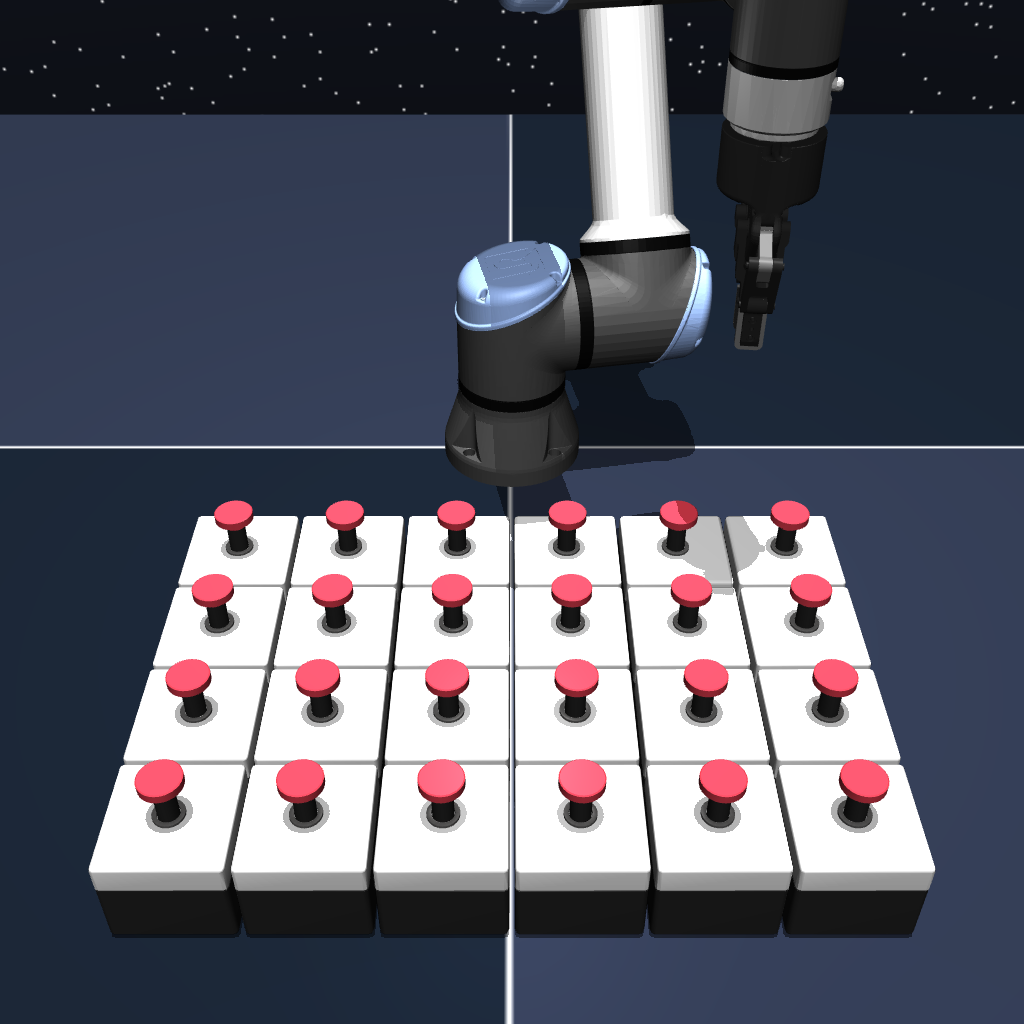}
            \vspace{-15pt}
            \captionsetup{font={stretch=0.7}}
            \caption*{\centering \texttt{task3}}
        \end{minipage}
        \hfill
        \begin{minipage}{0.18\linewidth}
            \centering
            \includegraphics[width=\linewidth]{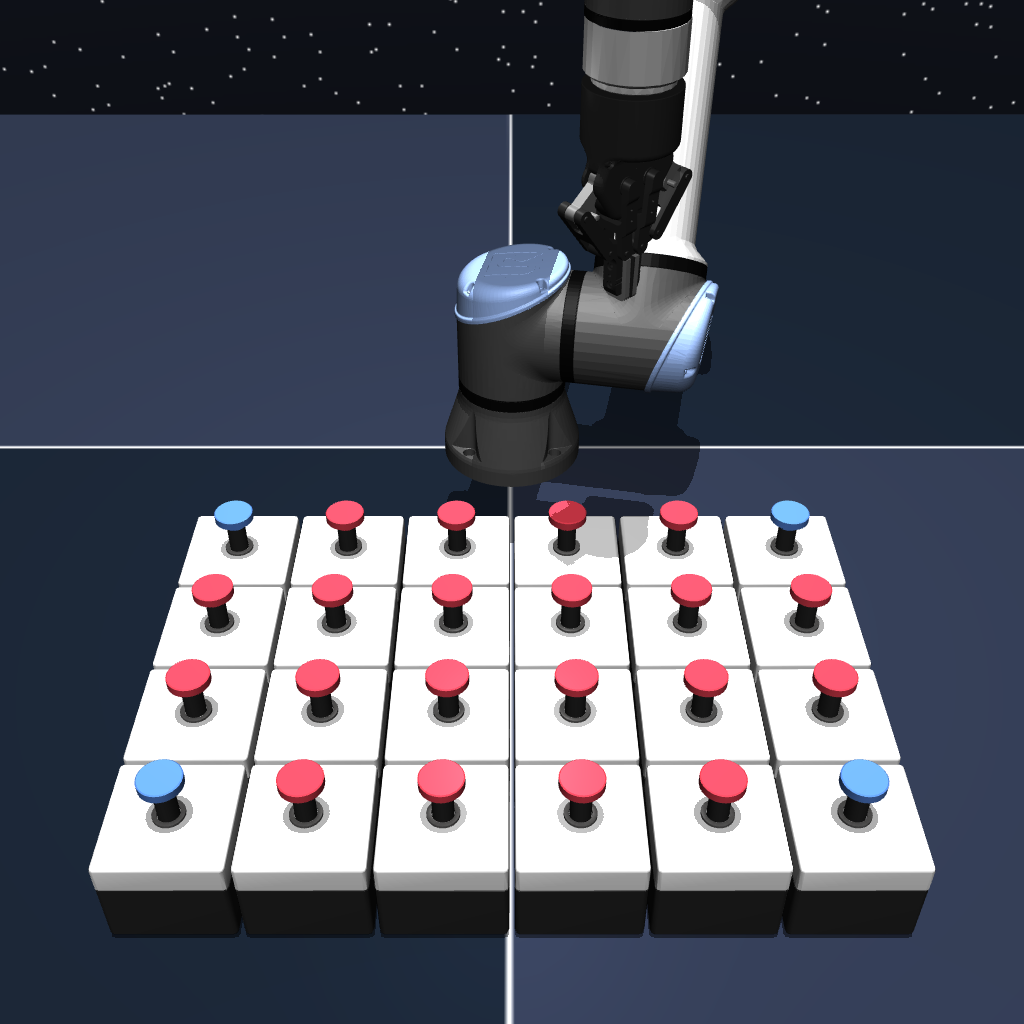}
            \begin{tikzpicture}
                \node at (0, 0.58) {};
                \draw[Triangle-, thick] (0,0.5) -- (0,0);
                \node at (0, -0.1) {};
            \end{tikzpicture}
            \includegraphics[width=\linewidth]{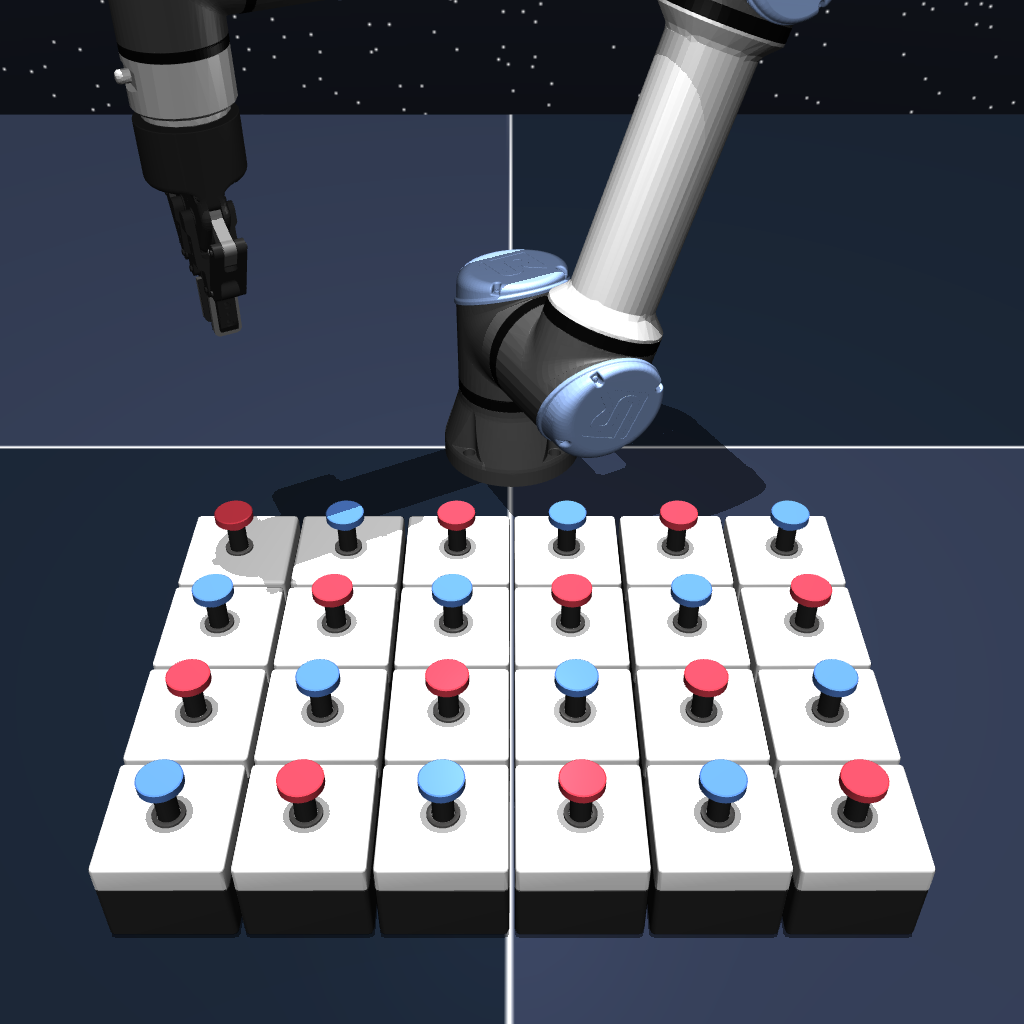}
            \vspace{-15pt}
            \captionsetup{font={stretch=0.7}}
            \caption*{\centering \texttt{task4}}
        \end{minipage}
        \hfill
        \begin{minipage}{0.18\linewidth}
            \centering
            \includegraphics[width=\linewidth]{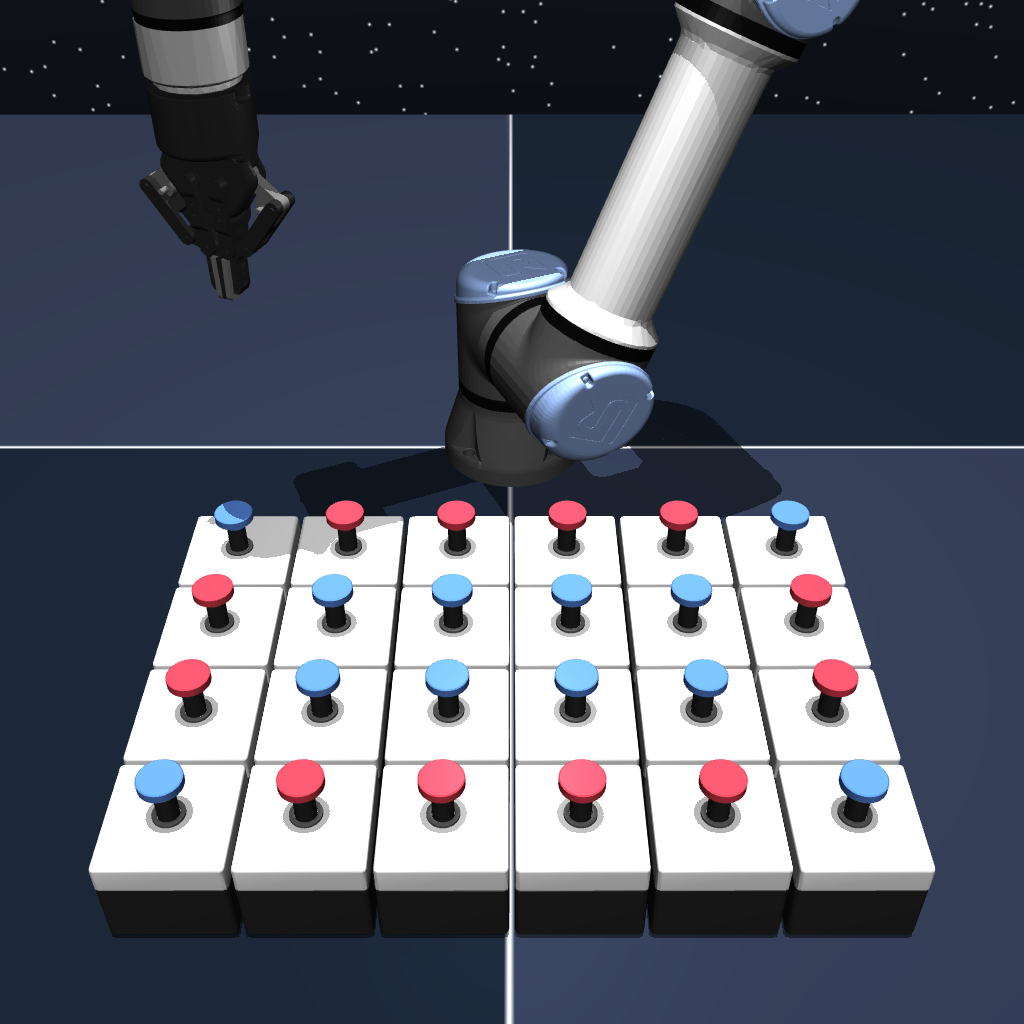}
            \begin{tikzpicture}
                \node at (0, 0.58) {};
                \draw[Triangle-, thick] (0,0.5) -- (0,0);
                \node at (0, -0.1) {};
            \end{tikzpicture}
            \includegraphics[width=\linewidth]{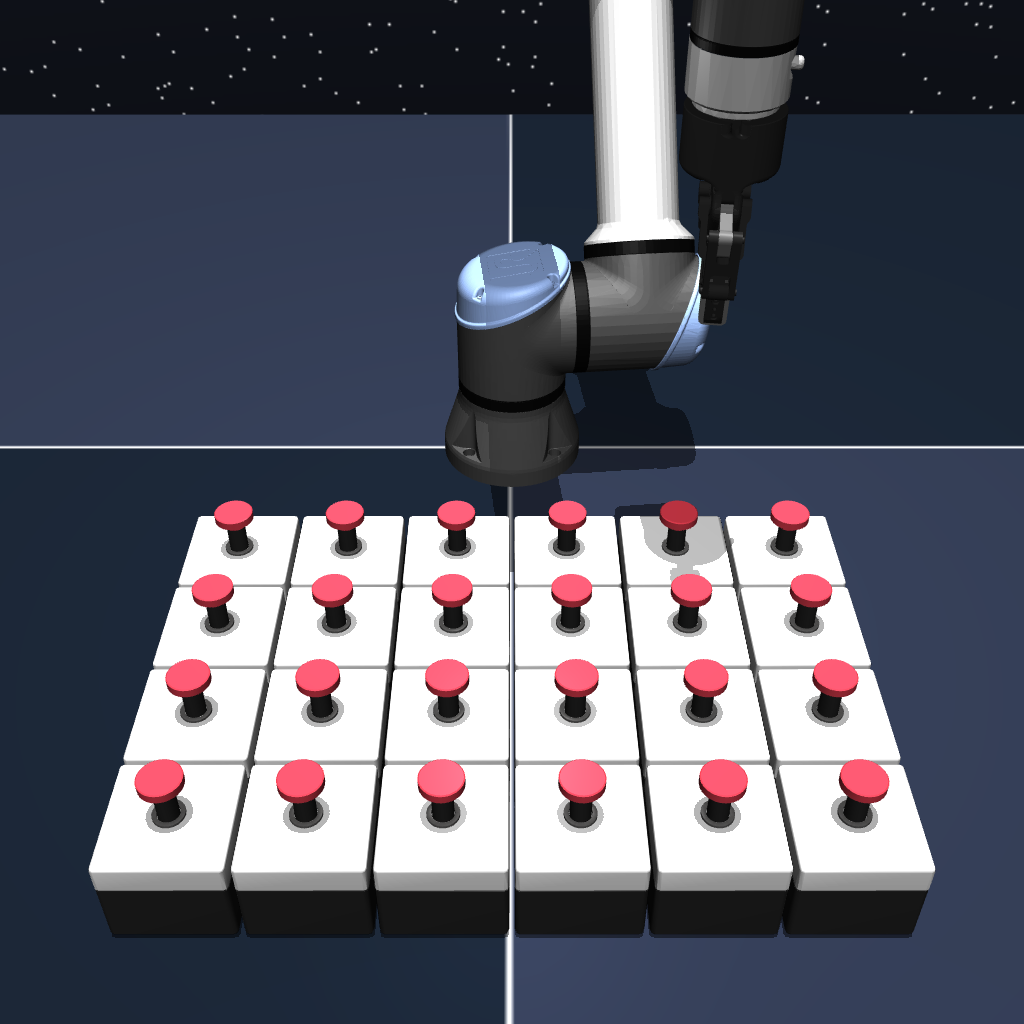}
            \vspace{-15pt}
            \captionsetup{font={stretch=0.7}}
            \caption*{\centering \texttt{task5}}
        \end{minipage}
    \end{minipage}
    \vspace{-3pt}
    \caption{\footnotesize \textbf{Evaluation goals for \tt{puzzle-4x6}.}}
    \label{fig:goals_3}
\end{figure}

\begin{figure}[h!]
    \begin{minipage}{\linewidth}
        \centering
        \begin{minipage}{0.18\linewidth}
            \centering
            \includegraphics[width=\linewidth]{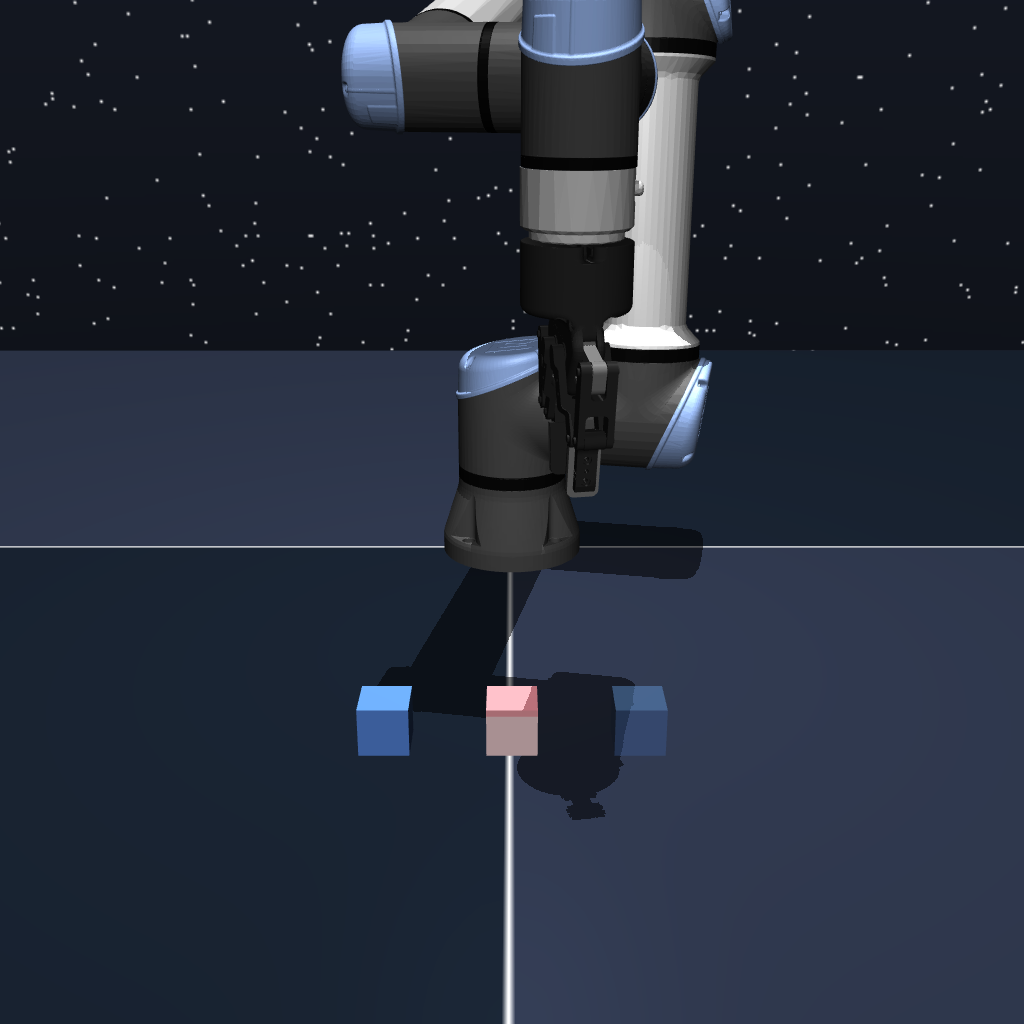}
            \vspace{-15pt}
            \captionsetup{font={stretch=0.7}}
            \caption*{\centering \texttt{task1} \par \texttt{\scriptsize single-pnp}}
        \end{minipage}
        \hfill
        \begin{minipage}{0.18\linewidth}
            \centering
            \includegraphics[width=\linewidth]{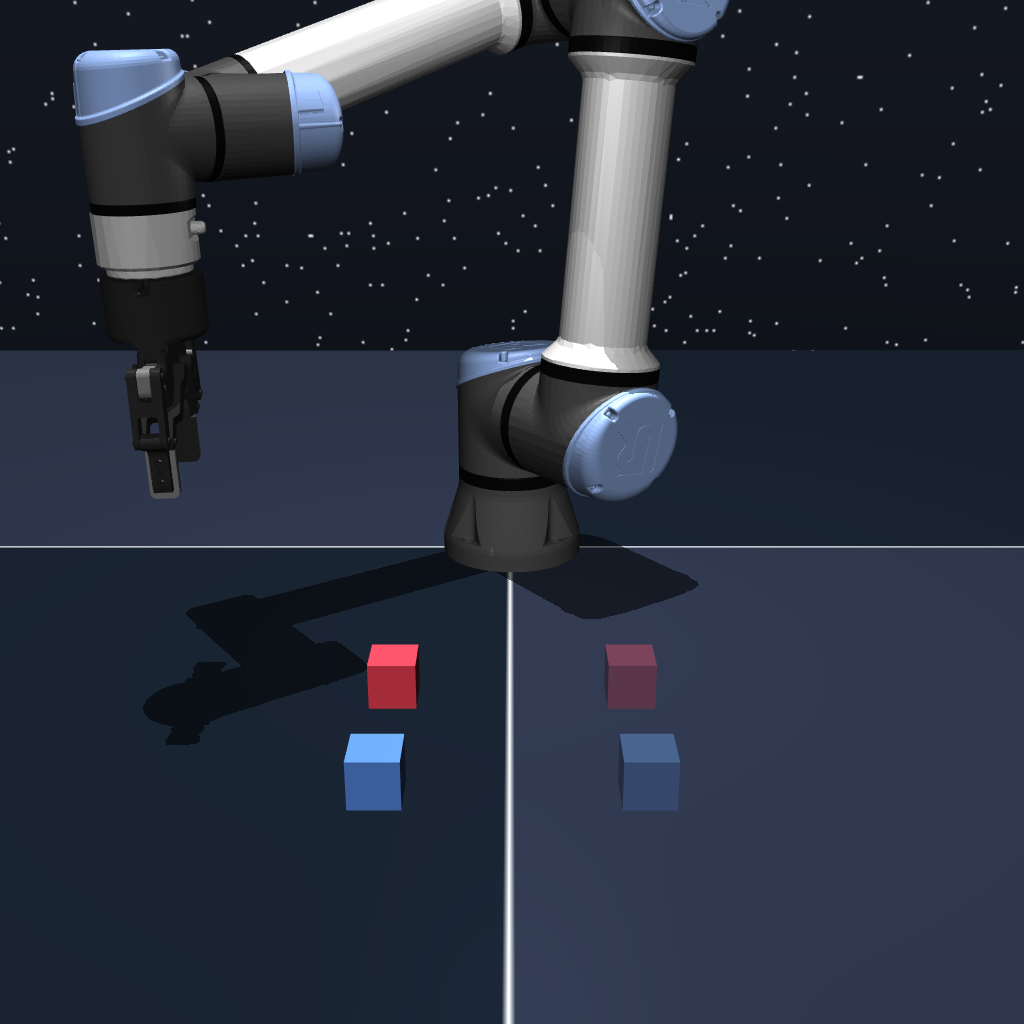}
            \vspace{-15pt}
            \captionsetup{font={stretch=0.7}}
            \caption*{\centering \texttt{task2} \par \texttt{\scriptsize double-pnp1}}
        \end{minipage}
        \hfill
        \begin{minipage}{0.18\linewidth}
            \centering
            \includegraphics[width=\linewidth]{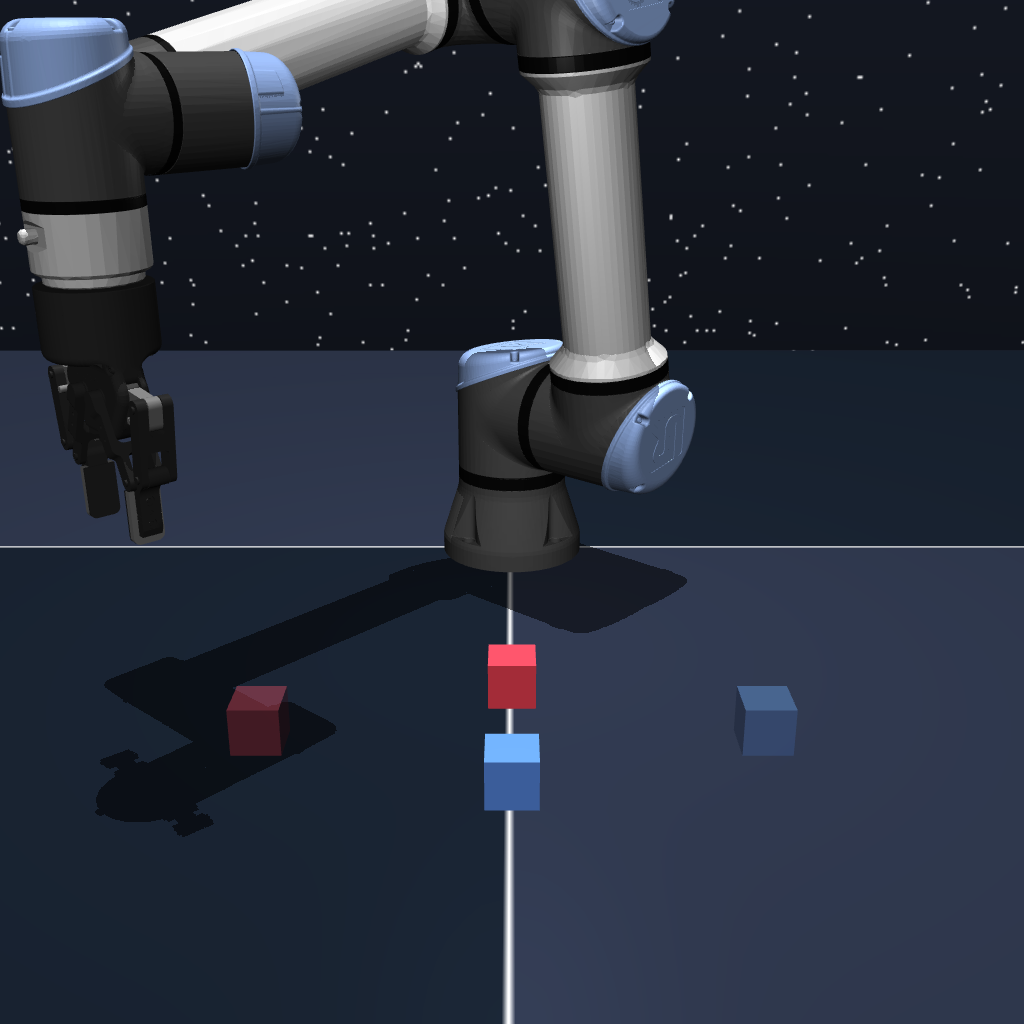}
            \vspace{-15pt}
            \captionsetup{font={stretch=0.7}}
            \caption*{\centering \texttt{task3} \par \texttt{\scriptsize double-pnp2}}
        \end{minipage}
        \hfill
        \begin{minipage}{0.18\linewidth}
            \centering
            \includegraphics[width=\linewidth]{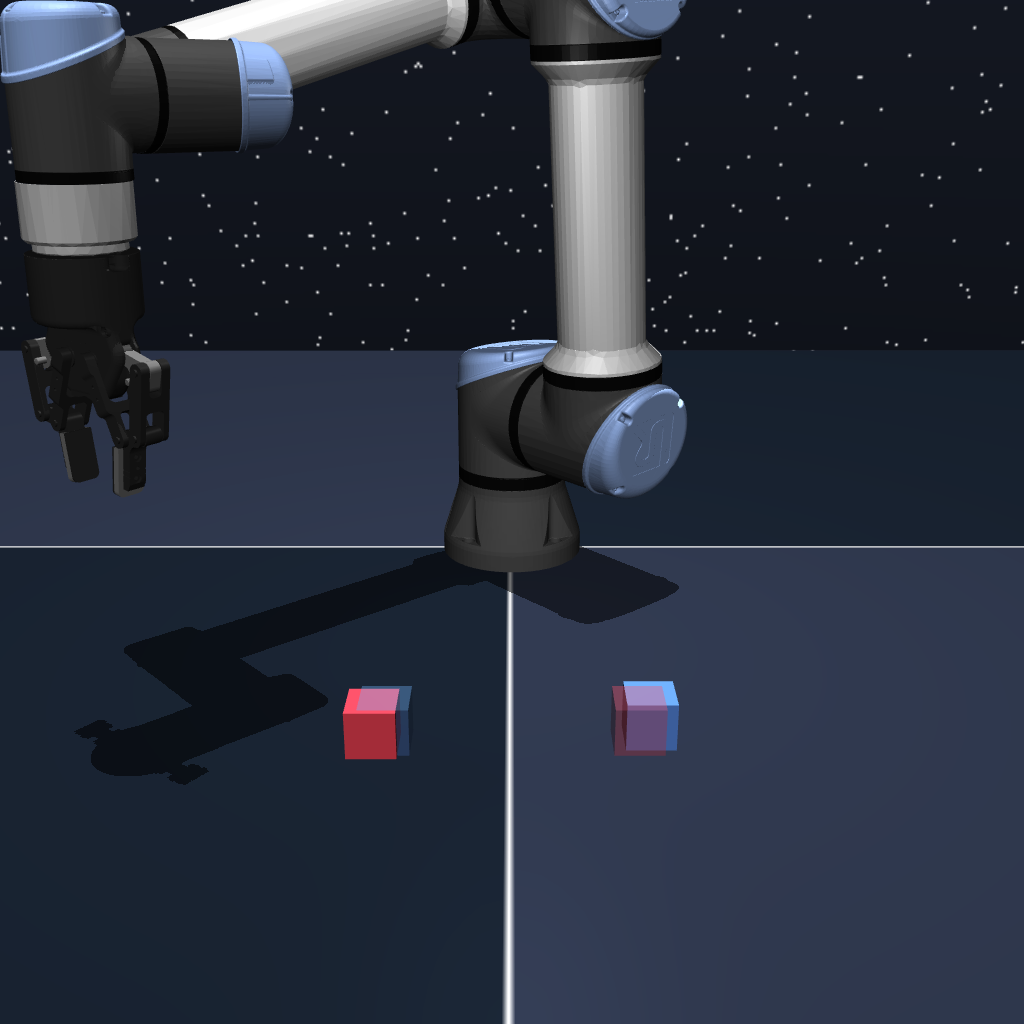}
            \vspace{-15pt}
            \captionsetup{font={stretch=0.7}}
            \caption*{\centering \texttt{task4} \par \texttt{\scriptsize swap}}
        \end{minipage}
        \hfill
        \begin{minipage}{0.18\linewidth}
            \centering
            \includegraphics[width=\linewidth]{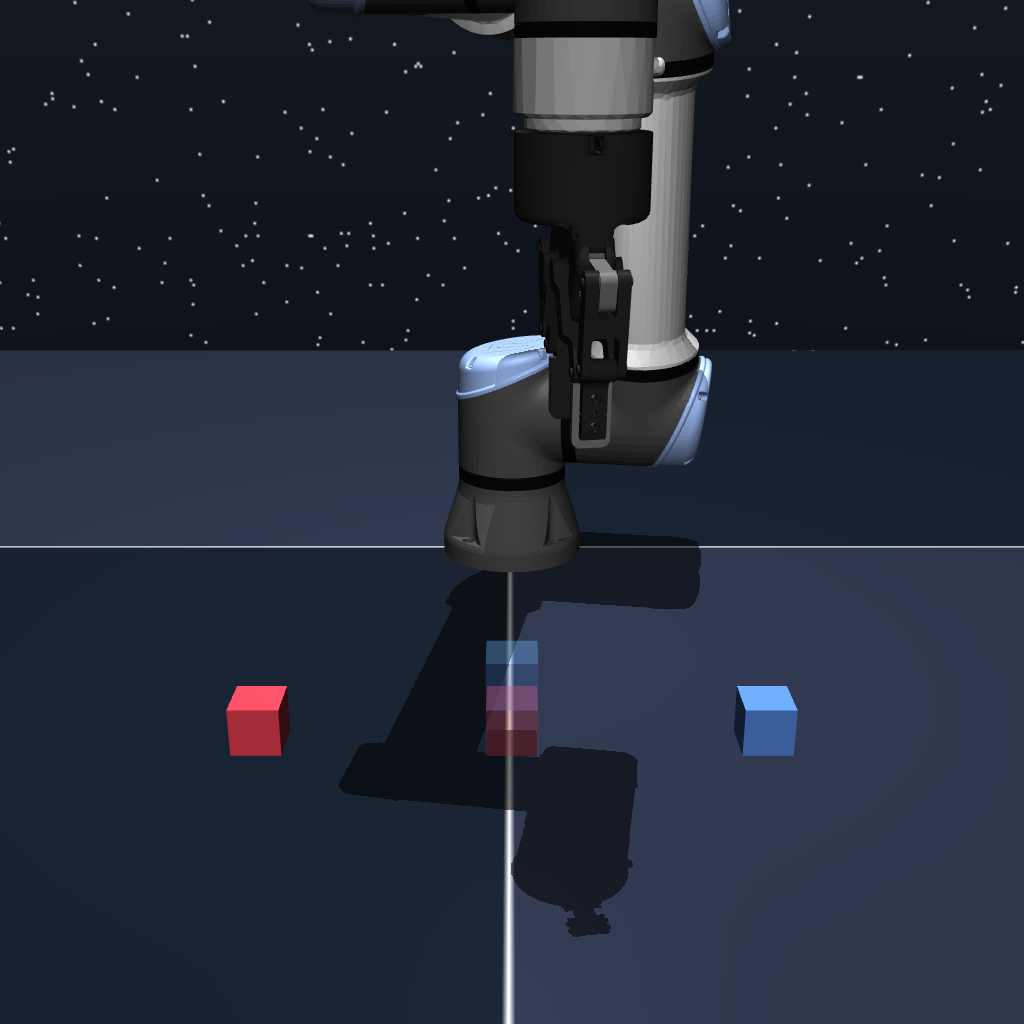}
            \vspace{-15pt}
            \captionsetup{font={stretch=0.7}}
            \caption*{\centering \texttt{task5} \par \texttt{\scriptsize stack}}
        \end{minipage}
    \end{minipage}
    \vspace{-3pt}
    \caption{\footnotesize \textbf{Evaluation goals for \tt{cube-double}.} As stated in \Cref{sec:exp_details_ogbench}, \tt{task4} is omitted from our evaluation.}
    \label{fig:goals_4}
\end{figure}

\begin{figure}[h!]
    \begin{minipage}{\linewidth}
        \centering
        \begin{minipage}{0.18\linewidth}
            \centering
            \includegraphics[width=\linewidth]{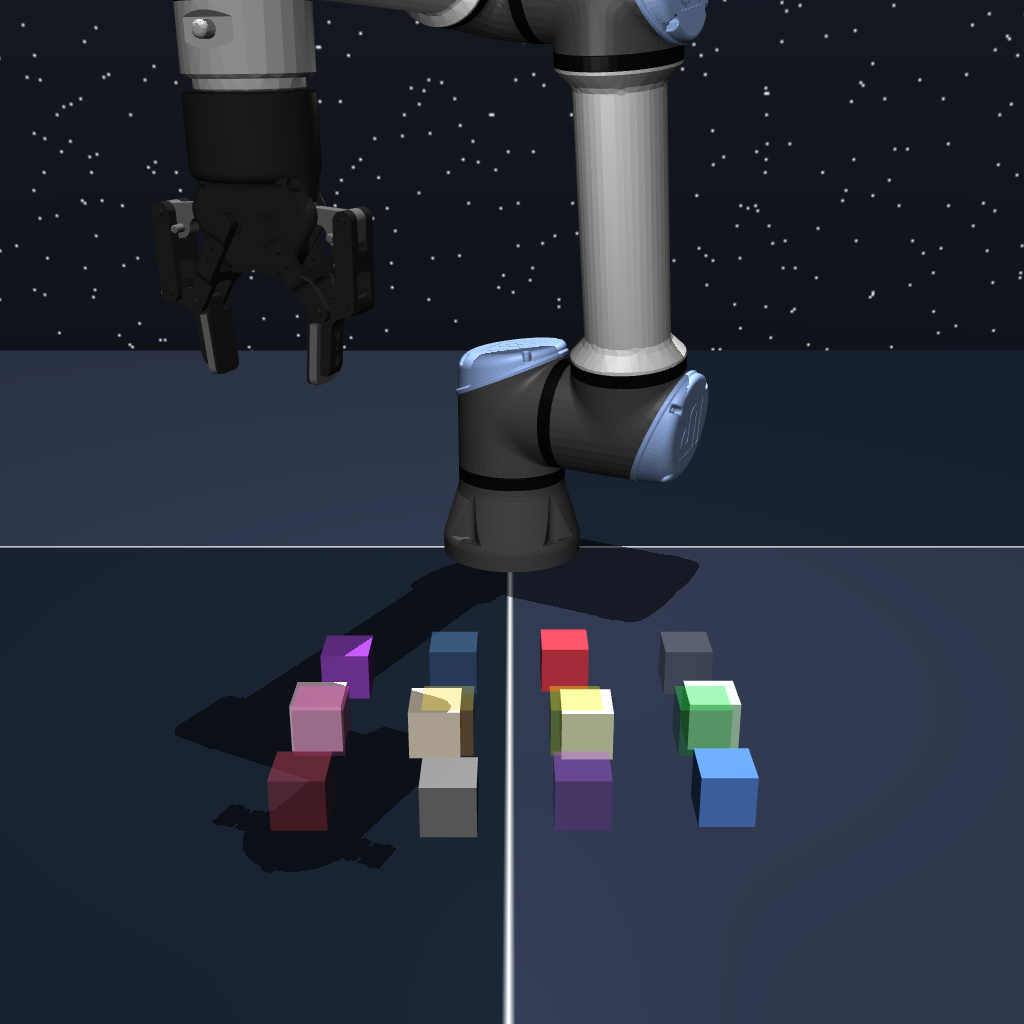}
            \vspace{-15pt}
            \captionsetup{font={stretch=0.7}}
            \caption*{\centering \texttt{task1} \par \texttt{\scriptsize quadruple-pnp}}
        \end{minipage}
        \hfill
        \begin{minipage}{0.18\linewidth}
            \centering
            \includegraphics[width=\linewidth]{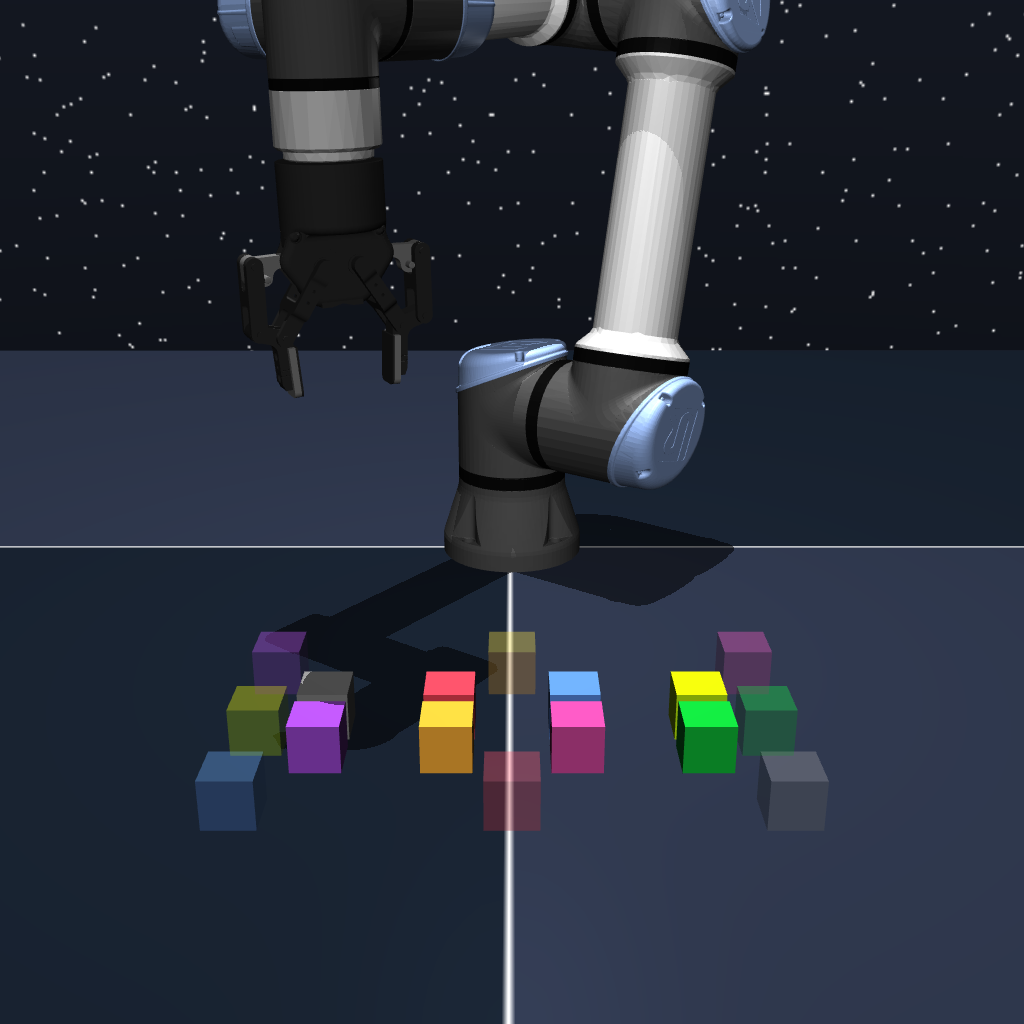}
            \vspace{-15pt}
            \captionsetup{font={stretch=0.7}}
            \caption*{\centering \texttt{task2} \par \texttt{\scriptsize octuple-pnp1}}
        \end{minipage}
        \hfill
        \begin{minipage}{0.18\linewidth}
            \centering
            \includegraphics[width=\linewidth]{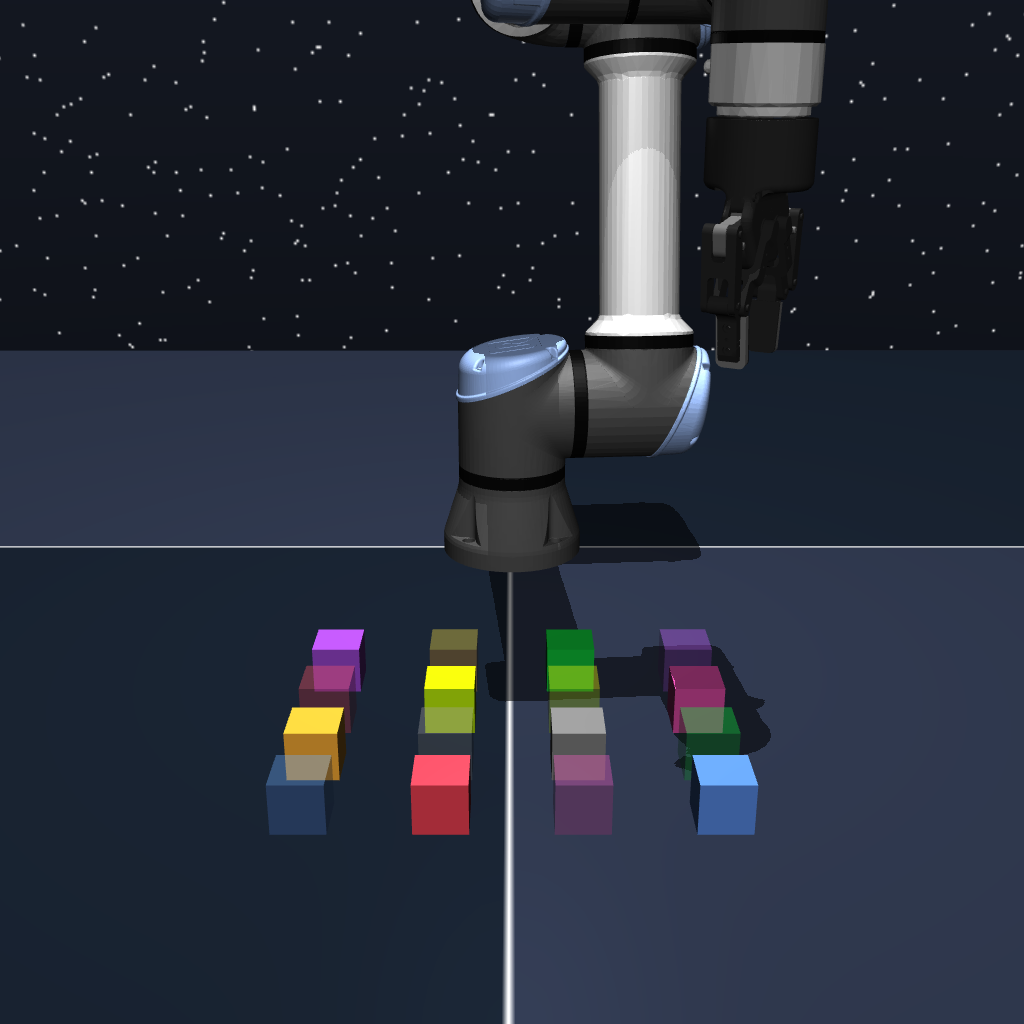}
            \vspace{-15pt}
            \captionsetup{font={stretch=0.7}}
            \caption*{\centering \texttt{task3} \par \texttt{\scriptsize octuple-pnp2}}
        \end{minipage}
        \hfill
        \begin{minipage}{0.18\linewidth}
            \centering
            \includegraphics[width=\linewidth]{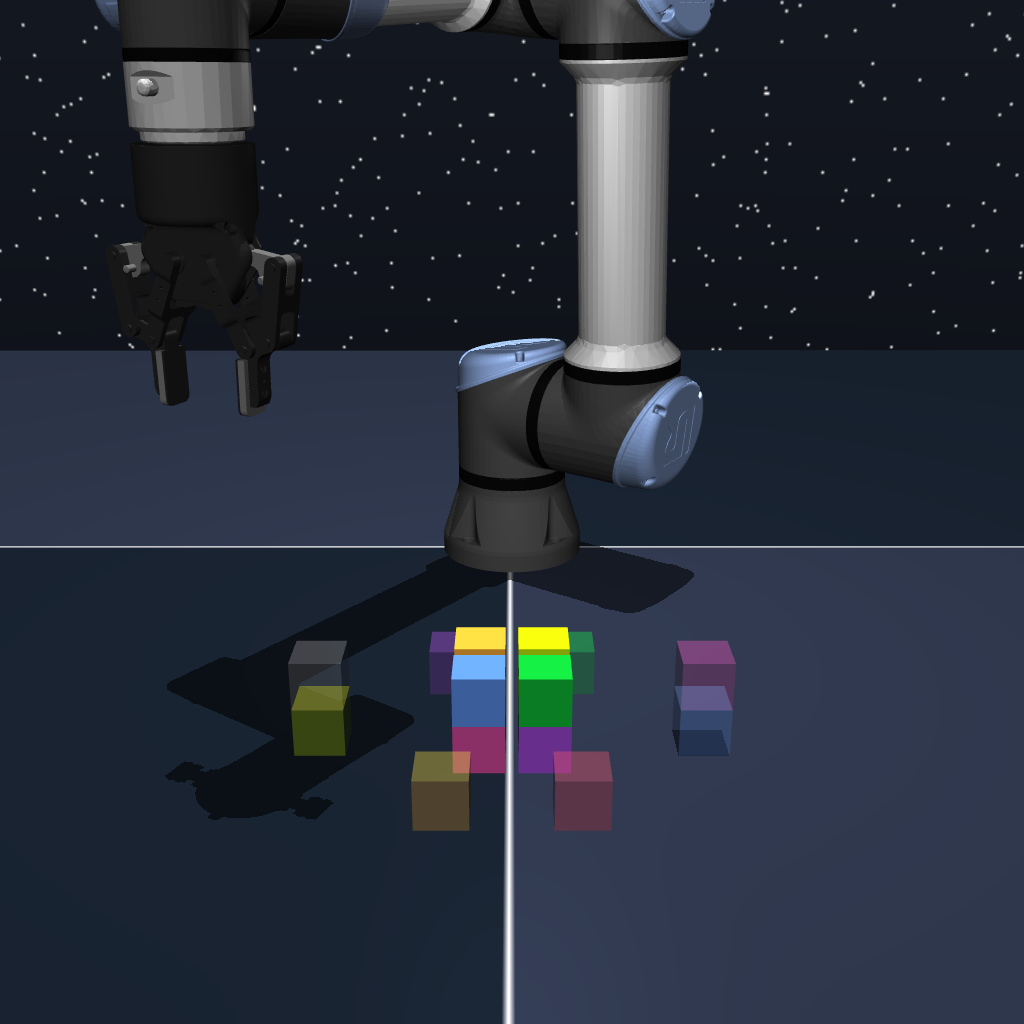}
            \vspace{-15pt}
            \captionsetup{font={stretch=0.7}}
            \caption*{\centering \texttt{task4} \par \texttt{\scriptsize stack1}}
        \end{minipage}
        \hfill
        \begin{minipage}{0.18\linewidth}
            \centering
            \includegraphics[width=\linewidth]{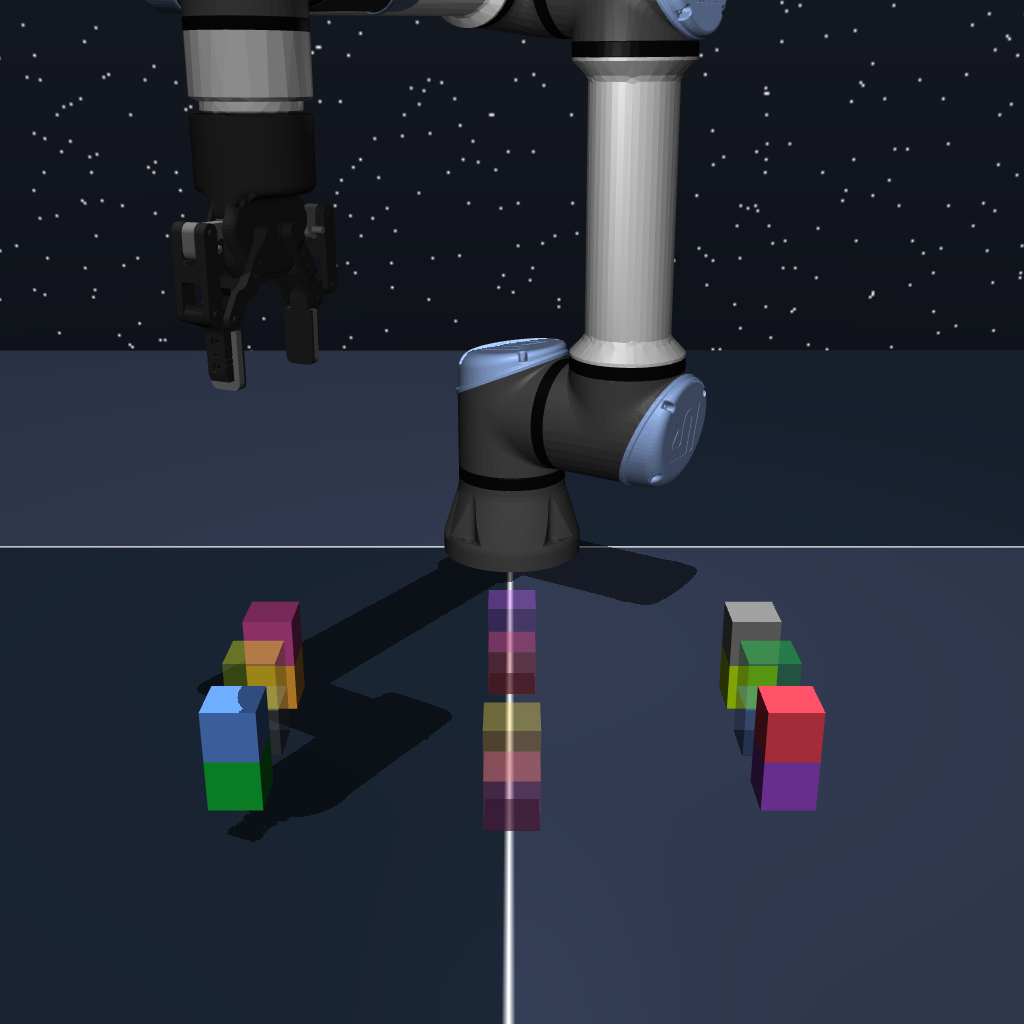}
            \vspace{-15pt}
            \captionsetup{font={stretch=0.7}}
            \caption*{\centering \texttt{task5} \par \texttt{\scriptsize stack2}}
        \end{minipage}
    \end{minipage}
    \vspace{-3pt}
    \caption{\footnotesize \textbf{Evaluation goals for \tt{cube-octuple}.}}
    \label{fig:goals_5}
\end{figure}

\begin{figure}[h!]
    \begin{minipage}{\linewidth}
        \centering
        \begin{minipage}{0.18\linewidth}
            \centering
            \includegraphics[width=\linewidth]{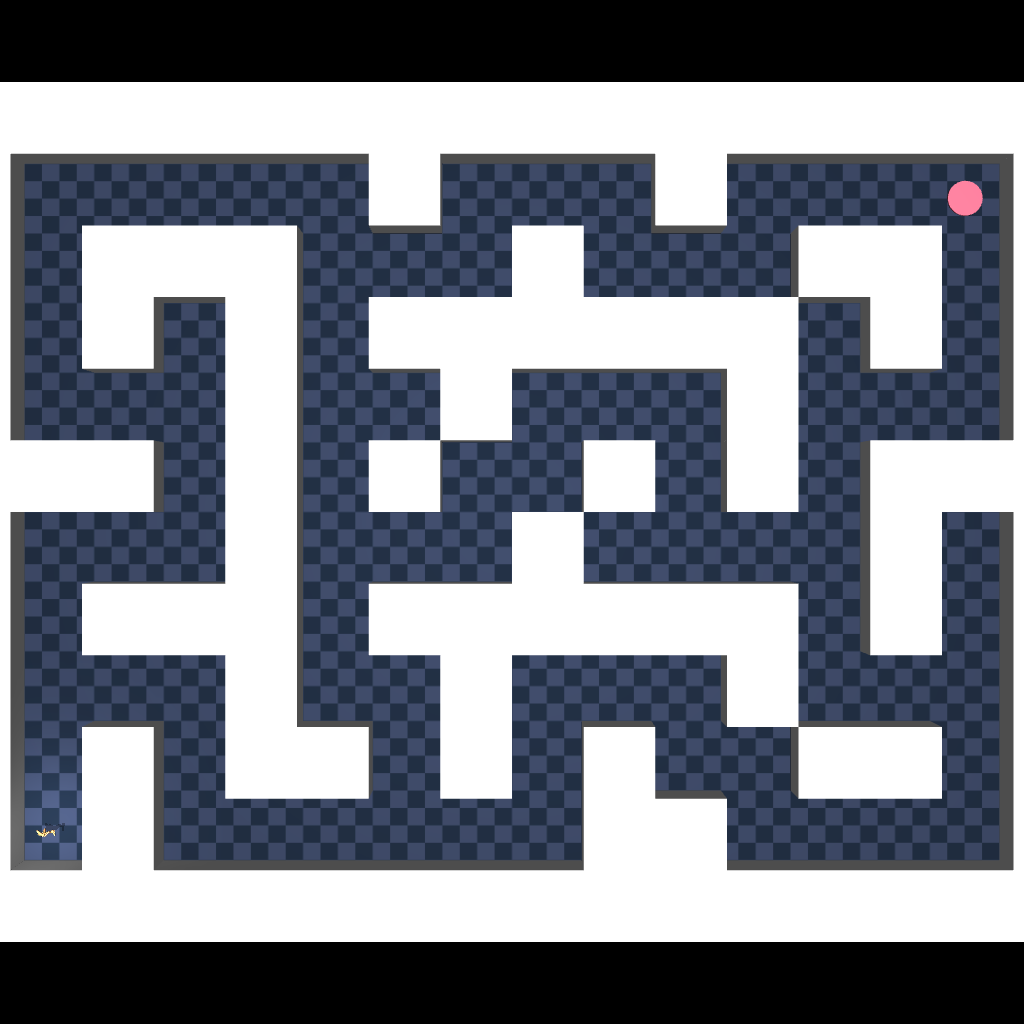}
            \vspace{-15pt}
            \captionsetup{font={stretch=0.7}}
            \caption*{\centering \texttt{task1}}
        \end{minipage}
        \hfill
        \begin{minipage}{0.18\linewidth}
            \centering
            \includegraphics[width=\linewidth]{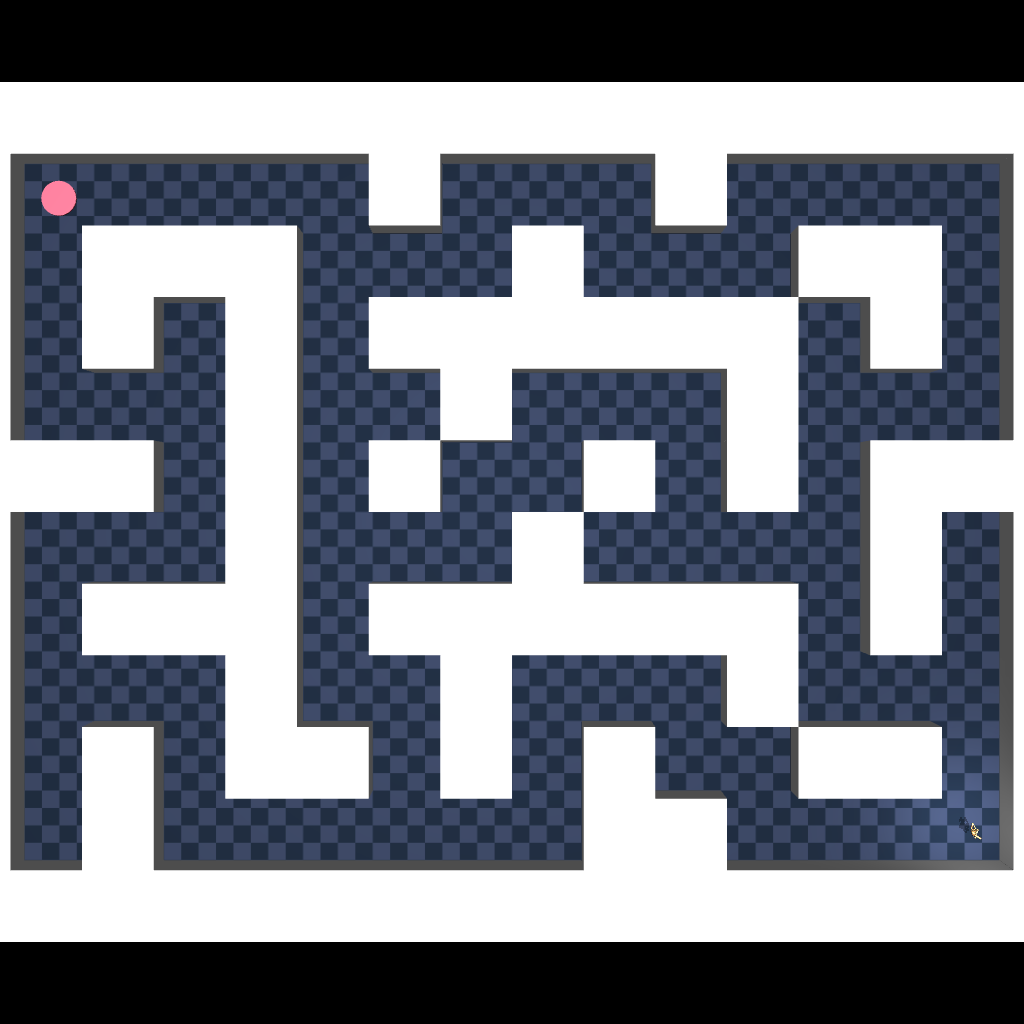}
            \vspace{-15pt}
            \captionsetup{font={stretch=0.7}}
            \caption*{\centering \texttt{task2}}
        \end{minipage}
        \hfill
        \begin{minipage}{0.18\linewidth}
            \centering
            \includegraphics[width=\linewidth]{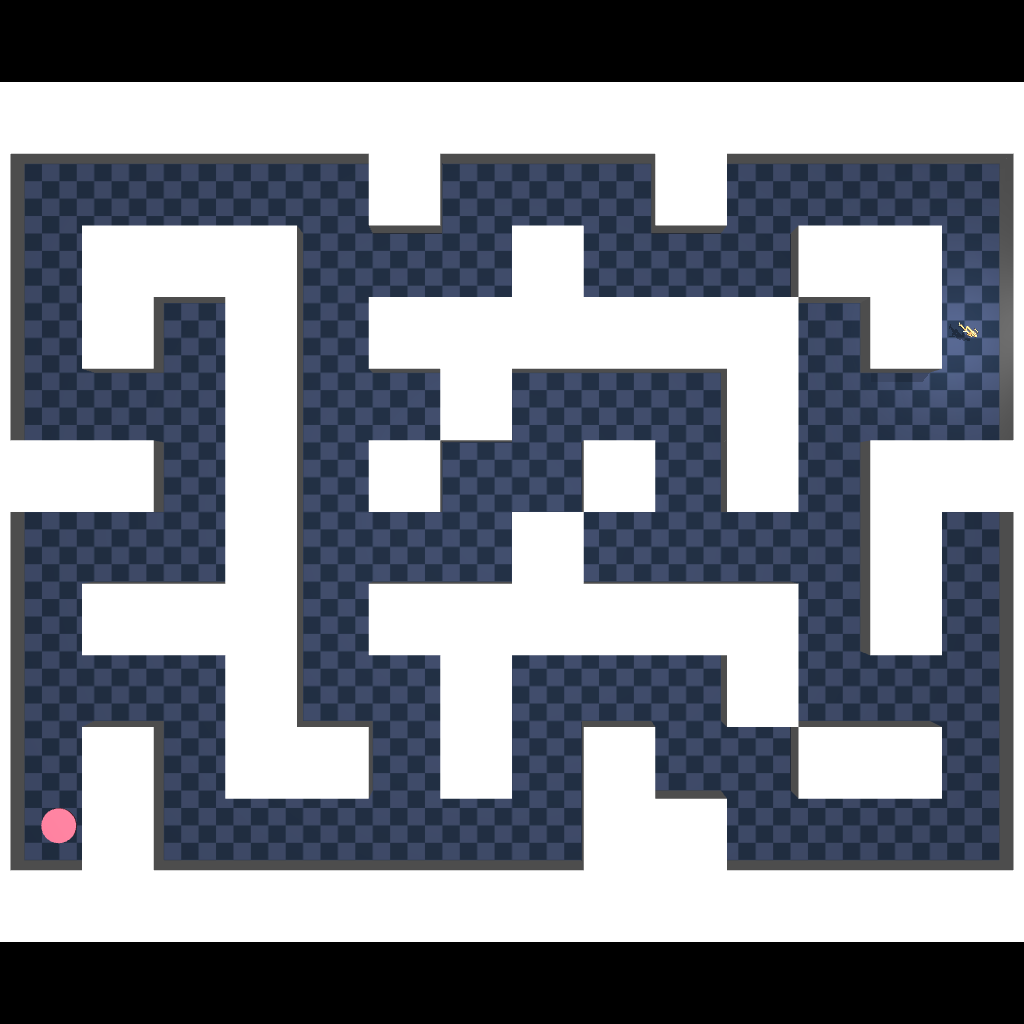}
            \vspace{-15pt}
            \captionsetup{font={stretch=0.7}}
            \caption*{\centering \texttt{task3}}
        \end{minipage}
        \hfill
        \begin{minipage}{0.18\linewidth}
            \centering
            \includegraphics[width=\linewidth]{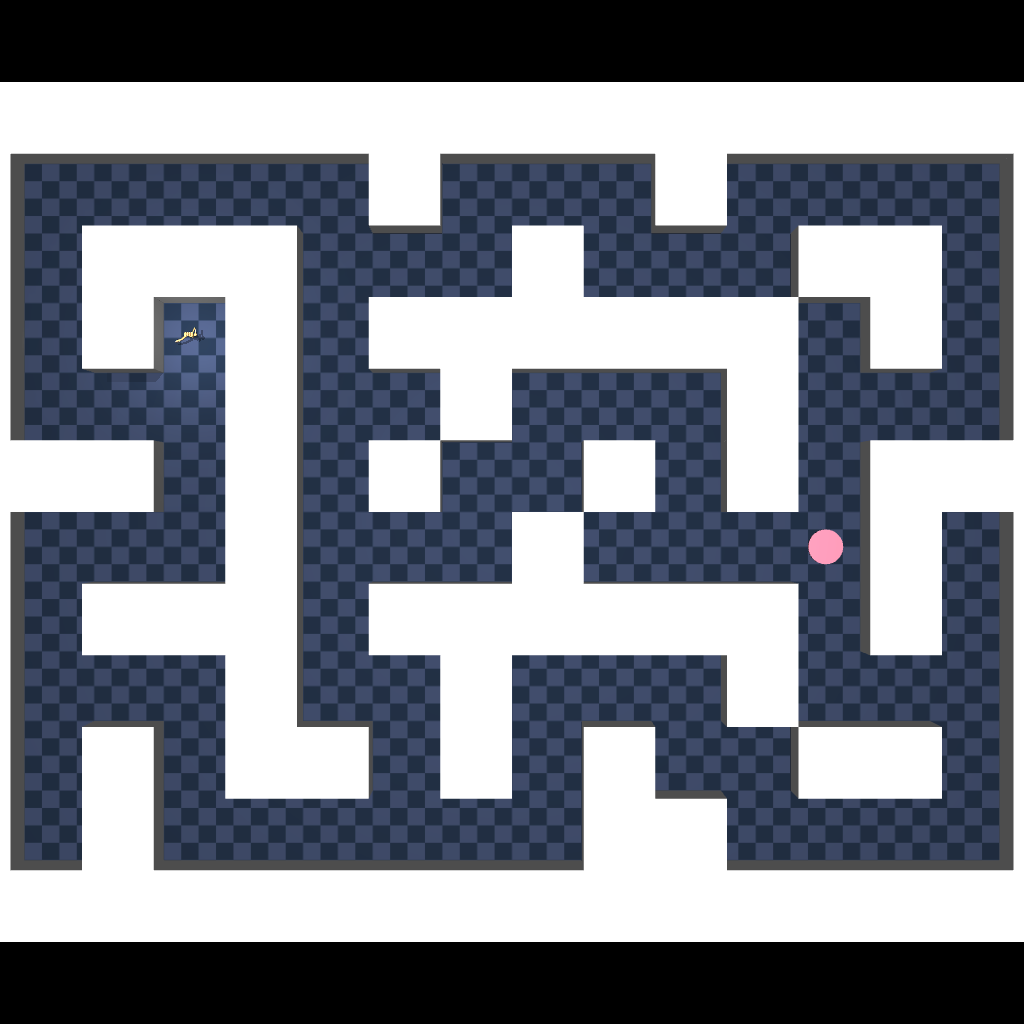}
            \vspace{-15pt}
            \captionsetup{font={stretch=0.7}}
            \caption*{\centering \texttt{task4}}
        \end{minipage}
        \hfill
        \begin{minipage}{0.18\linewidth}
            \centering
            \includegraphics[width=\linewidth]{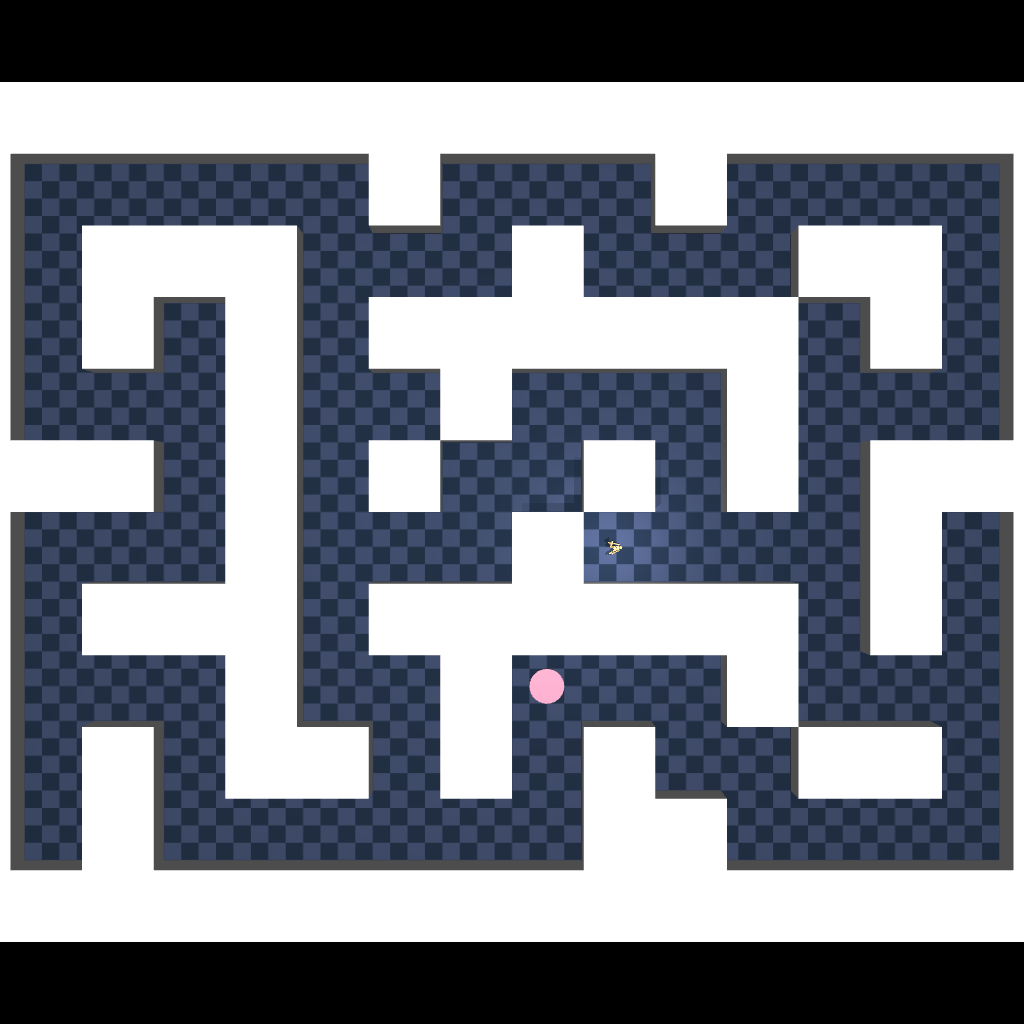}
            \vspace{-15pt}
            \captionsetup{font={stretch=0.7}}
            \caption*{\centering \texttt{task5}}
        \end{minipage}
    \end{minipage}
    \vspace{-3pt}
    \caption{\footnotesize \textbf{Evaluation goals for \tt{humanoidmaze-giant}.}}
    \label{fig:goals_6}
\end{figure}

\textbf{Tasks.}
We use three existing tasks and one new task from OGBench~\citep{ogbench_park2025}:
\tt{humanoidmaze-giant}, \tt{puzzle-4x5}, \tt{puzzle-4x6}, and \tt{cube-octuple}.
In the \texttt{cube} domain, we extend the most challenging existing task, \texttt{cube-quadruple} (with $4$ cubes),
to create a new task, \texttt{cube-octuple} (with $8$ cubes),
to further challenge the agents.
All these tasks are state-based and goal-conditioned.
We employ the \tt{oraclerep} variants from OGBench, which provide ground-truth goal representations
(\eg, in \tt{cube}, the goal is defined only by the cube positions, not including the agent's proprioceptive states).
This helps eliminate confounding factors related to goal representation learning.
For the \tt{cube-double} task used in \Cref{fig:flat_easy},
we exclude the swapping task (\tt{task4}) from the evaluation goals (\Cref{fig:goals_4}),
as we found that this task requires a non-trivial degree of distributional generalization.
We refer to \Cref{fig:goals_1,fig:goals_2,fig:goals_3,fig:goals_4,fig:goals_5,fig:goals_6}
for illustrations of the evaluation goals,
where the goal images for existing tasks are adopted from \citet{ogbench_park2025}.

\textbf{Datasets.}
On each of these tasks, we generate a $1$B-sized dataset using the original data-generation script provided by OGBench.
The \tt{cube} and \tt{puzzle} datasets consist of length-$1000$ trajectories,
and the \tt{humanoidmaze} dataset consists of length-$4000$ trajectories, as in the original datasets.
These datasets are collected by scripted policies that perform random tasks with a certain degree of noise.
In \tt{humanoidmaze}, the agent repeatedly reaches random positions using a (noisy) expert low-level controller;
in \tt{cube}, the agent repeatedly picks a random cube and places it in a random position;
in \tt{puzzle}, the agent repeatedly presses buttons in an arbitrary order.
Notably, these datasets are collected in an unsupervised, task-agnostic manner
(\ie, in the ``play''-style~\citep{play_lynch2019}).
In other words, the data-collection scripts are \emph{not} aware of the evaluation goals.

\textbf{Methods and hyperparameters.}
We generally follow the original implementations, hyperparameters, and evaluation protocols of \citet{ogbench_park2025}.
We train each offline RL algorithm for $5$M gradient steps ($2.5$M steps for simpler tasks in \Cref{fig:flat_easy})
and evaluate every $250$K steps.
At each evaluation epoch, we measure the success rate of the agent using
$15$ rollouts on each of the $5$ ($4$ for \tt{cube-double}) evaluation goals.
For data-scaling plots,
we compute the average success rate over the last three evaluation epochs (\ie, $4.5$M, $4.75$M, and $5$M steps),
following \citet{ogbench_park2025}.

The hyperparameters (in particular, the degree of behavioral regularization) of each algorithm
are individually tuned on each task based on the largest $1$B datasets.
We provide the full list of hyperparameters in \Cref{table:hyp,table:policy_hyp}, where we abbreviate $n$-step SAC+BC as $n$-SAC+BC and double SHARSA as DSHARSA.

\section{Result tables}
\label{sec:result_tables}

We provide result tables in \Cref{table:results_1,table:results_2},
where standard deviations are denoted by the ``$\pm$'' sign.
In the tables, we abbreviate flow BC as FBC, hierarchical flow BC as HFBC,
$n$-step SAC+BC as $n$-SAC+BC, and double SHARSA as DSHARSA.
We highlight values at or above $95\%$ of the best performance in bold, following \citet{ogbench_park2025}.

\clearpage

\begin{table}[h!]
\caption{
\footnotesize
\textbf{Hyperparameters for didactic experiments.}
}
\vspace{5pt}
\label{table:hyp_didactic}
\begin{center}
\scalebox{0.75}
{
\begin{tabular}{ll}
    \toprule
    \textbf{Hyperparameter} & \textbf{Value} \\
    \midrule
    Gradient steps & $5$M \\
    Optimizer & Adam~\citep{adam_kingma2015} \\
    Learning rate & $0.0003$ \\
    Batch size & $512$ \\
    MLP size & $[512, 512, 512]$ \\
    Nonlinearity & GELU~\citep{gelu_hendrycks2016} \\
    Layer normalization & True \\
    Target network update rate & $0.005$ \\
    Discount factor $\gamma$ & $1$ \\
    Horizon reduction factor $n$ & $64$ \\
    \bottomrule
\end{tabular}
}
\end{center}
\end{table}

\begin{table}[h!]
\caption{
\footnotesize
\textbf{Common hyperparameters for OGBench experiments.}
}
\vspace{5pt}
\label{table:hyp}
\begin{center}
\scalebox{0.75}
{
\begin{tabular}{ll}
    \toprule
    \textbf{Hyperparameter} & \textbf{Value} \\
    \midrule
    Gradient steps & $5$M (default), $2.5$M (\tt{cube-double}, \tt{puzzle-4x4}) \\
    Optimizer & Adam~\citep{adam_kingma2015} \\
    Learning rate & $0.0003$ \\
    Batch size & $1024$ \\
    MLP size & $[1024, 1024, 1024, 1024]$ \\
    Nonlinearity & GELU~\citep{gelu_hendrycks2016} \\
    Layer normalization & True \\
    Target network update rate & $0.005$ \\
    Discount factor $\gamma$ & $0.999$ (default), $0.99$ (\tt{cube-double}, \tt{puzzle-4x4}) \\
    Flow steps & $10$ \\
    Horizon reduction factor $n$ & $50$ (\tt{cube}, \tt{humanoidmaze}), $25$ (\tt{puzzle}) \\
    Expectile $\kappa$ (IQL) & $0.9$ \\
    Expectile $\kappa$ (HIQL) & $0.5$ (\tt{cube}, \tt{humanoidmaze}), $0.7$ (\tt{puzzle}) \\
    Value representation dimensionality $k$ (CRL) & $1024$ \\
    Goal representation dimensionality $k$ (HIQL) & $128$ \\
    Double Q aggregation (SAC+BC, SHARSA, FQL) & $\min(Q_1, Q_2)$ (\tt{cube}, \tt{puzzle}), $(Q_1 + Q_2)/2$ (\tt{humanoidmaze}) \\
    Value loss type (SAC+BC, SHARSA, FQL) & Binary cross entropy \\
    Actor $(p^\gD_\mathrm{cur}, p^\gD_\mathrm{geom}, p^\gD_\mathrm{traj}, p^\gD_\mathrm{rand})$ ratio (BC)
    & $(0, 1, 0, 0)$ \\
    Actor $(p^\gD_\mathrm{cur}, p^\gD_\mathrm{geom}, p^\gD_\mathrm{traj}, p^\gD_\mathrm{rand})$ ratio (others)
    & $(0, 1, 0, 0)$ (\tt{cube}), $(0, 0.5, 0, 0.5)$ (\tt{puzzle}), $(0, 0, 1, 0)$ (\tt{humanoidmaze}) \\
    Value $(p^\gD_\mathrm{cur}, p^\gD_\mathrm{geom}, p^\gD_\mathrm{traj}, p^\gD_\mathrm{rand})$ ratio (CRL)
    & $(0, 1, 0, 0)$ \\
    Value $(p^\gD_\mathrm{cur}, p^\gD_\mathrm{geom}, p^\gD_\mathrm{traj}, p^\gD_\mathrm{rand})$ ratio (others)
    & $(0.2, 0, 0.5, 0.3)$ \\
    Policy extraction hyperparameters & \Cref{table:policy_hyp} \\
    \bottomrule
\end{tabular}
}
\end{center}
\end{table} 

\begin{table}[h!]
\caption{
\footnotesize
\textbf{Policy extraction hyperparameters for OGBench experiments.}
}
\vspace{5pt}
\label{table:policy_hyp}
\begin{center}
\scalebox{0.75}
{
\begin{tabular}{lcccccccc}
    \toprule
    \tt{Task} & \tt{IQL} $\alpha$ & \tt{CRL} $\alpha$ & \tt{SAC+BC} $\alpha$ & \tt{FQL} $\alpha$ & \tt{n-SAC+BC} $\alpha$ & \tt{HIQL} $\alpha$ & \tt{SHARSA} $N$ & \tt{DSHARSA} $N$ \\
    \midrule
    \tt{cube-octuple} & $10$ & $3$ & $10$ & $3$ & $0.1$ & $10$ & $32$ & $32$ \\
    \tt{puzzle-4x5} & $1$ & $1$ & $0.3$ & $3$ & $0.1$ & $3$ & $32$ & $32$ \\
    \tt{puzzle-4x6} & $1$ & $1$ & $0.3$ & $3$ & $0.1$ & $3$ & $32$ & $32$ \\
    \tt{humanoidmaze-giant} & $0.3$ & $0.3$ & $0.1$ & $3$ & $0.03$ & $3$ & $32$ & $32$ \\
    \tt{cube-double} & $3$ & $10$ & $1$ & - & - & - & - & - \\
    \tt{puzzle-4x4} & $1$ & $3$ & $0.3$ & - & - & - & - & - \\
    \bottomrule
\end{tabular}
}
\end{center}
\end{table}


\clearpage

\begin{table}[t!]
\caption{
\footnotesize
\textbf{Full results (at {\color{myblue}$\mathbf{1}$M} epoch).}
}
\vspace{5pt}
\label{table:results_1}
\centering
\scalebox{0.52}
{

\begin{tabular}{lllccccccccc}
\toprule
\tt{Environment} & \tt{Dataset Size} & \tt{Task} & \tt{FBC} & \tt{IQL} & \tt{CRL} & \tt{SAC+BC} & \tt{n-SAC+BC} & \tt{HFBC} & \tt{HIQL} & \tt{SHARSA} & \tt{DSHARSA} \\
\midrule
\multirow[c]{24}{*}{\texttt{cube-octuple-play-oraclerep-v0}} & \multirow[c]{6}{*}{\texttt{1M}} & \texttt{task1} & $\mathbf{0}$ {\tiny $\pm 0$} & $\mathbf{0}$ {\tiny $\pm 0$} & $\mathbf{0}$ {\tiny $\pm 0$} & $\mathbf{0}$ {\tiny $\pm 0$} & $\mathbf{0}$ {\tiny $\pm 0$} & $\mathbf{0}$ {\tiny $\pm 0$} & $\mathbf{0}$ {\tiny $\pm 0$} & $\mathbf{0}$ {\tiny $\pm 0$} & $\mathbf{0}$ {\tiny $\pm 0$} \\
 &  & \texttt{task2} & $\mathbf{0}$ {\tiny $\pm 0$} & $\mathbf{0}$ {\tiny $\pm 0$} & $\mathbf{0}$ {\tiny $\pm 0$} & $\mathbf{0}$ {\tiny $\pm 0$} & $\mathbf{0}$ {\tiny $\pm 0$} & $\mathbf{0}$ {\tiny $\pm 0$} & $\mathbf{0}$ {\tiny $\pm 0$} & $\mathbf{0}$ {\tiny $\pm 0$} & $\mathbf{0}$ {\tiny $\pm 0$} \\
 &  & \texttt{task3} & $\mathbf{0}$ {\tiny $\pm 0$} & $\mathbf{0}$ {\tiny $\pm 0$} & $\mathbf{0}$ {\tiny $\pm 0$} & $\mathbf{0}$ {\tiny $\pm 0$} & $\mathbf{0}$ {\tiny $\pm 0$} & $\mathbf{0}$ {\tiny $\pm 0$} & $\mathbf{0}$ {\tiny $\pm 0$} & $\mathbf{0}$ {\tiny $\pm 0$} & $\mathbf{0}$ {\tiny $\pm 0$} \\
 &  & \texttt{task4} & $\mathbf{0}$ {\tiny $\pm 0$} & $\mathbf{0}$ {\tiny $\pm 0$} & $\mathbf{0}$ {\tiny $\pm 0$} & $\mathbf{0}$ {\tiny $\pm 0$} & $\mathbf{0}$ {\tiny $\pm 0$} & $\mathbf{0}$ {\tiny $\pm 0$} & $\mathbf{0}$ {\tiny $\pm 0$} & $\mathbf{0}$ {\tiny $\pm 0$} & $\mathbf{0}$ {\tiny $\pm 0$} \\
 &  & \texttt{task5} & $\mathbf{0}$ {\tiny $\pm 0$} & $\mathbf{0}$ {\tiny $\pm 0$} & $\mathbf{0}$ {\tiny $\pm 0$} & $\mathbf{0}$ {\tiny $\pm 0$} & $\mathbf{0}$ {\tiny $\pm 0$} & $\mathbf{0}$ {\tiny $\pm 0$} & $\mathbf{0}$ {\tiny $\pm 0$} & $\mathbf{0}$ {\tiny $\pm 0$} & $\mathbf{0}$ {\tiny $\pm 0$} \\
\bottomrule
\end{tabular}

}
\end{table}

\begin{table}[t!]
\caption{
\footnotesize
\textbf{Full results (averaged over {\color{myblue}$\mathbf{4.5}$M}, {\color{myblue}$\mathbf{4.75}$M}, and {\color{myblue}$\mathbf{5}$M} epochs).}
}
\vspace{5pt}
\label{table:results_2}
\centering
\scalebox{0.52}
{

\begin{tabular}{lllccccccccc}
\toprule
\tt{Environment} & \tt{Dataset Size} & \tt{Task} & \tt{FBC} & \tt{IQL} & \tt{CRL} & \tt{SAC+BC} & \tt{n-SAC+BC} & \tt{HFBC} & \tt{HIQL} & \tt{SHARSA} & \tt{DSHARSA} \\
\midrule
\multirow[c]{24}{*}{\texttt{cube-octuple-play-oraclerep-v0}} & \multirow[c]{6}{*}{\texttt{1M}} & \texttt{task1} & $\mathbf{0}$ {\tiny $\pm 0$} & $\mathbf{0}$ {\tiny $\pm 0$} & $\mathbf{0}$ {\tiny $\pm 0$} & $\mathbf{0}$ {\tiny $\pm 0$} & $\mathbf{0}$ {\tiny $\pm 0$} & $\mathbf{0}$ {\tiny $\pm 0$} & $\mathbf{0}$ {\tiny $\pm 0$} & $\mathbf{0}$ {\tiny $\pm 0$} & $\mathbf{0}$ {\tiny $\pm 0$} \\
 &  & \texttt{task2} & $\mathbf{0}$ {\tiny $\pm 0$} & $\mathbf{0}$ {\tiny $\pm 0$} & $\mathbf{0}$ {\tiny $\pm 0$} & $\mathbf{0}$ {\tiny $\pm 0$} & $\mathbf{0}$ {\tiny $\pm 0$} & $\mathbf{0}$ {\tiny $\pm 0$} & $\mathbf{0}$ {\tiny $\pm 0$} & $\mathbf{0}$ {\tiny $\pm 0$} & $\mathbf{0}$ {\tiny $\pm 0$} \\
 &  & \texttt{task3} & $\mathbf{0}$ {\tiny $\pm 0$} & $\mathbf{0}$ {\tiny $\pm 0$} & $\mathbf{0}$ {\tiny $\pm 0$} & $\mathbf{0}$ {\tiny $\pm 0$} & $\mathbf{0}$ {\tiny $\pm 0$} & $\mathbf{0}$ {\tiny $\pm 0$} & $\mathbf{0}$ {\tiny $\pm 0$} & $\mathbf{0}$ {\tiny $\pm 0$} & $\mathbf{0}$ {\tiny $\pm 0$} \\
 &  & \texttt{task4} & $\mathbf{0}$ {\tiny $\pm 0$} & $\mathbf{0}$ {\tiny $\pm 0$} & $\mathbf{0}$ {\tiny $\pm 0$} & $\mathbf{0}$ {\tiny $\pm 0$} & $\mathbf{0}$ {\tiny $\pm 0$} & $\mathbf{0}$ {\tiny $\pm 0$} & $\mathbf{0}$ {\tiny $\pm 0$} & $\mathbf{0}$ {\tiny $\pm 0$} & $\mathbf{0}$ {\tiny $\pm 0$} \\
 &  & \texttt{task5} & $\mathbf{0}$ {\tiny $\pm 0$} & $\mathbf{0}$ {\tiny $\pm 0$} & $\mathbf{0}$ {\tiny $\pm 0$} & $\mathbf{0}$ {\tiny $\pm 0$} & $\mathbf{0}$ {\tiny $\pm 0$} & $\mathbf{0}$ {\tiny $\pm 0$} & $\mathbf{0}$ {\tiny $\pm 0$} & $\mathbf{0}$ {\tiny $\pm 0$} & $\mathbf{0}$ {\tiny $\pm 0$} \\
\bottomrule
\end{tabular}

}
\end{table}

\clearpage

\end{document}